%% file: arxiv_main.tex
\documentclass[10pt,twocolumn]{article} 
\usepackage{simpleConference}
\usepackage{times}
\usepackage{graphicx}
\usepackage{amssymb}
\usepackage{url,hyperref}
\usepackage{multirow}
\usepackage{array}
\usepackage{commath}
\usepackage{pdfpages}
\usepackage{placeins}
\usepackage{cleveref}
\usepackage{booktabs} 
\usepackage{subcaption}

\begin{document}

\title{PhysicsGen: Can Generative Models Learn \\ from Images to Predict Complex Physical Relations?}

\author{%
Martin Spitznagel$^{1}$, Jan Vaillant$^{1,2}$, Janis Keuper$^{1,3}$ \\
\\
$^{1}$ Institute for Machine Learning and Analytics (IMLA), Offenburg University, Germany \\
$^{2}$ Herrenknecht AG \\
$^{3}$ University of Mannheim, Germany \\
\texttt{firstname.lastname@hs-offenburg.de}\\ \\
\textbf{Accepted at IEEE/CVF Conference on Computer Vision and Pattern Recognition (CVPR) 2025}
}

\maketitle
\begin{abstract}
The image-to-image translation abilities of generative learning models have recently made significant progress in the estimation of complex (steered) mappings between image distributions. While appearance based tasks like image in-painting or style transfer have been studied at length, we propose to investigate the potential of generative models in the context of physical simulations. Providing a dataset of 300k image-pairs and baseline evaluations for three different physical simulation tasks, we propose a benchmark to investigate the following research questions: i) are generative models able to learn complex physical relations from input-output image pairs? ii) what speedups can be achieved by replacing differential equation based simulations? While baseline evaluations of different current models show the potential for high speedups (ii), these results also show strong limitations toward the physical correctness (i). This underlines the need for new methods to enforce physical correctness.
Data, baseline models and evaluation code:
\url{http://www.physics-gen.org}.
\end{abstract}

\input{sections/01_introduction}
\input{sections/02_sound}

\input{sections/03_lens_distortion}
\input{sections/04_ball}
\input{sections/05_discussion}

\section*{Funding Acknowledgement}
The authors acknowledge the financial support by the German Federal 
Ministry of Education and Research (BMBF) in the program “Forschung an Fachhochschulen in Kooperation mit Unternehmen (FH-Kooperativ)” within the joint project "KI-Bohrer" under 
grant 13FH525KX1.

\bibliographystyle{abbrv}
\bibliography{cvpr_2025}
\newpage
\appendix
\clearpage
\section*{PhysicsGen -- Supplementary Material}

This Appendix provides additional details and supplementary material supporting the main content of the paper. Below are the sections included in this Appendix:

\begin{itemize}
    \item \textbf{Section A: Training Setup} \\
    Provides detailed information about the configurations and parameters used during the training phase of the models.    
    \item \textbf{Section B: Sound Propagation Appendix}
    \begin{itemize}
        \item \ref{app:sound_eval_metrics} Evaluation Metrics: Outlines the criteria and methods used to assess the performance and accuracy of sound propagation models within the simulation framework.
        \item \ref{app:location_sampling} Location Sampling: Describes the methods used to select and sample various urban locations.
        \item \ref{app:receiver_placement} Receiver Placement:  Details the strategies for placing receivers in the simulation environment.
        \item \ref{app:sound_physics} Sound Physics: Description of the physics behind the sound propagation.
        \item \ref{app:sound_runtime} Runtime Analysis: Runtime Analysis of sound propagation tasks vs. simulation framework.
        \item \ref{app:pipeline} Scalable Simulation Pipeline: Discusses the scalable simulation pipeline developed for processing sound propagation data.
        \item \ref{app:results} Additional Qualitative Results: Presents additional qualitative results, showcasing further analyses, visualizations, and interpretations of the sound propagation task.
    \end{itemize}
    \item \textbf{Section~\ref{a:lens}: Lens Appendix}
    \begin{itemize}
        \item ~\ref{app:face_landmark_metrics} Evaluation Metrics for Facial Landmark Detection: Outlines the metrics such as Euclidean distance and mean absolute errors to evaluate facial landmark detection accuracy.
        \item ~\ref{app:lens_physics} Lens Physics: Explains the mathematical models for simulating lens distortions, focusing on radial and tangential components.
        \item ~\ref{app:lens_distortion} Lens Distortion Application Pipeline: Outlines the process and tools used to apply lens distortions to dataset images, highlighting the use of Python and OpenCV.
        \item ~\ref{app:lens_results} Additional Qualitative Results: Visual results, comparing the model’s landmark predictions against actual distorted landmarks to assess accuracy.
    \end{itemize}
    \item \textbf{Section~\ref{a:ball}: Ball Appendix}
    \begin{itemize}
        \item ~\ref{a:ball_physics} Ball physics and kinematics: Description of the physics behind the rolling and bouncing movement of the ball
        \item ~\ref{a:BB_training_all} Bouncing ball: Further evaluation and analysis of the results of the three generative approaches for the \textbf{bouncing case}. Here the detailed result tables, some more predictions samples and some typical errors of the network can be found.
        \item ~\ref{a:RB_training_all} Rolling ball: Further evaluation and analysis of the results of the three generative approaches for the \textbf{rolling case}. Here the detailed result tables, some more predictions samples and some typical errors of the network can be found.
    \end{itemize}
    \item \textbf{Datacard} \ref{sec:datacard}.
\end{itemize}

\appendix
\input{appendix/model_architecture}

\input{appendix/sound_appendix}
\input{appendix/lens_appendix}
\input{appendix/ball_appendix}
\input{appendix/datacard}

\end{document}

%% file: sections/01_introduction.tex
\begin{figure}
\centering
\includegraphics[width=0.90\linewidth]{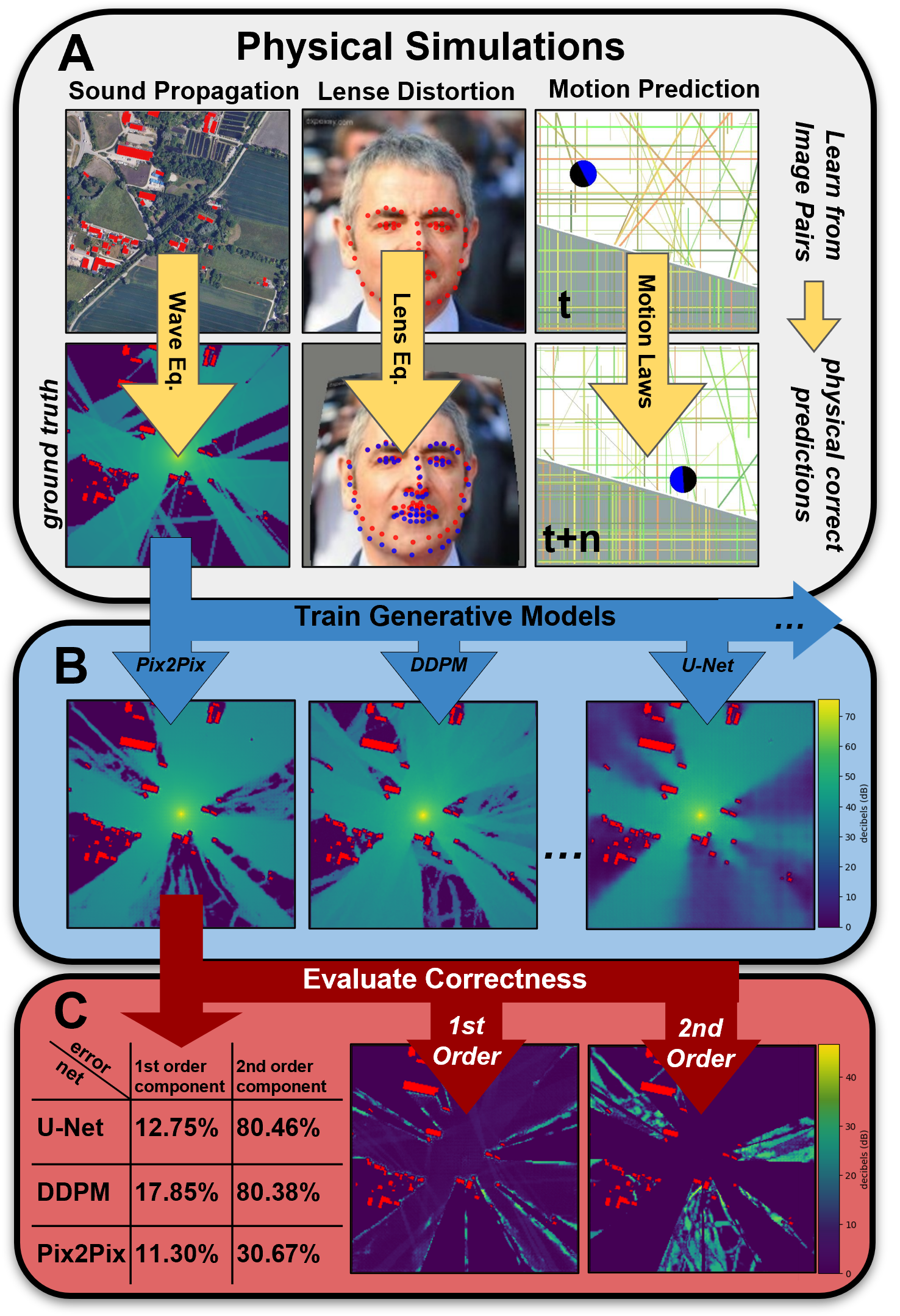}
\caption{\small
Overview of the physical problems, baseline generative models and their evaluation. \textbf{A:} We introduce three complex physical simulation tasks with 100k input-output image pairs each, providing ground truth simulations based on differential equations with varying complexity. \textbf{B:} We evaluate all tasks on independently trained image translation models; only results for the sound propagation task are visualized in this figure.
 \textbf{C:} While the evaluation of the baseline models shows a general ability of generative image models to learn physical relations from images, we observe significant performance drops for tasks that require a higher order term in the differential equations of their simulation.}  
\label{fig:overview}
\vspace{-60px}
\end{figure}
\section{Introduction}
The numerical simulation of physical processes and relations is a crucial tool in a wide range of scientific and engineering tasks. Traditionally depending on the solution of various types of differential equations, research of physical simulation methods is increasingly shifting towards data-driven and learnable approaches~\cite{carleo2019machine}.\\ 
In this work, we focus on a narrow but practically important sub-field: the inference of physical relations from images. Extending Tenenbaum's seminal idea of a \textit{intuitive physics engine}~\cite{battaglia2013simulation}, where machine learning models have been trained to predict coarse physical properties from images, we investigate the capabilities of modern generative models to learn detailed, complex and physically correct mappings from image pairs. Given such abilities, generative models like GANs~\cite{goodfellow2014generative} and Diffusion models~\cite{ho2020denoising} could be utilized to introduce hard to formalize but observable relations into physical models. Due to the fast inference times, generative models could also be used to accelerate simulations~\cite{thuerey2021pbdl}.      

\noindent\textbf{Introducing the \textit{PhysicsGen} Benchmark.} A first step towards such data-driven models is a thorough analysis of the capabilities of current generative approaches. Hence, we introduce a novel benchmark that allows to systematically investigate the inference of a diverse set of physical problems from images (as visualized in \cref{fig:overview}):
\begin{itemize}
\item We introduce \textit{\textbf{PhysicsGen}}, a \textbf{novel Benchmark} for the evaluation of generative image models learning to \textbf{infer physical simulations tasks from images}.  
\item \textit{PhysicsGen} provides a \textbf{Dataset} with a total of \textbf{300k image-pairs}, representing three diverse physical simulation problems: i) iterative wave propagation (\cref{sec:sound}), ii) closed form lens distortion (\cref{sec:lens}) and iii) time series prediction of motion dynamics (\cref{sec:ball}).
\item We provide a \textbf{baseline evaluation} that measures speedups and physical accuracy compared to full physical simulations across multiple generative image-to-image models. The evaluated models include various generative image-to-image models: Generative Adversarial Networks (GANs)~\cite{goodfellow2014generativeadversarialnetworks}, specifically Pix2Pix~\cite{pix2pix}; an U-Net~\cite{unet}; Convolutional Autoencoders (convAE) and Variational Autoencoders (VAEs)~\cite{kingma2022autoencodingvariationalbayes}. We also assess diffusion models, including Denoising Diffusion Probabilistic Models (DDPMs)~\cite{ddpm}, Stable Diffusion~\cite{rombach2022highresolutionimagesynthesislatent}, and Denoising Diffusion Bridge Models (DDBMs)~\cite{zhou2023denoisingdiffusionbridgemodels}. (see \cref{app:gen_models} for details of the model architectures and training - the model architectures and training procedures are identical for all three tasks). 
\item Our \textbf{Analysis} of the baseline results shows potentially high speed-ups at good accuracies for simple 1st order simulation tasks. However, our experiments also show that current generative models have \textbf{fundamental problems to learn higher order physical relations}. 
\end{itemize}

\subsection{Related Work.}
The integration of learnable methods into physical modeling is a wide field with a rich body of literature which is very hard to capture in the limited space of this paper. In the following, we try to refer to the most important works in direct relation to our benchmark and some more general key concepts. We refer the reader to~\cite{thuerey2021pbdl} and~\cite{carleo2019machine} for a comprehensive overview of the field.\\ 
\noindent\textbf{Generative models for physical simulations.} The integration of physical principles within generative models is an active and burgeoning area of research. Recent models such as PUGAN~\cite{PUGAN} and FEM-GAN~\cite{FEM-GAN} have demonstrated the potential of combining GANs with physical modeling to enhance performance in environments governed by complex physical laws. Advances in fluid dynamics and structural system identification using physics-guided GANs have shown significant improvements in efficiency and precision by incorporating physics-based loss functions and simulations~\cite{Kim2018Deep,YU2024122339}.\\
\noindent\textbf{Predicting physical relations.}
Also, significant progress has been made in understanding physics through machine learning. Initial research demonstrated that models could comprehend the dynamics of block towers, progressing beyond simple memorization to genuinely understanding physical interactions~\cite{lerer2016learning}. Practical applications have been explored, including fall detection using body part tracking through machine learning~\cite{li2016fall}. Additionally, advancements in generating physically plausible human animations highlight the potential of incorporating physical principles into machine learning frameworks~\cite{PhysDiff}.\\
\noindent\textbf{Grey-Box Models.} Physics-guided approaches in AI have laid a solid foundation for incorporating physical laws into model training. Prior research has demonstrated the effectiveness of embedding physical constraints within AI models to enhance their performance and reliability~\cite{PhysicsGuided,PhysicallyGrounded}. In generative models, diffusion models have shown promise but often face challenges when generating complex geometrical structures due to issues like multi-modality and noise sensitivity. The development of Geometric Bayesian Flow Networks addresses these challenges by enhancing the robustness and efficiency of molecule geometry modeling~\cite{song2024unified}.\\
\noindent\textbf{Physical Simulations for Reinforcement Learning.} Deep reinforcement learning settings benefit from a substantial availability of physics-based environments and datasets, which facilitate the training of models to understand and predict physical phenomena~\cite{duan2016benchmarking}. Resources such as OpenAI Gym, Robosuite, and PyBullet provide robust platforms for developing and testing reinforcement learning models in simulated physical settings~\cite{zhu2022robosuite,greff2022kubric,brockman2016openai}. However, in the domain of generative AI, there is a notable scarcity of physics-informed datasets and environments, limiting the ability to train models on complex physical systems.\\
\section{Benchmark Tasks Overview}
\label{sec:benchmark_overview}
Selecting an appropriate set of tasks for the \textit{PhysicsGen} benchmark is crucial to comprehensively evaluate the capabilities of generative image models in inferring complex physical relations. Given the vast diversity of possible physical simulation tasks, it is impractical to encompass the entire spectrum. Therefore, we strategically selected three distinct tasks based on the following criteria:

\begin{itemize}
    \item \textbf{Diversity of Physical Problems}: To cover different types of physical phenomena, ensuring that models are tested across various domains.
    \item \textbf{Variety of Numerical Solution Strategies}: Incorporating both iterative and closed-form (non-iterative) solvers to assess models' adaptability to different computational approaches.
    \item \textbf{Simplicity of Evaluation Metrics}: Facilitating straightforward and effective model performance evaluation.
    \item \textbf{Accessibility to the ML Community}: Choosing intuitively understandable tasks without requiring specialized physics knowledge, thereby making the benchmark widely applicable to the machine learning community.
\end{itemize}

\noindent Based on these objectives, we introduce the following three simulation tasks within the \textit{PhysicsGen} benchmark:

\noindent\textbf{Urban Sound Propagation}: This task models the behavior of sound waves in urban environments, accounting for phenomena such as diffraction and reflections (\cref{sec:sound}). By providing 100,000 input-output image pairs based on differential equation-based simulations, it challenges generative models to accurately capture intricate sound distribution patterns influenced by urban structures.

\noindent\textbf{Lens Distortion}: Focusing on optical aberrations, this task utilizes the Brown-Conrady distortion model to simulate lens-induced geometric distortions in images (\cref{sec:lens}). With 100,000 image pairs depicting varying degrees of distortion, models are tasked with learning precise, closed-form physical relations that alter image geometry based on lens parameters.

\noindent\textbf{Dynamics of Rolling and Bouncing Movements}: This task involves simulating the motion of a rolling or bouncing ball on an inclined surface, incorporating both linear and rotational dynamics (\cref{sec:ball}). Providing 100,000 image pairs that capture the ball's position and rotation over time, it assesses models' ability to predict time-series dynamics and handle higher-order physical relations inherent in motion equations.

\noindent Collectively, these tasks offer a diverse framework for assessing the strengths and limitations of generative models in learning and replicating complex physical simulations from image data. The subsequent sections delve into each task in detail, outlining dataset generation, underlying physics, evaluation metrics, and baseline model performances.

%% file: sections/02_sound.tex
\section{Urban Sound Propagation\label{sec:sound}}
The simulation of wave propagation in complex scenes, including diffraction and reflection at multiple objects is a representative task for a wide range of physical problems which are typically solved via (expensive) iterative solving of higher order differential equations. Due to its intuitive setting and accessible data, we investigate wave propagation problems by a concrete example where sound is propagated in urban environments.\\
\noindent\textbf{Dataset.} 
We sampled $25k$ geo-locations across 5 cities, leveraging open-source geo-data from the \textit{Overpass API}\footnote{\url{http://overpass-api.de/api/map}} and processed it with \textit{GeoPandas}~\cite{kelsey_jordahl_2020_3946761} to represent urban environments in a 500m$^2$ area with buildings and open spaces encoded in black and white pixels, respectively. Our selection criteria ensured diverse urban scenarios by mandating a minimum of 10 buildings within 200 meters and a 50-meter clearance from buildings to the sample point.
\begin{figure*}
\setlength{\belowcaptionskip}{-20pt}
\centering
\begin{minipage}{0.625\linewidth} 
    \centering
    \includegraphics[width=\linewidth]{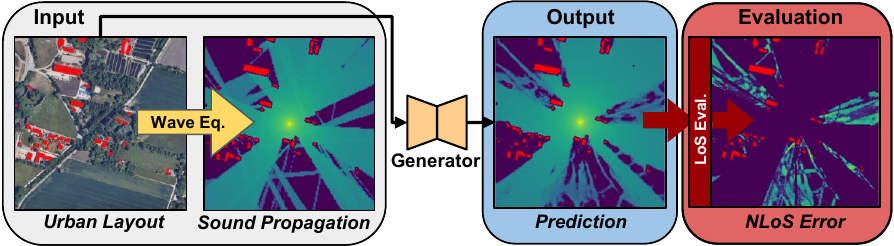}
\end{minipage}%
\hfill
\begin{minipage}{0.35\linewidth} 
    \caption{\small The sampling pipeline for the sound propagation dataset utilizes the \textit{NoiseModelling} framework~\cite{noisemodelling_framework} to generate sound propagation maps based on specific urban layouts. The generators are then trained to replicate these sound propagation patterns for given locations and source parameters. Predictions are evaluated by specifically analyzing errors in relation to the line of sight (see \cref{app:sound_eval_metrics} for details).}
    \label{fig:sound_pipeline}
\end{minipage}
\end{figure*}
We utilized the \textit{NoiseModelling v4} framework~\cite{noisemodelling_framework}, adhering to the \textit{CNOSSOS-EU} standards~\cite{CNOSSOS-EU}, to simulate sound propagation maps, placing sound receivers on a grid with a step-size of 5 meters and at building boundaries, excluding indoor placements (see \cref{app:receiver_placement} for details). The resulting simulation output was then interpolated into 512x512 or 256x256 pixel maps, representing decibel levels on a $[0,255]$ scale.
Four simulation tasks (\textit{Baseline, Diffraction, Reflection, Combined}) were defined to model sound propagation under varying conditions, accumulating in a dataset of $100,000$ samples. The \textit{Baseline} task models a constant 95 dB noise source at 500 Hz, excluding diffraction and reflection. The \textit{Diffraction} task adds horizontal sound wave diffraction, the \textit{Reflection} task simulates sound reflections with a standard absorption coefficient, and the \textit{Combined} task introduces variable sound levels (60-115 dB) and environmental factors, incorporating both reflection and diffraction to mimic realistic urban sound behavior. Refer to \cref{fig:sound_pipeline} and \cref{app:pipeline} for a detailed description and visualization of the dataset generation pipeline.\\
\noindent\textbf{Physics of Sound Propagation.}
Mathematically, the propagation of sound over time is described via partial differential wave equations. Due to space constraints and the practical nature of our problem setting, we will neglect the derivation from continuous wave equations and directly focus on the discrete and iterative implementations of sound propagation which have been applied for our ground-truth simulations.
Following \cite{vorlander2007auralization}, for a discrete set of receivers $R$, the amplitude  $L^{j}_{{R_k}}$ of receiver $R_k$ at frequency $j$ is computed via iterative differences:
\begin{equation}
    L^{j}_{{R_k}} = L^{j}_{W} - A_{div_{R_{k}}} - A^{j}_{atm_{R_k}} - A^{j}_{dif_{R_{k}}} \quad (j, k \in \mathbb{N}),\label{eq:sound}
\end{equation}
where $L^{j}_{W}$ represents the source level, $A_{div_{R_{k}}}$ captures the geometrical spreading, $A^{j}_{atm_{R_k}}$ denotes the atmospheric absorption, and $A^{j}_{dif_{R_{k}}}$ models diffraction. Note that the ground effect $A^{j}_{grd_{R_{k}}}$ is neglected in our study.\\
\noindent Additionally, the model accounts for reflections by adjusting the power level $L^{j}_{W}$ based on the absorption coefficient $\alpha_{vert}$ of the surfaces involved. This adjustment is performed using the equation
\begin{equation}
\label{eq:reflection}
L^{(n_{ref})}_{W} =  L^{(n_{ref} - 1)}_{{W}} + n_{ref} \times 10 \log_{10} (1 -{\alpha}_{vert})
\end{equation}
where $n_{ref}$ indicates the number of reflections considered. Specular reflections are modeled using the image receiver method, which provides a computationally efficient way to account for the angle of incidence being equal to the angle of reflection~\cite{vorlander2007auralization}. The sound level at each receiver reflects the cumulative effect of direct, diffracted, and reflected sound paths. As the number of reflections increases, the complexity of calculating the power level $L^{j}_{W}$ also increases, growing in $O(N^2)$, where $N$  denotes the number of reflective surfaces considered in the computation.\\
\noindent\textbf{Evaluation.}
Our baseline evaluation focuses on the ability of standard image-to-image generative models to accurately predict sound propagations. Prediction quality is determined by the pixel-wise difference between actual and predicted sound distributions. We use Mean Absolute Error (MAE) to quantify the average magnitude of prediction errors. Additionally, we introduce Weighted Mean Absolute Percentage Error (wMAPE) to specifically penalize errors in predictions that inaccurately show high sound amplitudes in regions expected to have low amplitudes, such as areas behind buildings. To evaluate model performance in capturing sound reflections or diffractions, metrics were specifically calculated for areas both within and outside the direct line of sight (LoS and NLoS) from the sound source. This methodology allows us to assess model predictions for sound propagations through direct paths as well as reflections and diffractions.

\begin{table}[h]
\footnotesize
  \caption{\small Quantitative evaluation across all tasks for all architectures with a batch size of 16 during inference. The complete results containing the \textbf{Combined} task are given in \cref{app:results}. Best overall results in \textbf{bold}, second best \underline{underlined}.}
  \label{tab:combined_metrics}
  \setlength\tabcolsep{4pt}
  \centering
  \begin{tabular}{llccccc}
    \textbf{Model} & \textbf{Task} & \multicolumn{2}{c}{\textbf{MAE} $\downarrow$} & \multicolumn{2}{c}{\textbf{wMAPE} $\downarrow$} & \textbf{Runtime/} $\downarrow$ \\
    & & \textbf{LoS} & \textbf{NLoS} & \textbf{LoS} & \textbf{NLoS} & \textbf{Sample (ms)}\\
      \hline
      Simulation & Base & 0.00 & 0.00 & 0.00 & 0.00 & 204700 \\
      convAE & Base & 3.67 & 2.74 & 20.24 & 67.13 & 0.128  \\ 
      VAE \cite{kingma2022autoencodingvariationalbayes} & Base & 3.92 & 2.84 & 21.33 & 75.58 & 0.124\\
      UNet \cite{unet} & Base & 2.29 & 1.73 & \underline{12.91} & \underline{37.57} & 0.138 \\
      Pix2Pix \cite{pix2pix} & Base & \underline{1.73} & \underline{1.19} & \textbf{9.36} & \textbf{6.75} & 0.138 \\
      DDPM \cite{ddpm} & Base & 2.42 & 3.26 & 15.57 & 51.08 & 3986.353 \\
      SD(w.CA) \cite{rombach2022highresolutionimagesynthesislatent} & Base & 3.76 & 3.34 & 17.42 & 35.18 & 2961.027\\
      SD & Base & 2.12 & \textbf{1.08} & 13.23 & 32.46 & 2970.86 \\
      DDBM \cite{zhou2023denoisingdiffusionbridgemodels}& Base & \textbf{1.61} & 2.17 & 17.50 & 65.24 & 3732.21 \\
      \hline
      Simulation & Dif. & 0.00 & 0.00 & 0.00 & 0.00 & 206000 \\
      convAE & Dif. & 3.59 & 8.04 & 13.77 & 32.09 & 0.128 \\
      VAE & Dif. & 3.92 & 8.22 & 14.46 & 32.57 & 0.124 \\
      UNet & Dif. & \underline{0.94} & \textbf{3.27} & \underline{4.22} & 22.36 & 0.138 \\
      Pix2Pix & Dif. & \textbf{0.91} & \underline{3.36} & \textbf{3.51} & \textbf{18.06} & 0.138\\
      DDPM & Dif. & 1.59 & \textbf{3.27} & 8.25 & \underline{20.30} & 3986.353  \\
      SD(w.CA) & Dif. & 2.46 & 7.72 & 10.14 & 31.23 & 2961.027 \\
      SD & Dif. & 1.33 & 5.07 & 8.15 & 24.45 & 2970.86 \\
      DDBM & Dif. & 1.35 & 3.35 & 11.22 & 23.56 & 3732.21 \\
      \hline
      Simulation & Refl. & 0.00 & 0.00 & 0.00 & 0.00 & 251000 \\
      convAE & Refl. & 3.83 & 6.56 & 20.67 & 93.54 & 0.128  \\
      VAE & Refl.  & 4.15 & 6.32 & 21.57 & 92.47 & 0.124 \\
      UNet & Refl.  & 2.29 & 5.72 & \underline{12.75} & 80.46 & 0.138\\
      Pix2Pix & Refl.  & \underline{2.14} & \textbf{4.79} & \textbf{11.30} & \textbf{30.67} & 0.138\\
      DDPM & Refl.  & 2.74 & 7.93 & 17.85 & 80.38 & 3986.353 \\
      SD(w.CA) & Refl.  & 3.81 & 6.82 & 19.78 & 81.61 & 2961.027 \\
      SD & Refl. & 2.53 & \underline{5.26} & 15.04 & \underline{55.27} & 2970.86 \\
      DDBM & Refl.  & \textbf{1.93} & 6.38 & 18.34 & 79.13 & 3732.21 \\
      \end{tabular}
\end{table}
\begin{figure}[b]
\centering
\includegraphics[width=1\linewidth]{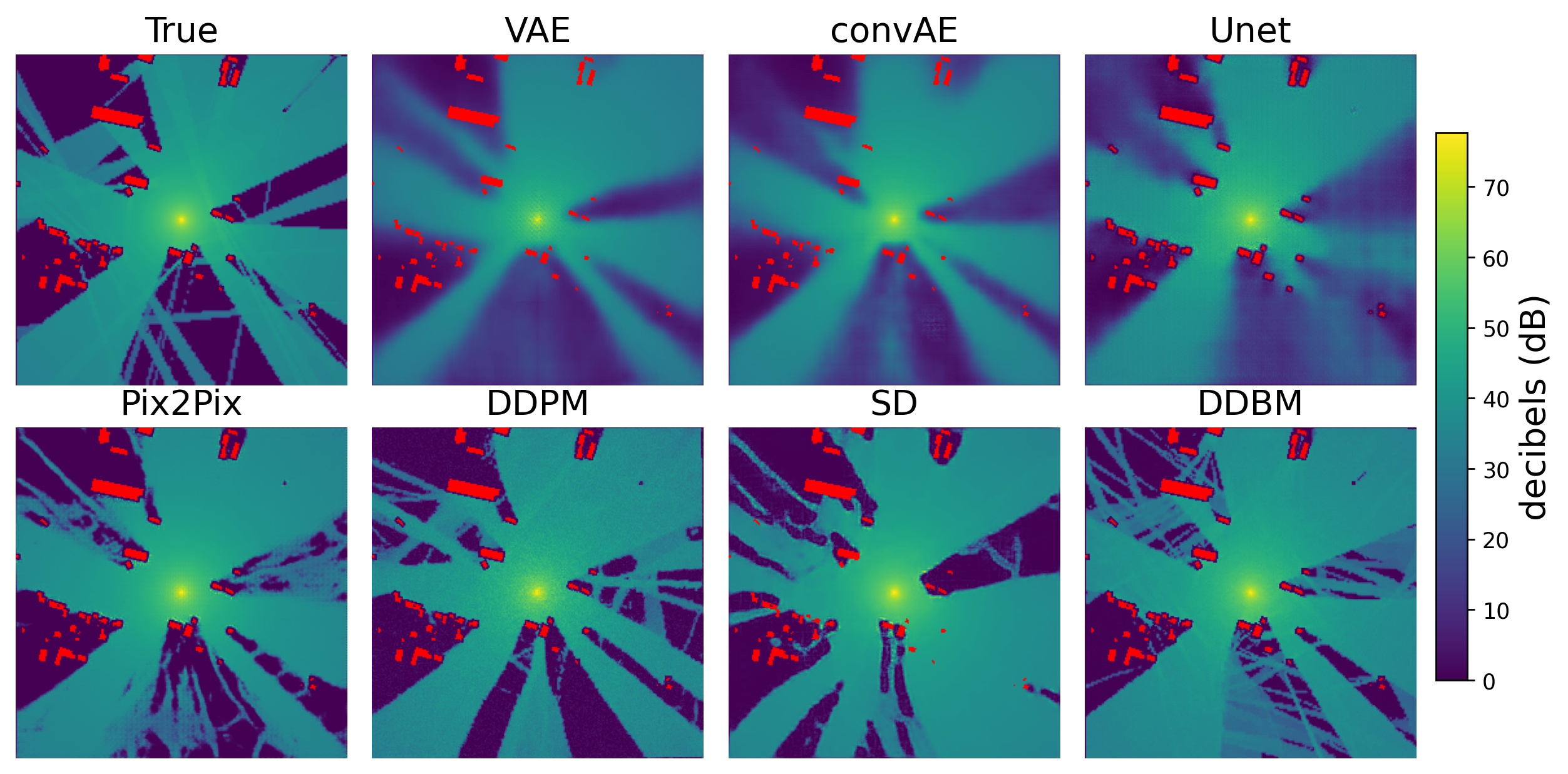}
\caption{\small Qualitative results comparing the ground-truth simulation with the prediction for a single sample within the \textbf{reflection} task. Additional results can be found in \cref{app:results}}
\label{fig:sight_loss}
\end{figure}
\noindent\textbf{Analysis.} 
The results in \cref{tab:combined_metrics} show that generative models significantly outperform traditional sound propagation simulations in processing speed, achieving up to a $20k$ factor improvement in runtime (see \cref{tab:model_simulation_runtime_comparison} in the Appendix for further analysis). Regarding the prediction quality, we observe that models encountering tasks with reflections or diffractions show an increase in MAE for NLoS regions across all tested architectures. For the Pix2Pix model this rise in NLoS error is moderate, while this effect is more pronounced in the UNet and Diffusion models. Models without skip connections, such as VAE and ConvAE, tend to blur most of the image and struggle to capture any reflection patterns accurately. This trend highlights how reflections and diffractions introduce complex higher-order dependencies (see \cref{eq:reflection} and \cref{app:sound_physics}) that complicate the image-to-image sound propagation tasks. Although the models successfully replicate sound reflection patterns visually (see \cref{fig:sight_loss}), there exists a significant gap between its visual outputs and the actual acoustic accuracy. This discrepancy suggests that while visually detailed, the models do not fully capture the complexities of sound physics.

%% file: sections/03_lens_distortion.tex
\section{Lens Distortion\label{sec:lens}}
\begin{figure}[b]
\setlength{\belowcaptionskip}{-20pt}
\centering
\includegraphics[width=1.0\linewidth]{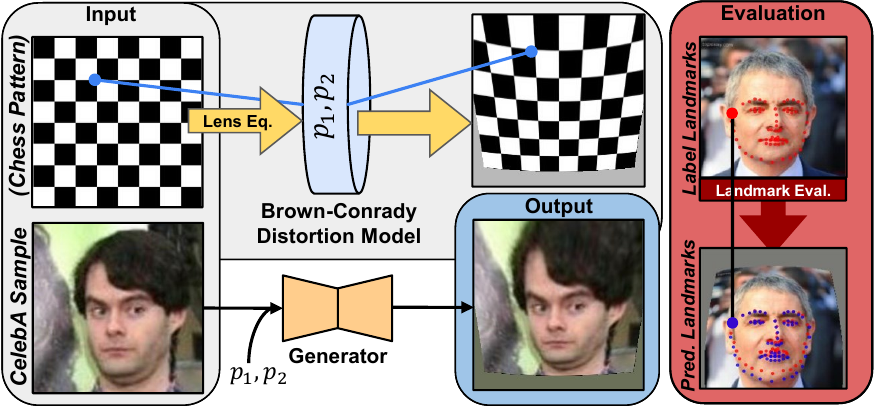}
\caption{\small Sampling and evaluation pipeline for the lens distortion dataset. The Brown-Conrady distortion model generates the true distorted images based on parameters $p_1$ and $p_2$ (we depict a chess pattern for visualization). The conditioned generators are then trained to replicate these distortions for given images and parameters. The models are evaluated by comparing predicted against true facial landmarks, using a 2D facial landmark detection based on the Facial Alignment Network (FAN)~\cite{Bulat_2017}.}
\label{fig:lens_overview}
\end{figure}

\begin{figure}
\centering
\includegraphics[width=1\linewidth]{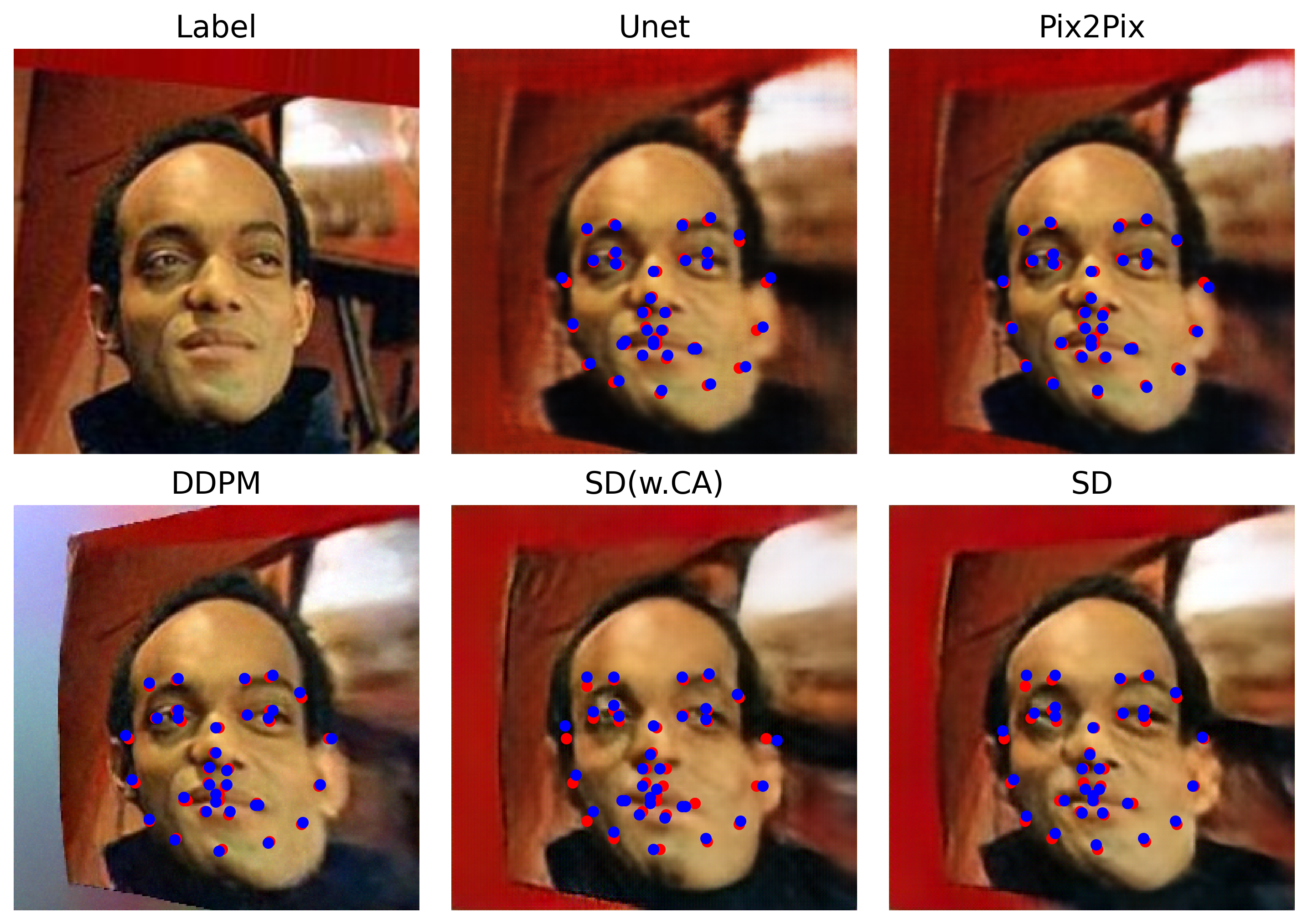}
\caption{\small Qualitative visualization of lens distortion predictions using different generative models on a CelebA dataset sample. $p_2$ distortion: Original image at top-left; subsequent panels show U-Net, Pix2Pix, Diffusion models. Red dots mark actual landmark positions and blue dots for predictions.}
\label{fig:lens_model_preds}
\end{figure}

Our second benchmark is designed to investigate a class of physical problems for which solutions can be computed directly via closed equations, rather than iterative solvers. We find a representative and intuitive task in the modeling of optical lens distortions~\cite{Brown1971CloseRangeCC,Faig1975CALIBRATIONOC}, a common problem in computer vision. Generative models are trained to simulate the appearance of faces distorted by lens effects, using the Brown-Conrady~\cite{brownConrady} model. These distortions are computed using OpenCV~\cite{opencv_library}, providing ground truth for training and evaluation of generative models. \\
\noindent\textbf{Dataset.} 
We leverage a subset of the CelebA dataset~\cite{liu2015faceattributes}, comprising 50,000 facial images, which were cropped to center the faces within a 256x256 pixel frame. The resulting images where split into two distinct groups to isolate the effects of tangential distortion: one focusing on horizontal distortion and the other on vertical distortion. This allows us to specifically measure the impact of each type of distortion on the accuracy of the generative models (see \cref{fig:lens_overview} and Appendix~\ref{a:lens} for more details).\\
\noindent\textbf{Physics of Lens Distortions.}
The Brown-Conrady model is a widely adopted method to describe and correct lens distortions. It provides a mathematical framework to represent both radial and tangential distortions through a series of coefficients. For our study, we concentrate exclusively on the tangential distortion components, utilizing relevant coefficients to model the distortion characteristics of specific lenses. The mathematical representation of tangential distortion is expressed through the following equation, where $x_{dist}$ and $y_{dist}$ denote the distorted coordinates:
\begin{equation}
\begin{aligned}
x_{dist} &= x + \left[2p_1xy + p_2\left(r^2 + 2x^2\right)\right], \\
y_{dist} &= y + \left[p_1\left(r^2 + 2y^2\right) + 2p_2xy\right],
\end{aligned}
\end{equation}
where $r^2 = x^2 + y^2$ represents the squared radial distance from the image center, and $p_1$ and $p_2$ are the coefficients that model the tangential distortion effects. By strategically adjusting $p_1$ and $p_2$ and isolating their effects via controlled nullification, we can identify how each parameter individually contributes to the image distortion:
\begin{equation}
\begin{aligned}
x_{dist} &=
\begin{cases}
x + 2p_1xy & \text{if } p_2 = 0,\\
x + p_2(r^2 + 2x^2) & \text{if } p_1 = 0,
\end{cases} \\
y_{dist} &=
\begin{cases}
y + p_1(r^2 + 2y^2) & \text{if } p_2 = 0,\\
y + 2p_2xy & \text{if } p_1 = 0.
\end{cases}
\end{aligned}
\end{equation}
These conditional equations illustrate the isolated impact of each coefficient, revealing that the correction complexity for $y_{dist}$ escalates with the term $p_1(r^2 + 2y^2)$ when $p_1$ is non-zero and $p_2$ is zero. Conversely, the complexity of $x_{dist}$ correction intensifies with $p_2(r^2 + 2x^2)$ under the condition that $p_2$ is non-zero and $p_1$ is zero.
\\
\noindent\textbf{Evaluation.}
Given the challenges of accurately capturing lens distortion through traditional pixel-based metrics, our evaluation strategy is specifically designed to focus on accurately reproducing designated landmarks in the images. We detect facial landmarks using a 2D facial landmark detection method based on the Facial Alignment Network (FAN)~\cite{Bulat_2017}, applied to the faces in the CelebA dataset (see Appendix~\ref{a:lens} for details). We use these landmarks to assess the geometric distortions introduced by lens settings, represented by the coefficients $p_1$ and $p_2$.
We then compute the Euclidean distance between the actual position of a landmark in the true distorted image and its predicted position in the image generated by the models as a quality metric. This measurement is performed separately for the horizontal (X) and vertical (Y) coordinates, resulting in X Error and Y Error respectively. This approach not only allows us to gauge the accuracy of the models in simulating the specific effects of lens distortion on facial features but also helps in understanding how different distortion coefficients impact the positional accuracy of landmarks along each axis.\\
\noindent\textbf{Analysis.} Our experiments with the CelebA dataset under varying conditions of $p_1$ and $p_2$ showcase different patterns in error rates that align with the theoretical complexity of the distortion model. As shown in \cref{tab:landmark_err} and \cref{fig:lens_model_preds}, the generative models exhibit differing accuracies in landmark prediction under the influence of lens distortion parameters $p_1$ and $p_2$. Notably, the increased complexity in $x_{dist}$ and $y_{dist}$ equations under separate conditions ($p_1$ nonzero vs. $p_2$ nonzero) slightly correlates with heightened error rates in respective dimensions.
The models exhibit higher Y Error when only $p_1$ is active, indicative of the complex correction challenge introduced by the higher order dependency on $y$. Similarly, a slight increase in X error and a slight decrease in Y error are observed when $p_2$ is exclusively nonzero.
%


%% file: sections/04_ball.tex
\section{Dynamics of bouncing movements\label{sec:ball}}
\begin{figure*}
\setlength{\belowcaptionskip}{-20pt}
\centering
\begin{minipage}{0.625\linewidth} 
    \centering
    \includegraphics[width=\linewidth]{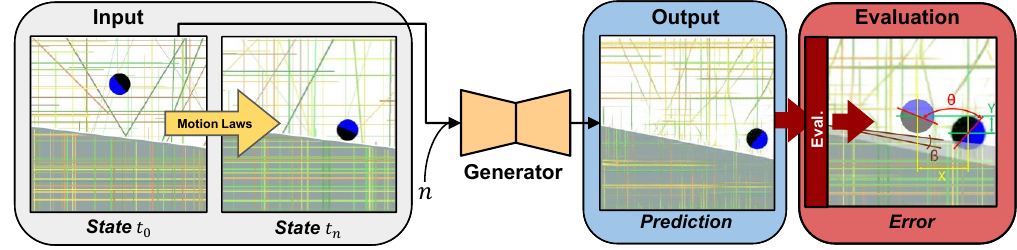}
\end{minipage}%
\hfill
\begin{minipage}{0.35\linewidth} 
    \caption{\small The sampling and evaluation pipeline for the rolling and bouncing ball movements simulates ball trajectories. A physics-based model is used to predict the ball’s movement. Evaluations focus on the accuracy of predictions related to key movement metrics, including bounce height and horizontal displacement (see \cref{a:ball} for details).
}
    \label{fig:ball_overview}
\end{minipage}
\end{figure*}

\begin{figure}
\centering
\includegraphics[width=1\linewidth]{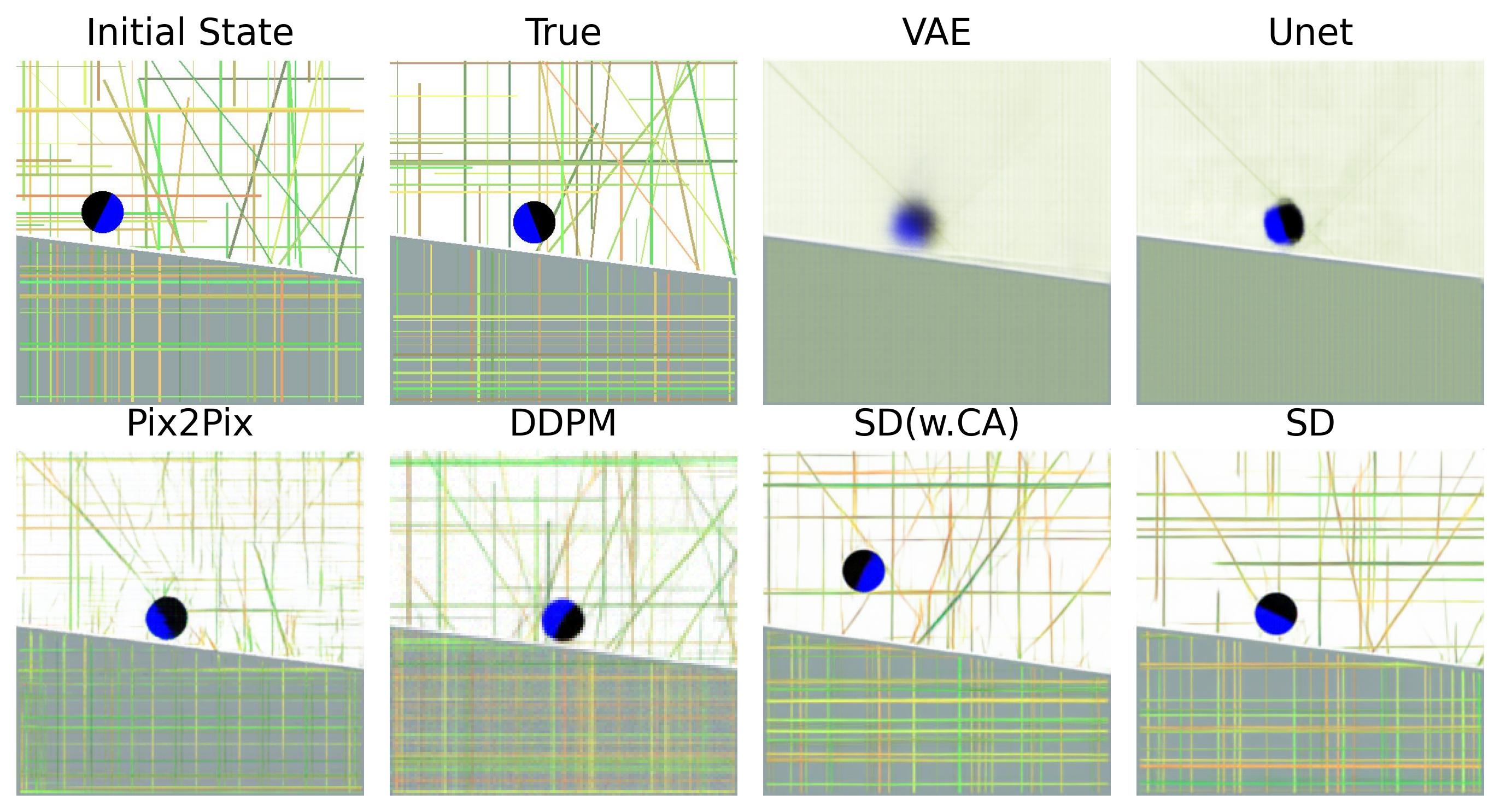}
\caption{\small Qualitative results for predictions of the bouncing ball problem. Initial State and True label image at top-left; subsequent panels show results for VAE, UNet, Pix2Pix and Diffusion models.}
\label{fig:BB_PredImgs_1}
\end{figure}

The third physical problem under investigation is the movement of a rolling or bouncing ball. The aim here is to evaluate the ability of generative models to map kinematic movements of physics - a task which should generalize well to the investigation of various time series problems. Specifically, we define the task to predict the position and rotation of a ball along an inclined surface for a defined time interval after the input image. Only the bouncing case is discussed below; however, all results for the rolling ball can be found in \cref{a:RB_training_all}.
\begin{table}
\footnotesize
  \caption{\small Quantitative X-Y error and shift comparison across different models for varying activation of parameters $p_1$ and $p_2$ with a batch size of 16 during inference. Best overall results in \textbf{bold}, second best \underline{underlined}.}
  \label{tab:landmark_err}
  \centering
  \setlength\tabcolsep{4pt}
  \begin{tabular}{c|ccccc}
    & \textbf{Comb} $\downarrow$ & \textbf{X} $\downarrow$  & \textbf{Y} $\downarrow$ & \abs{\textbf{X-Y}} $\downarrow$ & \textbf{Runtime/} \\
    \textbf{Model} &  & \textbf{Err} & \textbf{ Err} & \textbf{Shift} & \textbf{Sample (ms)}\\
    \midrule
    & \multicolumn{5}{c}{$p1 \neq 0, p2 = 0$}\\
    Sim. & 0.00 & 0.00 & 0.00 & 0.00 & 153.205 \\
    convAE & 11.93 & 6.75 & 8.13 & 1.38 & 0.110 \\
    VAE & 11.53 & 6.55 & 7.83 & 1.28 & 0.122 \\
    UNet & 2.82 & 1.28 & 2.15 & 0.87 & 0.118 \\
    Pix2Pix & \underline{2.00} & \underline{0.99} & \underline{1.43} & \textbf{0.44} & 0.122\\
    DDPM & \textbf{1.93} & \textbf{0.94} & \textbf{1.39} & \underline{0.45} & 3970.603\\
    SD(w.CA) & 3.09 & 1.59 & 2.21 &0.62 & 2991.678\\
    SD & 2.79 & 1.41 & 2.01 & 0.60 & 2997.576\\
    \midrule
    & \multicolumn{5}{c}{$p1 = 0, p2 \neq 0$} \\
    Sim. & 0.00 & 0.00 & 0.00 & 0.00 & 153.205 \\
    convAE & 10.56 & 8.35 & 4.77 & 2.21 & 0.110 \\
    VAE & 10.40 & 8.26 & 4.62 & 3.64 & 0.122 \\
    UNet & 2.36 & \underline{1.33} & 1.60 & 0.27 & 0.117 \\
    Pix2Pix & \textbf{1.77} & \textbf{1.02} & \textbf{1.14} & \textbf{0.13} & 0.123 \\
    DDPM & \underline{2.13} & 1.39 & \underline{1.23} & \underline{0.16} & 3970.603 \\
    SD(w.CA) & 2.85 & 1.60 & 1.94 & 0.34 & 2991.678 \\
    SD & 2.44 & 1.38 & 1.64 & 0.26 & 2997.576 \\
  \end{tabular}
\end{table}
\begin{figure}[h!]
    \centering
    \includegraphics[width=0.4\textwidth]{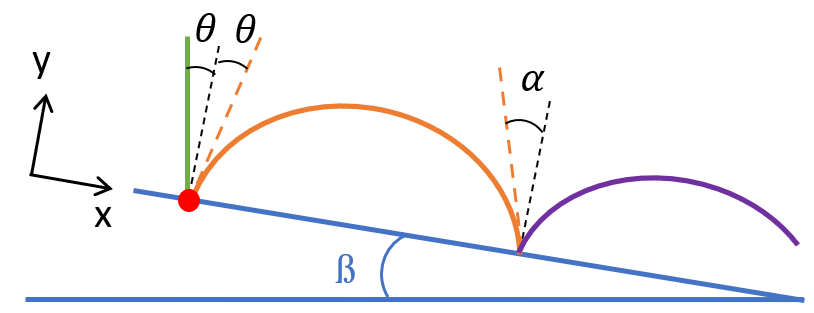}
    \vspace{-1em} 
    
    \begin{equation}
        \color{green}
        F_g = m \ast a \Leftrightarrow m \ast \ddot{y} = -m \ast g \Leftrightarrow \ddot{y} = -g
        \label{eq:BB_free_fall_main}
    \end{equation}
    \vspace{-0.8em} 
    \begin{equation}
        \color{red}
        m \ast \ddot{y} + d \ast \dot{y} - c \ast (r - y) = -m \ast g
        \label{eq:BB_ball_ground_bounce}
    \end{equation}
    \vspace{-0.6em} 
    \begin{equation}
        \color{orange}
        \begin{split}
            m \ast \ddot{x} = m \ast g \ast \sin(\beta) & \text{\:\:along x axis} \\
            m \ast \ddot{y} = -m \ast g \ast \cos(\beta) & \text{\:\:along y axis}
        \end{split}
        \label{eq:BB_oblique_throw}
    \end{equation}
    \caption{\small \label{fig:BB_movement_main}Physical equations describing the bouncing ball movement divided in 3 parts: free fall (green), impact \& bounce of the ball on the ground (red dot) and oblique throw with initial velocity from ball bounce (orange).}
\end{figure}
\\
\noindent\textbf{Dataset.}
We generated our data set with \textit{Pymunk}~\cite{Pymunk}, a Python physics simulation toolbox. For the bouncing case we use 50k pairs of training images (256x256px) and perform the evaluation of the generative networks with 1600 images. Three parameters on the input images are variable: the ground inclination, the start height of the ball movement and the ball position. Additionally, the time interval between the input and target images is also variable. All other physical properties of the problem (ball's mass or friction coefficients) are kept constant. A multi-coloured structure with randomly distributed lines is superimposed on the simulation images in order to obtain greater diversity in the training data and is also used to analyze consistency, e.g. how well the generative networks can map fine multicoloured patterns (see \cref{fig:ball_overview}). The total time required to create a sample using the available physics simulation CPU code (with a customized background) is $3.8$ seconds, while the inference times for the generative models remain consistent with those reported in \cref{tab:combined_metrics}.
\\
\noindent\textbf{Physics of bouncing balls.}
To model the physics behind the movement of a bouncing ball, the motion is divided into different sections which are represented with different colors in \cref{fig:BB_movement_main}. 
First, the ball is released from a defined height without any velocity and falls down according to the Newton's law of the free fall (green part) (see equation~\ref{eq:BB_free_fall_main}). The impact of the ball on the ground and its rebound can be described in simplified terms by a spring-damper model. The aim of this second part of the movement (red dot) is to calculate the rebound speed and angle. This is done with equation~\ref{eq:BB_ball_ground_bounce}, which contains the constant of the elements of the simplified model (see details in Appendix~\ref{a:BB_physics}) and the angle corresponds to the ground inclination for the first impact. Finally, the orange part of the movement is assumed to be an oblique throw with the initial velocity $v_0$ calculated with equation~\ref{eq:BB_ball_ground_bounce}. The equations of motion describing this oblique throw along the X and Y axes are also derived from the Newton's law (see equation~\ref{eq:BB_oblique_throw}). When the ball hits the ground for the second time, the principle "angle of incidence = angle of reflection" is used to determine the angle of rebound $\theta$ and the next part of the motion is again defined by equations~\ref{eq:BB_ball_ground_bounce} and~\ref{eq:BB_oblique_throw}.\\
\noindent\textbf{Evaluation.}
The generated images are examined according to 4 precise error criteria: the ball position split in X and Y directions, where the error is defined as the number of pixels between the theoretical and the predicted ball center. The second error measure is the ball rotation describing its angle in degrees during the movement due to the inclined ground surface. The next quality indicator of the algorithms is the out-of-roundness of the ball also measured in pixels by taking all identified points of the ball contour and calculating the difference between the theoretical ball radius of 15px and the predicted one. The standard deviation of all radius errors is taken as the error measure. Finally, the correct representation of the ground inclination is checked: the angular error between simulation and prediction is also measured in degrees.
In addition, there are three "general" error criteria that give preciser indications of the correct representation of the physics which are explained with the complete results in the Appendix~\ref{a:ball}.\\
The results of the seven examined generative algorithms are summarized in table~\ref{tab:BB_MainResults} (only for the four most important error criteria). The last column of the table contains the average number of non-evaluable predicted images over the four error criteria. A detailed analysis of the advantages and disadvantages of the individual methods with all the evaluation parameters can be found in the Appendix~\ref{a:ball}, as well as the evaluations obtained for individual runs for the rolling and bouncing case.
\setlength{\textfloatsep}{6pt}
\begin{table}[h!]
\footnotesize
    \centering
    \setlength\tabcolsep{3pt}
    \begin{tabular}{c|c|c|c|c|c}
    & \textbf{Position X} & \textbf{Position Y} & \textbf{Rotation} & \textbf{Roundness} & \textbf{Error} \\
    \hline
    convAE & $4.24 \pm 3.9 $ & $6.08 \pm 5.9 $ & $12.2 \pm 8.6 $ & $1.06 \pm 0.0 $ & 99\% \\
    VAE & $4.69 \pm 6.1 $ & $6.25 \pm 6.9 $ & $31.0 \pm 40 $ & $0.90 \pm 0.1 $ & 95\% \\
    UNet & $ 5.53\pm 7.5 $ & $10.8 \pm 12 $ & $15.2 \pm 23 $ & $0.74 \pm 0.2 $ & 28\% \\
    Pix2Pix & $ \textbf{6.28} \boldsymbol{\pm} \textbf{8.0} $ & $\textbf{11.7} \boldsymbol{\pm} \textbf{13} $ & $\textbf{17.2} \boldsymbol{\pm} \textbf{21} $ & $\textbf{0.56} \boldsymbol{\pm} \textbf{0.1}$ & 11\% \\
    DDPM & $ 7.91\pm 9.0 $ & $15.5 \pm 14 $ & $32.9 \pm 34$ & $0.61 \pm 0.2 $ & 5.7\% \\
    SD(w.CA) & $40.0 \pm 49 $ & $24.8 \pm 23 $ & $61.1 \pm 52 $ & $ \textbf{0.53} \boldsymbol{\pm} \textbf{0.2} $ & 7.3\% \\
    SD & $8.55 \pm 12 $ & $16.2 \pm 14 $ & $34.2 \pm 38 $ & $ \textbf{0.47} \boldsymbol{\pm} \textbf{0.1} $ & $ \textbf{2}\boldsymbol{\%} $\\
    
    \end{tabular}
    \caption{\small Prediction results for the bouncing ball experiment in mean$\pm$std and mean percentage of non-evaluable predictions for the four criteria (best overall results in bold).Average runtime of the simulation: 3.8s, generative models: same as in\cref{tab:combined_metrics}  }
    \label{tab:BB_MainResults}
\end{table}
\setlength{\textfloatsep}{6pt}
\\
Overall, the GAN (Pix2Pix model) provides the most stable representation of the bouncing ball problem. It achieves the best results across all evaluation criteria, with the most reliable physics predictions and fewest failed samples. 
The UNet also represents the ball position and angle on the generated samples very well, but produces more error images that cannot be evaluated with mostly either no ball or only a shredded shape. The greatest weakness of this algorithm lies in the clean representation of the ball roundness. The auto-encoder approaches (VAE and convAE) produce very blurred images in which the ball is often distorted and hardly recognizable. Therefore, the mean error values are not very representative, as over 90\% of the predictions cannot be analyzed. 
\\The denoising diffusion approach (DDPM) produces slightly poorer results in absolute terms. Especially the prediction of the ball rotation is worse, with an error approximately twice as high compared to the two other approaches.
\\Finally, to test another type of diffusion, two variations of stable diffusion were compared. The one with cross-attention is generating more error pictures and show high error values. In contrast, the normal stable diffusion performs very well: the mapping of the position is almost as accurate as with the Pix2Pix approach and the representation of the ball roundness is the best. In addition, the network generates very few error images, so that it could almost be placed on the same level as the GAN but comes as second due to a poorer angle mapping.
\\
As can be seen in the qualitative result images below (figure~\ref{fig:BB_PredImgs_1}), GAN and diffusion (especially the SD) networks depict small details much more accurately. The background structure of the images with the different colors is completely blurred by the UNet but not by the GAN and the diffusion. The DDPM predictions are blurred because the images have been scaled up from 64px to 256px. If the output of the diffusion model is set to 256px, the target image is sharp again but several or no balls and many more artefacts appear on the net predictions (see figure~\ref{fig:Err_DiffTraining_highRes} in the Appendix). To summarize, the Pix2Pix and Stable Diffusion approaches also are visually performing the best.
\\
Finally, slight correlations can be observed between the simulation's parameters and the generated errors. For example the ball position in the X direction is predicted worse for all nets when the time interval between input and prediction images becomes larger. More details on the correlations of each method can be found in the Appendix~\ref{a:BB_eval_results}.\\
\noindent\textbf{Analysis.} When comparing the result in table~\ref{tab:BB_MainResults} (and tables \ref{tab:BB_Results_SumTable},~\ref{tab:RB_Results_SumTable} in the Appendix) with the physical motion equations, it shows that the errors of the generative networks increase with higher order terms. The second-order terms, such as the movement along the X-axis and the ball angle represent the largest errors for the rolling scenario. In the bouncing case, a second-order derivative is also present to describe the ball movement in the Y-direction due to the ball’s impact on the ground. Therefore the calculation rule along the Y-direction consists of zero, first and second derivative terms, making it more complex than the equation in the X-direction. This is reflected in the error values of all the investigated approaches: the Y-error clearly exceeds that in the X-direction, as indicated by the complexity of the motion equations, although the X-position and angular errors remain roughly the same between the rolling and bouncing cases.
In general, it can be concluded that the errors of the analyzed  approaches increase with the order and complexity of the physical formulation.

%% file: sections/05_discussion.tex
\section{Discussion}
With \textit{PhysicsGen}, we present a comprehensive benchmark for the estimation of complex physical relations by image-to-image generative models. The three different datasets contained in \textit{PhysicsGen} cover a wide range of diverse simulation scenarios and hence allow a systematic evaluation of learning based image generation models.

\noindent\textbf{Key insights from the baseline-evaluation.}
Despite the diversity of the three given physical problems and the fundamentally different network types used the base-line evaluation experiments, we can report several consistent findings:
\begin{itemize}
\item Due to the unavailability of optimized GPU code for the used simulations it is impossible to provide a theoretically sound runtime comparison. Since a highly parallel implementation of the simulation algorithms is highly non trivial (if possible at all), our analysis focuses on the runtime analysis of the generative models. However, replacing physical simulations with GPU powered generative models allows \textbf{high computational speed-ups in practice}, especially when replacing highly iterative solvers like in the urban sound propagation with Pix2Pix or UNets (where we report speed-up factors of up to 20k).
\item In terms of physical correctness and consistency, our evaluation shows \textbf{good results for simple 0th and 1st order problems}, but also indicates that generative models mostly \textbf{fail to predict more complex physical relations} (which are typically formulated by higher order terms). This finding needs further investigation whether this is an effect of the principal theoretical limitations of current models or just a lack of sufficient training data.  
\item We also observe \textbf{distinct error pattern for different model types} which represents how the different model types are dealing with complexity and uncertainty. Please refer to Appendix \ref{a:BB_eval_results} for examples in the following exemplary discussion for the bouncing ball problem (similar effects can be observed in Appendix~\ref{app:results} for the sound propagation tasks as well): while GANs often produce physically possible outputs, they still can fail to generate the correct output for given inputs. This results in consistent scenes with a complete ball at the wrong position. UNets, VAE and convAE on the other hand, often fail to produce consistent scenes and tend to blur uncertain predictions, resulting in deformed balls and destructed backgrounds. Lastly, diffusion models often produce multiple candidate solutions (multiple balls in one scene) and tend to fail to estimate more complex relations like ball rotation. This phenomenon also needs further investigation.
\end{itemize}
\noindent\textbf{Limitations.}
The provided run-time analysis is based on the comparison of practically available implementations and is not fair from a theoretical perspective as no efforts were made to optimize the simulation codes to GPUs. Also, even though we carefully selected the three proposed physical benchmark problems such that they are likely to transfer to a wide range of other simulation tasks, our evaluation provides only a small subset of possible applications of generative models in physical modeling. Hence, results from our benchmark might not transfer to all possible scenarios. \\
\noindent\textbf{Broader Impact Statement.}
The simulation of physical systems forms the backbone for a wide range of scientific and engineering tasks. Hence, it is important to have benchmarks for the emerging field of neural enhanced simulations to compare and validate these novel approaches. This work provides such a benchmark for the specific, but widely applicable case, where physical simulations are used to map between 2D input and output structures (visualized as images). Our analysis shows that while current generative models are promising significant speedups, there are still many open questions regarding their physical correctness.  

%% file: appendix/model_architecture.tex
\section{Training Setup}
\label{app:gen_models}

This appendix provides a detailed overview of the architecture, input specifications, hyperparameters, and computer resources used for the generative models utilized in this research. Each experiment was conducted on a workstation equipped with a single NVIDIA RTX 4090 GPU (24 GB VRAM) unless specified otherwise.

The U-Net model architecture, adopted from \cite{unet}, and the Pix2Pix setup, based on \cite{pix2pix}, are designed to process either grayscale or RGB image inputs. When necessary, the input is extended by appending additional parameters as a separate dimension.
We followed the methodology described in \cite{ddpm} for the Diffusion model while incorporating conditional inputs as separate dimensions alongside the noised input image. Each model was constructed upon a unified U-Net backbone, scaling from 64 to 1028 channels and reconverging to 64, ensuring consistency in model design across all datasets. During training, different loss functions were employed to suit each model's architecture: Mean Squared Error (MSE) loss for the U-Net and Diffusion models, and a combination of Binary Cross Entropy (BCE) loss and L1 loss for the GAN. These architectures were applied consistently across all tasks introduced in this paper, with each task trained and evaluated independently.
The ConvAE and VAE \cite{kingma2022autoencodingvariationalbayes} models are both based on the U-Net architecture, similar to Pix2Pix and the standard U-Net, but adapted for their respective tasks. The ConvAE uses the U-Net structure with skip connections disabled. The VAE employs a similar hierarchical design.

The Stable Diffusion models were trained following the training procedure outlined in \cite{rombach2022highresolutionimagesynthesislatent}, with variations in conditioning mechanisms. In the standard Stable Diffusion (SD) setup, the conditioned image and parameters were passed to the denoising model via cross-attention and as additional dimensions with the noised input image. In contrast, the Stable Diffusion with Cross Attention Only (SD wCA) model exclusively employed cross-attention for passing the conditioned image and parameters.
The DDBMs were trained following the exact training procedure outlined in \cite{zhou2023denoisingdiffusionbridgemodels}, ensuring adherence to the methods described in the original work. DDBMs were trained on four NVIDIA A100 GPUs (80 GB VRAM each) to accommodate the increased computational demand, while Stable Diffusion was trained on a single A100 GPU. All other models, including U-Net, Pix2Pix, and standard Diffusion models, were trained on a single NVIDIA RTX 4090 GPU.

\begin{table}
\centering
\caption{UNet Training Hyperparameter}
\label{tab:hyp_unet}
\begin{tabular}{l|l}
\hline
\textbf{Hyperparameter} & \textbf{Value} \\
\hline
Batchsize & 18 \\
Learning Rate & $1 \times 10^{-4}$ \\
Epochs & 50 \\
Optimizer & Adam \\
Adam Betas & (0.5, 0.999) \\
\hline
\end{tabular}
\end{table}

\begin{table}
\centering
\caption{Pix2Pix Training Hyperparameter}
\label{tab:hyp_gan}
\begin{tabular}{l|l}
\hline
\textbf{Hyperparameter} & \textbf{Value} \\
\hline
Batchsize & 18 \\
Learning Rate Discriminator & $1 \times 10^{-4}$ \\
Learning Rate Generator & $2 \times 10^{-4}$ \\
Epochs & 50 \\
L1 Lambda & 100 \\
Lambda GP & 10 \\
Optimizer & Adam \\
Adam Betas & (0.5, 0.999) \\
\hline
\end{tabular}
\end{table}

\begin{table}
\centering
\caption{DDPM Training Hyperparameter}
\label{tab:hyp_dif}
\begin{tabular}{l|l}
\hline
\textbf{Hyperparameter} & \textbf{Value} \\
\hline
Batchsize & 18 \\
Learning Rate & $1 \times 10^{-4}$ \\
Epochs & 50 \\
Noise Steps & 1000 \\
Optimizer & Adam \\
Adam Betas & (0.5, 0.999) \\
\hline
\end{tabular}
\end{table}

\begin{table}[h!]
\centering
\caption{StableDiffusion Training Hyperparameters}
\begin{tabular}{l|l}
\hline
\textbf{Hyperparameter}                      & \textbf{Value}              \\ \hline
Number of Timesteps                     & 1000                      \\ 
Beta Start                              & 0.0015                    \\ 
Beta End                                & 0.0195                    \\ \hline
Down Channels                           & {[}256, 384, 512, 768{]} \\ 
Mid Channels                            & {[}768, 512{]}           \\ 
Number of Heads                         & 16                        \\ \hline
LDM Batch Size                          & 16                        \\ 
Autoencoder Batch Size                  & 4                         \\ 
Discriminator Start                     & 5000                      \\ 
LDM Learning Rate                       & $1 \times 10^{-4}$                    \\ 
Autoencoder Learning Rate               & $1 \times 10^{-5}$                   \\ 
Codebook Weight                         & 1                         \\ 
Commitment Beta                         & 0.2                       \\ 
Perceptual Weight                       & 1                         \\ 
KL Weight                               & 0.000005                  \\ 
LDM Epochs                              & 100                       \\ \hline
\end{tabular}
\label{tab:training_params}
\end{table}

\begin{table}[h!]
\centering
\caption{DDBM Training Hyperparameters}
\begin{tabular}{l|l}
\hline
\textbf{Hyperparameter}                      & \textbf{Value}              \\ \hline
$\sigma_{\text{max}}$                   & 80.0                      \\ 
$\sigma_{\text{min}}$                   & 0.002                     \\ 
$\sigma_{\text{data}}$                  & 0.5                       \\ 
Covariance ($C_{XY}$)                   & 0                         \\ 
Number of Channels                      & 256                       \\ 
Attention Levels                        & {[}32, 16, 8{]}           \\ 
Number of Residual Blocks               & 2                         \\ 
Sampler                                 & real-uniform    \\ 
Attention Type                          & flash            \\ 
Learning Rate                           & 0.0001                    \\ 
Dropout                                 & 0.1                       \\ 
EMA Rate                                & 0.9999                    \\ 
Number of Head Channels                 & 64                        \\ \hline
\end{tabular}
\label{tab:ddbm_params}
\end{table}

%% file: appendix/sound_appendix.tex
\clearpage
\section{Sound Propagation Appendix}
\label{a:sound}

\subsection{Evaluation Metrics} \label{app:sound_eval_metrics}

This appendix section details the evaluation metrics used, focusing on the Weighted Mean Absolute Percentage Error (wMAPE) and the implementation of ray tracing to determine line-of-sight (LoS) conditions between a sound source and various points on a sound propagation map.

\noindent\textbf{Weighted Mean Absolute Percentage Error:} The calculation of the wMAPE is adjusted to address cases where the true value of the noise map is zero but the predicted value is non-zero. In such instances, the script sets the error for these specific pixels to 100\%. This method accurately quantifies the total error, especially when the model predicts sound in areas where sound waves could not realistically reach according to the true data.

\noindent\textbf{Line of Sight:} Ray tracing is employed to establish LoS conditions between a sound source and various points on a sound propagation map. The process initiates by establishing a grid that represents the mapped area, with the sound source positioned at the center. For each point on this grid, a direct line is drawn from the source to the point. The script then checks for any obstructions along this line.

The ray-tracing function progresses by incrementally moving along the line from the source to the target point. It checks if any part of the line intersects with obstacles, represented by zero values on a binary image map. If an obstruction is encountered before the line reaches the target point, that point is marked as not having a line of sight to the source; otherwise, it is considered to have a line of sight.

\subsection{Location Sampling} \label{app:location_sampling}
\begin{figure}
\centering
\includegraphics[width=0.65\linewidth]{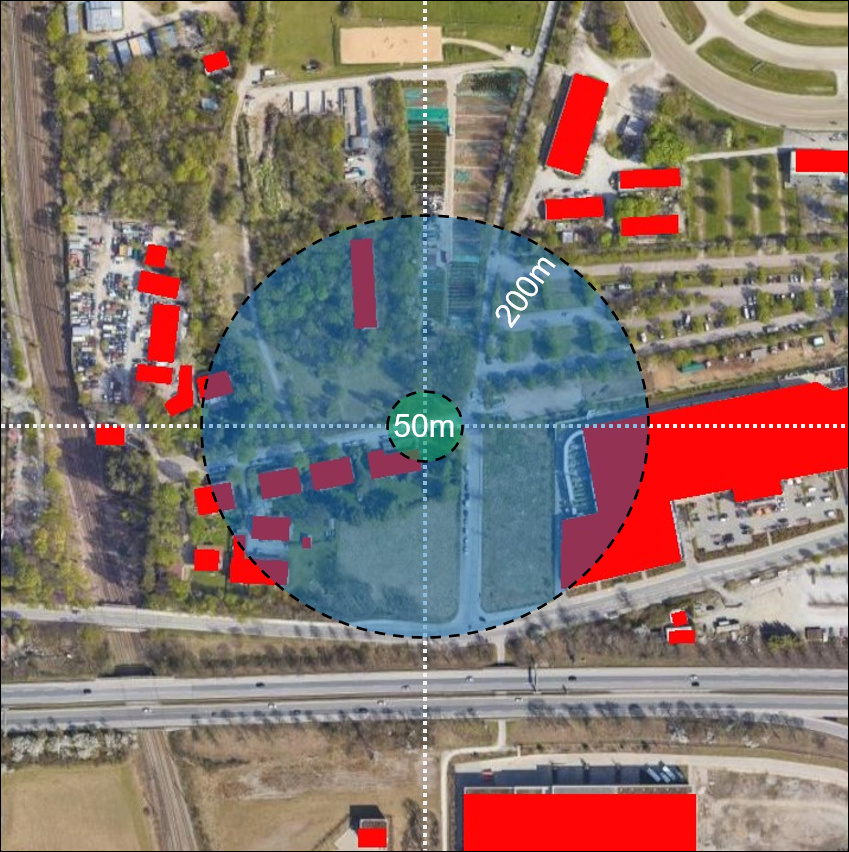}
\caption{\small Satellite image of a sampled location for sound propagation studies. Buildings are marked in red, illustrating the distribution within the area. The blue circle represents the 200-meter radius within which at least ten buildings are required, while the green circle indicates the 50-meter radius that must be free of any buildings to ensure a clear propagation path from the sound source.}
\label{fig:location_sampling}
\end{figure}

In order to conduct sound propagation studies that are reflective of diverse urban environments, our location sampling was designed with specific criteria to ensure a balanced and comprehensive dataset. The location sampling methodology for sound propagation studies required each selected location to contain at least ten buildings within a 200-meter radius of the designated sound source. Additionally, to avoid anomalous acoustic results and to simulate a realistic setting where sound sources are typically not positioned directly against structures, no buildings were allowed within a 50-meter radius of the sound source. Figure~\ref{fig:location_sampling} illustrates the urban areas selected for this study, showcasing the distribution of buildings relative to the sound source in the center.

Locations were randomly sampled across 5 cities, providing a broad geographical spread and a variety of urban layouts. The locations sampled include: Hamburg, Hannover, Augsburg, Bonn and Munich.

For each city, 5,000 unique locations were selected, contributing to a total of 25,000 data points for the study. The dataset was then divided into training, evaluation, and testing sets with a split of 80\%, 15\%, and 5\%.

\subsection{Receiver Placement} \label{app:receiver_placement}

The receiver placement for our sound propagation simulations is a critical component that directly influences the accuracy and relevance of our results. To achieve an optimal setup, we utilized the NoiseModelling framework~\cite{noisemodelling_framework}. Our specific placement strategy required a combination of precision and broad coverage, which was not fully supported by any single existing script within the framework. Therefore, we integrated two available scripts and further developed a custom WPS (Web Processing Service) to meet our specific needs.

The first script we used was \textbf{Regular\_Grid}, which calculates a regular grid of receivers. This script uses a single geometry or a table of geometries to generate receivers evenly spaced by a specified distance (\textit{delta}) on the Cartesian plane, measured in meters. This method ensures a systematic and uniform coverage across the studied area, providing a comprehensive baseline for sound propagation assessments.

\begin{figure}[h]
\centering
\begin{subfigure}[b]{0.3\linewidth}
    \includegraphics[width=\linewidth]{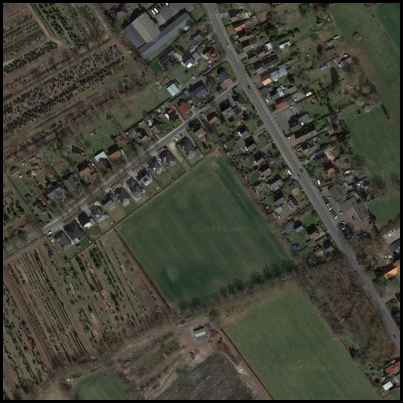}
    \caption{\small Sattelite image}
    \label{fig:imageA}
\end{subfigure}
\hfill
\begin{subfigure}[b]{0.3\linewidth}
    \includegraphics[width=\linewidth]{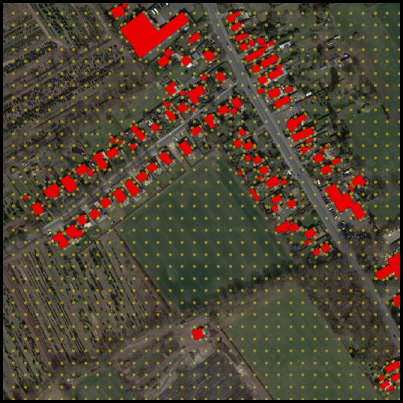}
    \caption{\small Receiver grid}
    \label{fig:imageB}
\end{subfigure}
\hfill
\begin{subfigure}[b]{0.3\linewidth}
    \includegraphics[width=\linewidth]{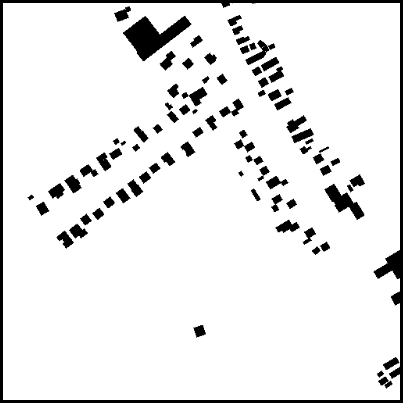}
    \caption{\small Urban layout}
    \label{fig:imageC}
\end{subfigure}
\caption{\small Starting with the selection of a $500\,\mathrm{m}^2$ area (a), buildings are identified, followed by placing a receiver grid (b). The urban layout (c) is then used for creating samples for the dataset.}
\label{fig:app_overview}
\end{figure}

The second script, \textbf{Building\_Grid}, is designed to place receivers specifically around building facades. This script generates receiver points approximately 2 meters from the building facades at a given height, facilitating detailed analysis of sound interactions with building surfaces, which are critical for urban acoustic modeling.

By combining these two approaches, we crafted a receiver placement strategy that not only maintains a uniform grid for broad coverage but also includes strategically placed receivers around building facades and edges. This hybrid approach ensures that our simulations accurately reflect sound propagation both across open areas and in close proximity to structural barriers, thereby enhancing the precision of the simulation results at key locations.

\subsection{Sound Physics} \label{app:sound_physics}
Following \cite{vorlander2007auralization}, for a discrete set of receivers $R$, the amplitude  $L^{j}_{{R_k}}$ of receiver $R_k$ at frequency $j$ is computed via iterative differences:
\begin{align}
\label{eq:noise}
L^{j}_{{R_k}} &= L^{j}_{W} - A_{div_{R_{k}}} - A^{j}_{atm_{R_k}} \notag\\
&\quad - A^{j}_{dif_{R_{k}}} - A^{j}_{grd_{R_{k}}} 
\end{align}
\textrm{where}
\begin{align*}
L^{j}_{W} &\textrm{models the source,}\\
A_{div_{R_{k}}} &\textrm{captures the geometrical spreading,}\\
A^{j}_{atm_{R_k}} &\textrm{represents the atmospheric absorption,}\\
A^{j}_{dif_{R_{k}}} &\textrm{models diffraction,}\\
A^{j}_{grd_{R_{k}}} &\textrm{the ground effect - which is neglected in our study.}
\end{align*}

Several environmental factors influence the sound level at a receiver:

\textbf{Geometrical Spreading:} This factor accounts for the dispersion of sound waves as they propagate through the medium. The decrease in sound intensity due to geometrical spreading is given by:
\begin{equation}
A_{div_{R_{k}}} = 20 \log_{10} (d_i) + 11
\end{equation}
where \(d_i\) is the distance between the sound source and the receiver.

\textbf{Atmospheric Absorption:} This factor represents the loss of sound energy as it travels through the atmosphere, influenced by the atmospheric conditions such as humidity and temperature. It is calculated as follows:
\begin{equation}
A^{j}_{atm_{R_k}} = \alpha_{air} \frac{d_i}{1000}
\end{equation}
where \(\alpha_{air}\) is the atmospheric absorption coefficient.

\textbf{Diffraction:} Sound waves can bend around obstacles, a phenomenon modeled by the diffraction factor:
\begin{equation}
A^{j}_{dif_{R_{k}}} = \begin{cases}
10\log_{10}(3+\frac{40}{\lambda}C''\delta) & \text{if } \frac{40}{\lambda}C''\delta \geq - 2 \\
0 & \text{otherwise}
\end{cases}
\end{equation}
where \(\lambda\) is the wavelength of the sound, \(C''\) is the diffraction coefficient, and \(\delta\) is the path length difference expressed in m between for direct and diffracted paths.

\textbf{Reflection Adjustments:} Reflections off various surfaces can significantly modify the sound level. This factor adjusts the sound power level based on the number of reflections and the properties of the reflective surfaces:
\begin{equation}
\label{eq:reflection}
L^{(n_{ref})}_{W} =  L^{(n_{ref} - 1)}_{{W}} + n_{ref} \times 10 \log_{10} (1 -{\alpha}_{vert})
\end{equation}
where \(n_{ref}\) is the number of reflections and \(\alpha_{vert}\) is the absorption coefficient of the reflecting surfaces.

\subsection{Runtime Analysis}\label{app:sound_runtime}
The runtime analysis, as detailed in Table~\ref{tab:model_simulation_runtime_comparison}, highlights the performance comparison between various models and simulation approaches for single sample processing.
\begin{table}
  \caption{\small Model vs. Simulation Performance Comparison for Single Sample Processing. The complex source is a single test sample for a more complex source with 28 descriptive sound signal sources for the simulation. This illustrates how the processing time increases significantly with more complex signal sources. It is important to note that this analysis may not provide a completely fair assessment from a theoretical perspective, as no efforts were made to optimize the simulation codes for GPU execution.}
  \label{tab:model_simulation_runtime_comparison}
  \centering
    \begin{tabular}{lc}
      \textbf{Model - Condition} & \textbf{Mean Runtime (ms)}\\
      \hline
      convAE & 0.128 \\
      UNet & 0.126 \\
      Pix2Pix & 0.138 \\
      DDPM & 3986.353\\
      SD(w.CA) & 2961.027\\
      SD & 2970.86 \\
      DDBM & 3732.21 \\
      \hline
      Simulation - Baseline & 20471.7 \\
      Simulation - Diffraction & 20602.7 \\
      Simulation - Reflection & 25097.3 \\
      Simulation - Combined & 29239.5 \\
     \hline
      Simulation - Combined & 186229.5 \\
      - 3rd Order Reflections &  \\
      Simulation - Combined & 540000\\
      - Complex Source
    \end{tabular}
\end{table}

\subsection{Scalable Simulation Pipeline \label{app:pipeline}}
\begin{figure}[h]
\begin{center}
\includegraphics[width=1\linewidth]{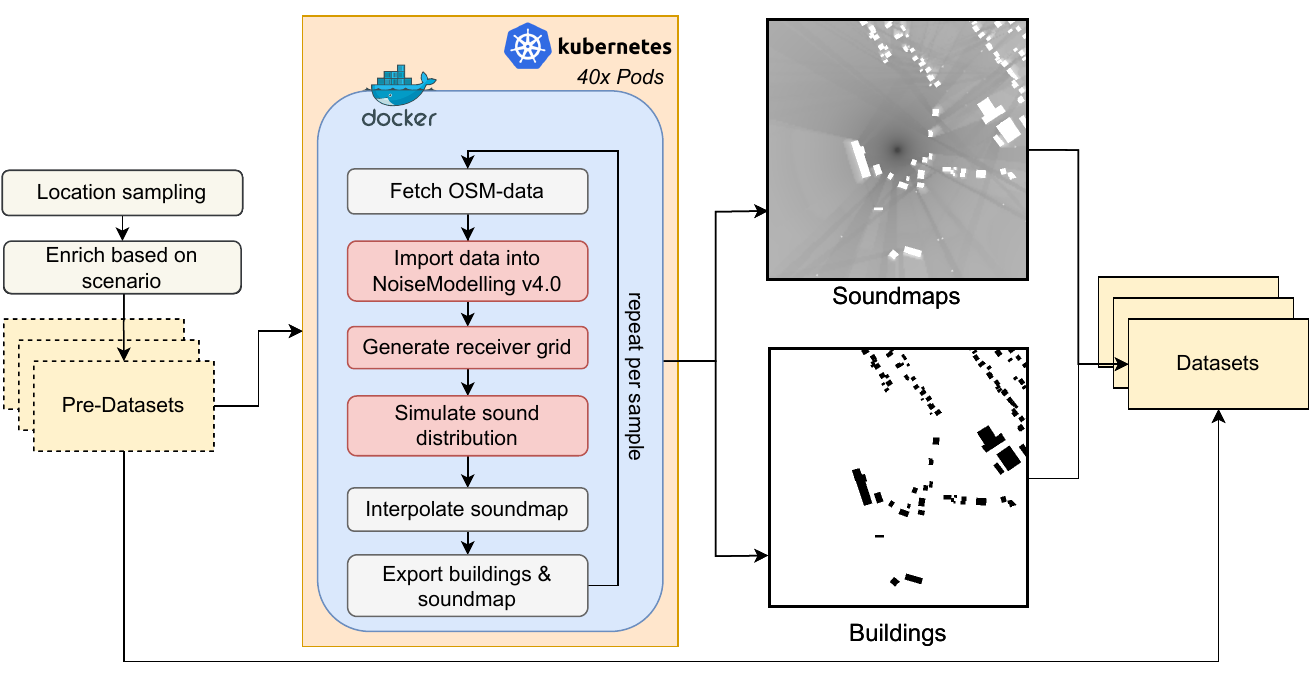}
\end{center}
\caption{\small Detailed visualization of the dataset generation pipeline.}
\label{fig:pipeline}
\end{figure}

The dataset generation pipeline visualized in Figure~\ref{fig:pipeline} is a crucial component of our study, developed to efficiently process sound propagation data in diverse urban settings. Utilizing the NoiseModelling framework~\cite{noisemodelling_framework}, we have automated the data input and simulation processes within a Docker-containerized environment. The pipeline commences with the automatic download of a 500m² area map from OpenStreetMap for each location using the Overpass API, followed by their import into the NoiseModelling framework alongside the signal source.

\begin{figure}
\centering
\begin{subfigure}[b]{1\linewidth}
    \includegraphics[width=\linewidth]{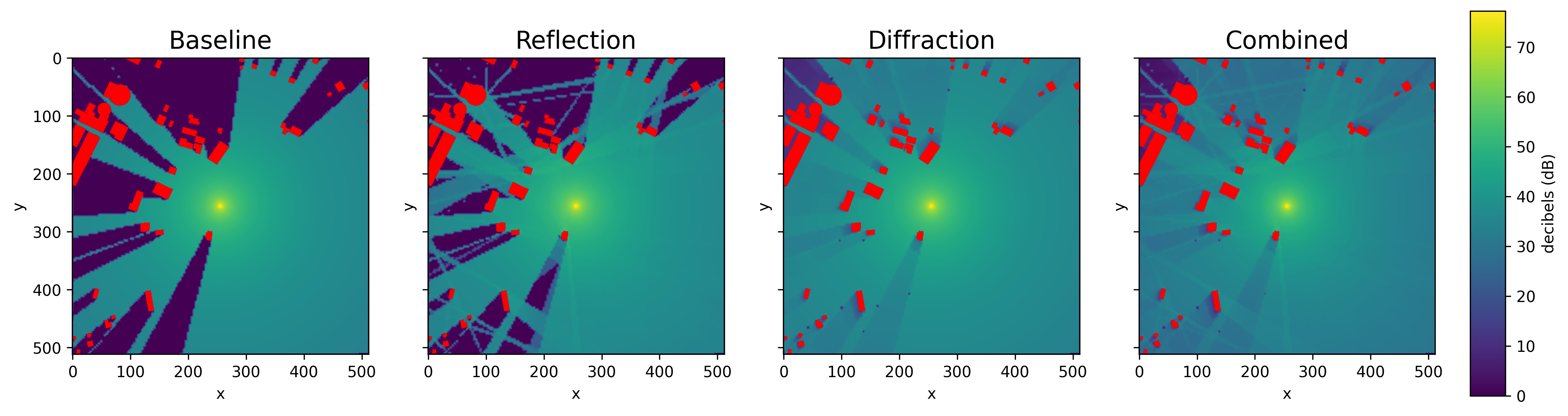}
    \label{fig:imageA}
\end{subfigure}
\begin{subfigure}[b]{1\linewidth}
    \includegraphics[width=\linewidth]{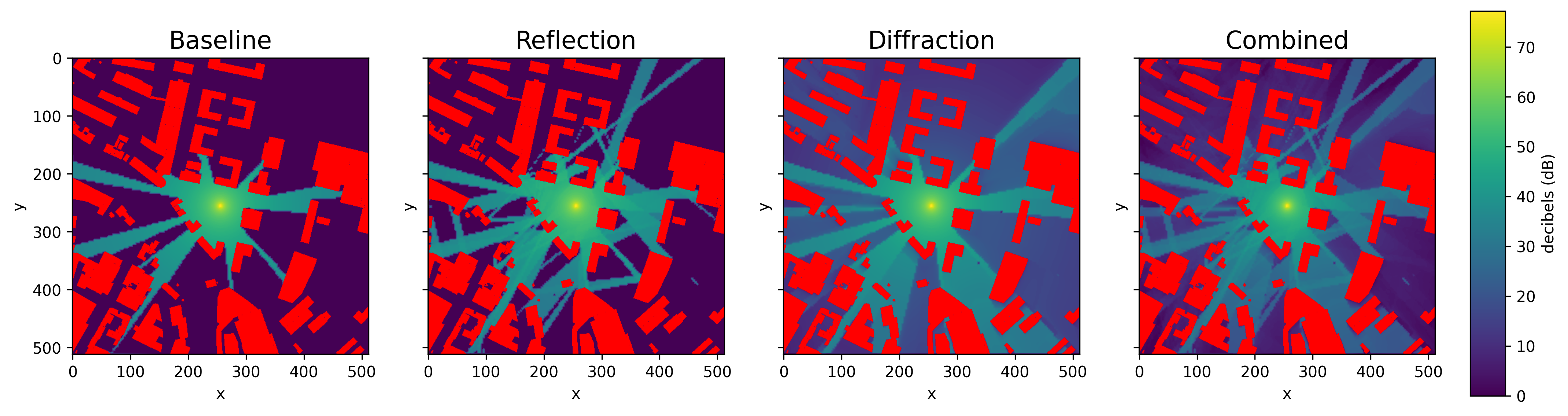}
    \label{fig:imageB}
\end{subfigure}
\caption{\small Comparison of true labels for \textbf{baseline}, \textbf{reflection}, \textbf{diffraction}, and \textbf{combined} tasks for two samples to visualize their differences.}
\end{figure}

\noindent Considering the computational intensity of this process, with an average duration of 30 seconds per sample, our pipeline is structured for scalability. It operates on a Kubernetes cluster with 40 pods, enabling us to complete the generation of the entire dataset, encompassing 25,000 data points for each complexity level, in approximately 20 hours.\\

\subsection{Additional Qualitative Results \label{app:results}}

\begin{figure}
\centering
\includegraphics[width=1\linewidth]{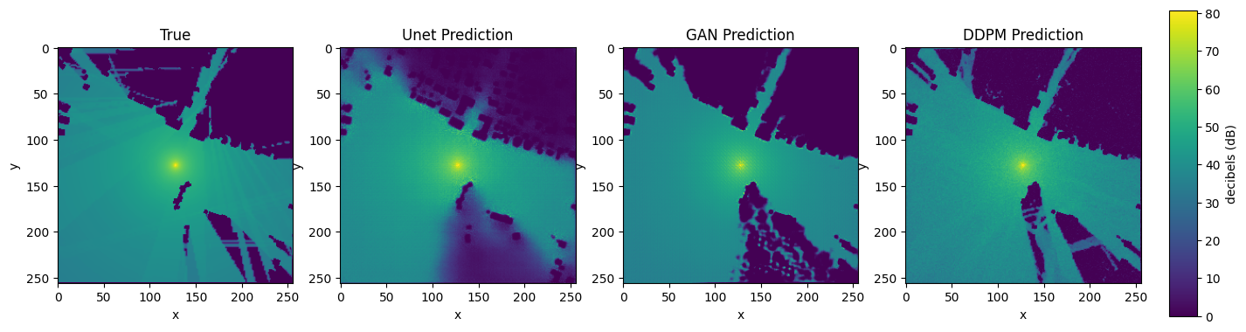}
\caption{\small Visualization of unique error patterns for Pix2Pix, UNet, and DDPM models in sound propagation simulations. Each model's approach to uncertain areas and its replication of reflection patterns are depicted.}
\label{fig:error_pattern}
\end{figure}

Figure \ref{fig:error_pattern} visualizes the error patterns for Pix2Pix, Unet and DDPM, illustrating behavior that is similar across all the architectures evaluated. The Unet model tends to blur areas where it is uncertain about the sound propagation results. Instead of accurately replicating the complex reflection patterns found in the dataset, the GAN attempts to fill these uncertain areas with a simplified, repetitive texture that does not match the true reflection dynamics.

In contrast, the diffusion model attempts a more direct replication of the reflection patterns visible in the training data. However, the error patterns observed with diffusion models, including DDPM, SD, and DDBMs, are highly similar. These models tend to overcompensate, introducing reflection patterns into areas where they are not physically plausible. This results in a noisy and sometimes unrealistic simulation output, particularly in regions where sound reflections should be minimal or absent based on the physical layout.

\begin{table}
\footnotesize
  \caption{\small Quantitative evaluation across all tasks for all architectures with a batch size of 16 during inference. Best overall results in \textbf{bold}, second best \underline{underlined}.}
  \label{tab:appendix_combined_metrics}
  \setlength\tabcolsep{4pt}
  \centering
  \begin{tabular}{llccccc}
    \textbf{Model} & \textbf{Task} & \multicolumn{2}{c}{\textbf{MAE} $\downarrow$} & \multicolumn{2}{c}{\textbf{wMAPE} $\downarrow$} & \textbf{Runtime/} $\downarrow$ \\
    & & \textbf{LoS} & \textbf{NLoS} & \textbf{LoS} & \textbf{NLoS} & \textbf{Sample (ms)}\\
      \hline
      Simulation & Base & 0.00 & 0.00 & 0.00 & 0.00 & 204700 \\
      convAE & Base & 3.67 & 2.74 & 20.24 & 67.13 & 0.128  \\ 
      VAE \cite{kingma2022autoencodingvariationalbayes} & Base & 3.92 & 2.84 & 21.33 & 75.58 & 0.124\\
      UNet \cite{unet} & Base & 2.29 & 1.73 & \underline{12.91} & \underline{37.57} & 0.138 \\
      Pix2Pix \cite{pix2pix} & Base & \underline{1.73} & \underline{1.19} & \textbf{9.36} & \textbf{6.75} & 0.138 \\
      DDPM \cite{ddpm} & Base & 2.42 & 3.26 & 15.57 & 51.08 & 3986.353 \\
      SD(w.CA) \cite{rombach2022highresolutionimagesynthesislatent} & Base & 3.76 & 3.34 & 17.42 & 35.18 & 2961.027\\
      SD & Base & 2.12 & \textbf{1.08} & 13.23 & 32.46 & 2970.86 \\
      DDBM \cite{zhou2023denoisingdiffusionbridgemodels}& Base & \textbf{1.61} & 2.17 & 17.50 & 65.24 & 3732.21 \\
      \hline
      Simulation & Dif. & 0.00 & 0.00 & 0.00 & 0.00 & 206000 \\
      convAE & Dif. & 3.59 & 8.04 & 13.77 & 32.09 & 0.128 \\
      VAE & Dif. & 3.92 & 8.22 & 14.46 & 32.57 & 0.124 \\
      UNet & Dif. & \underline{0.94} & \textbf{3.27} & \underline{4.22} & 22.36 & 0.138 \\
      Pix2Pix & Dif. & \textbf{0.91} & \underline{3.36} & \textbf{3.51} & \textbf{18.06} & 0.138\\
      DDPM & Dif. & 1.59 & \textbf{3.27} & 8.25 & \underline{20.30} & 3986.353  \\
      SD(w.CA) & Dif. & 2.46 & 7.72 & 10.14 & 31.23 & 2961.027 \\
      SD & Dif. & 1.33 & 5.07 & 8.15 & 24.45 & 2970.86 \\
      DDBM & Dif. & 1.35 & 3.35 & 11.22 & 23.56 & 3732.21 \\
      \hline
      Simulation & Refl. & 0.00 & 0.00 & 0.00 & 0.00 & 251000 \\
      convAE & Refl. & 3.83 & 6.56 & 20.67 & 93.54 & 0.128  \\
      VAE & Refl.  & 4.15 & 6.32 & 21.57 & 92.47 & 0.124 \\
      UNet & Refl.  & 2.29 & 5.72 & \underline{12.75} & 80.46 & 0.138\\
      Pix2Pix & Refl.  & \underline{2.14} & \textbf{4.79} & \textbf{11.30} & \textbf{30.67} & 0.138\\
      DDPM & Refl.  & 2.74 & 7.93 & 17.85 & 80.38 & 3986.353 \\
      SD(w.CA) & Refl.  & 3.81 & 6.82 & 19.78 & 81.61 & 2961.027 \\
      SD & Refl. & 2.53 & \underline{5.26} & 15.04 & \underline{55.27} & 2970.86 \\
      DDBM & Refl.  & \textbf{1.93} & 6.38 & 18.34 & 79.13 & 3732.21 \\
      \hline
      Simulation & Comb. & 0.00 & 0.00 & 0.00 & 0.00 & 251000 \\
      convAE & Comb. & 2.93 & 5.19 & 18.61 & 48.92 & 0.128 \\
      VAE & Comb.  & 3.08 & 5.35 & 19.59 & 50.24 & 0.124 \\
      UNet & Comb.  & 1.39 & \underline{2.63} & \underline{10.10} & 45.15 & 0.138\\
      Pix2Pix & Comb.  & \underline{1.37} & 2.67 & \textbf{9.80} & \underline{40.68} & 0.138\\
      DDPM & Comb.  & \textbf{1.26} & \textbf{2.21} & 13.07 & \textbf{40.38} & 3986.353 \\
      \end{tabular}
\end{table}

\begin{figure}
\centering
\includegraphics[width=\linewidth]{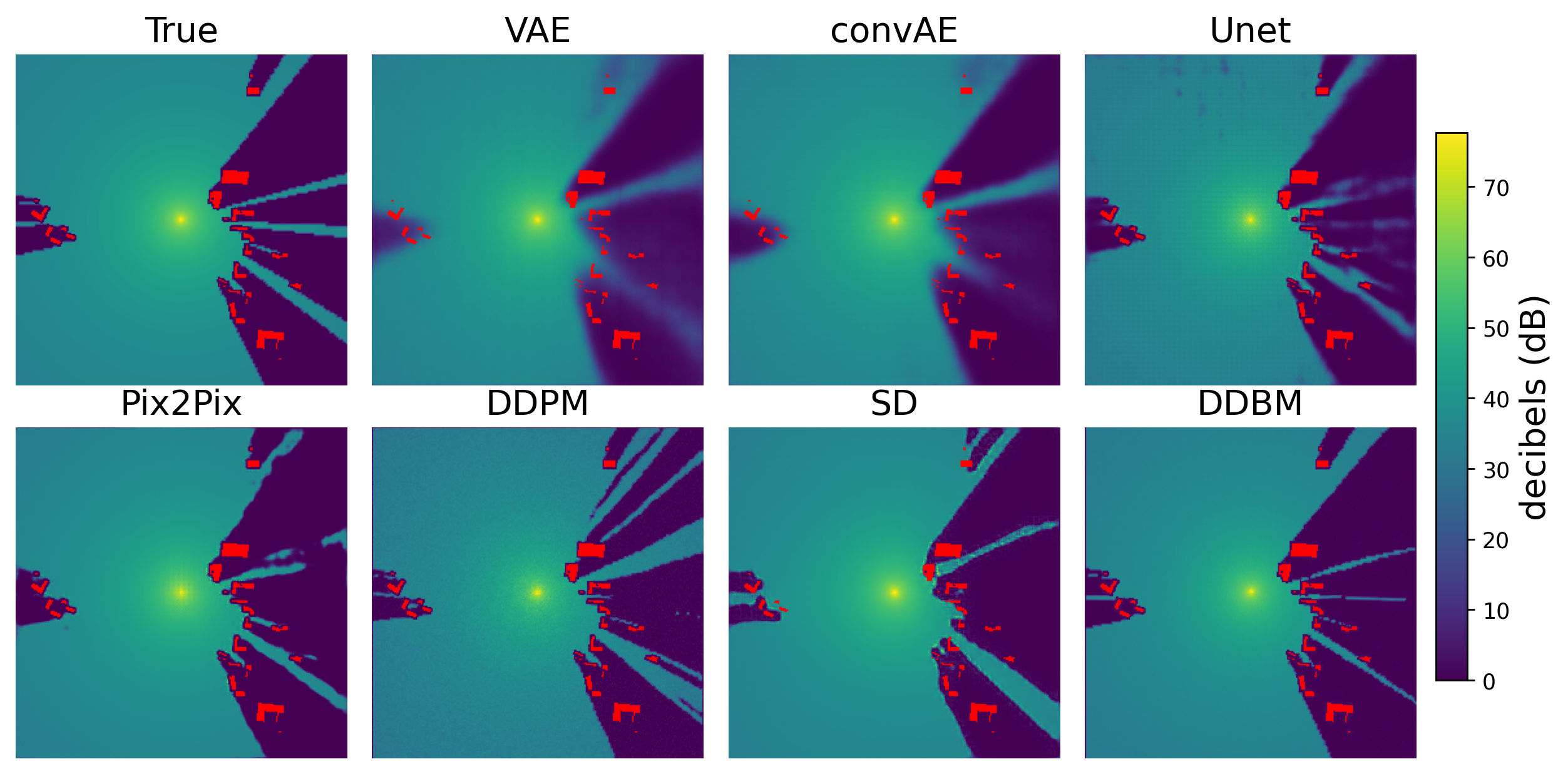}
\caption{\small Comparing the output of the physical simulation with the predictions for a single sample within the \textbf{baseline} task dataset.}

\end{figure}

\begin{figure}
\centering
\includegraphics[width=\linewidth]{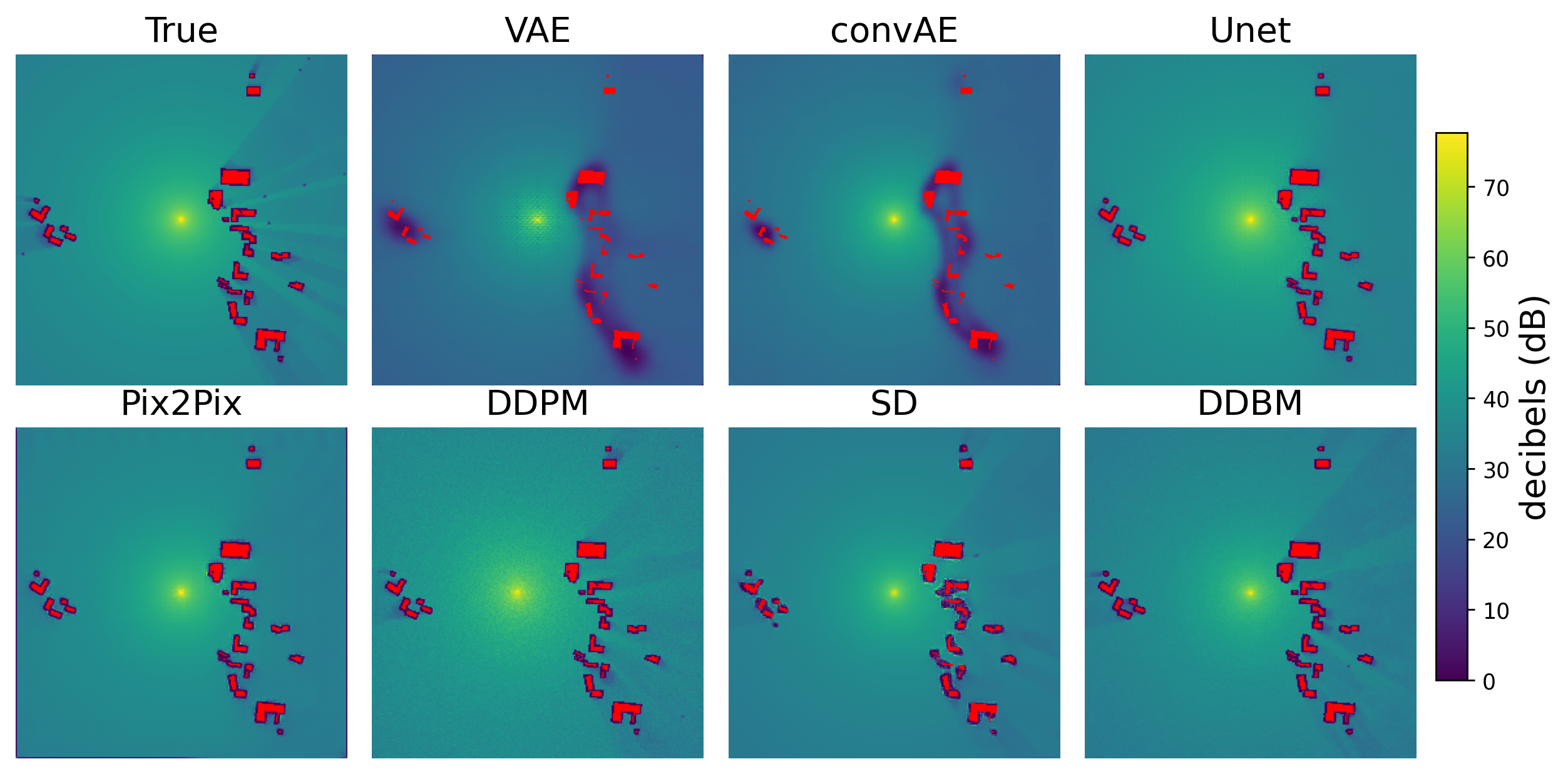}
\caption{\small Comparing the output of the physical simulation with the predictions for a single sample within the \textbf{diffraction} task dataset.}

\end{figure}

\begin{figure}
\centering
\includegraphics[width=\linewidth]{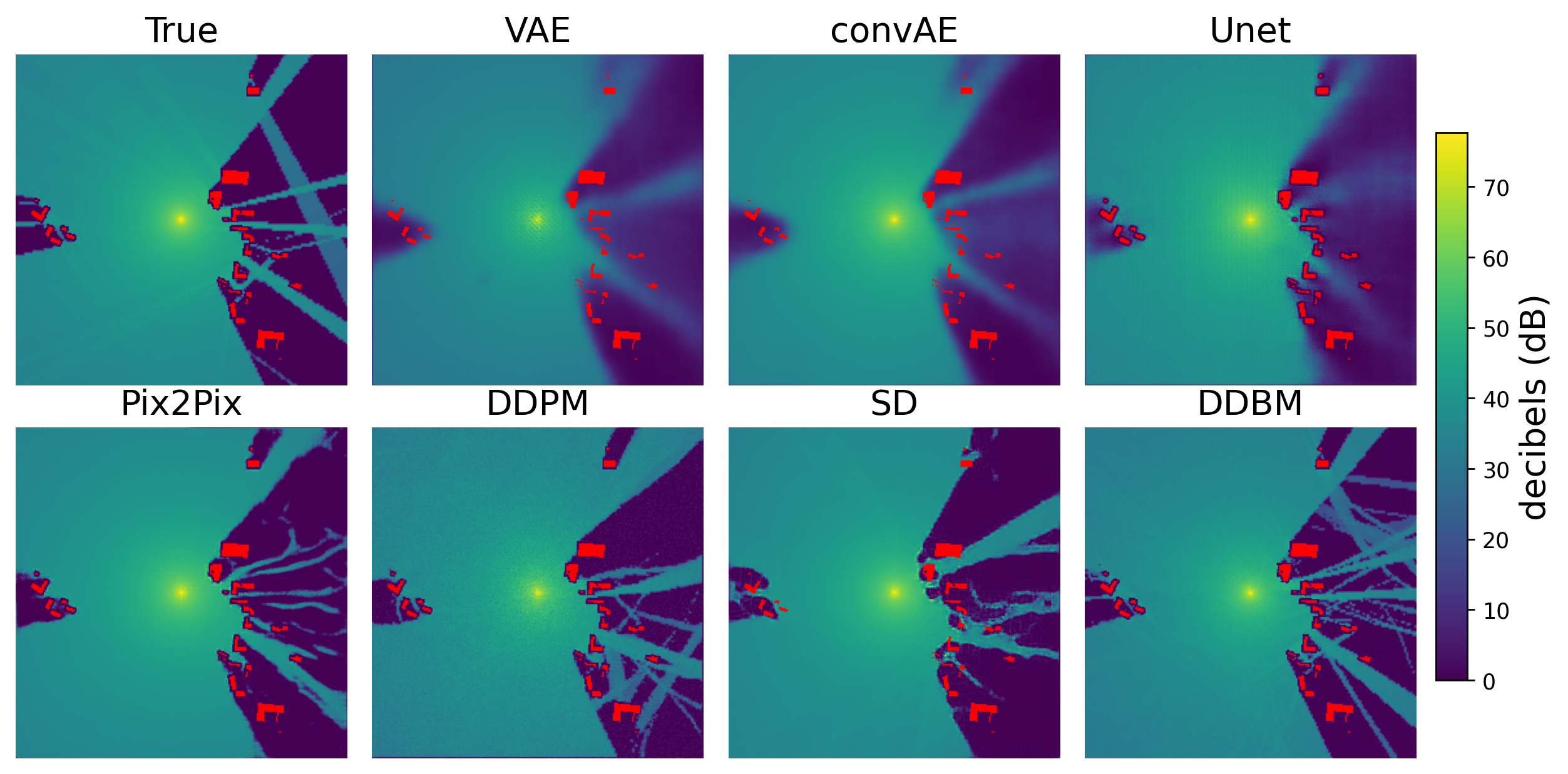}
\caption{\small Comparing the output of the physical simulation with the predictions for a single sample within the \textbf{reflection} task dataset.}

\end{figure}

\begin{figure}
\centering
\begin{subfigure}[b]{1\linewidth}
    \includegraphics[width=\linewidth]{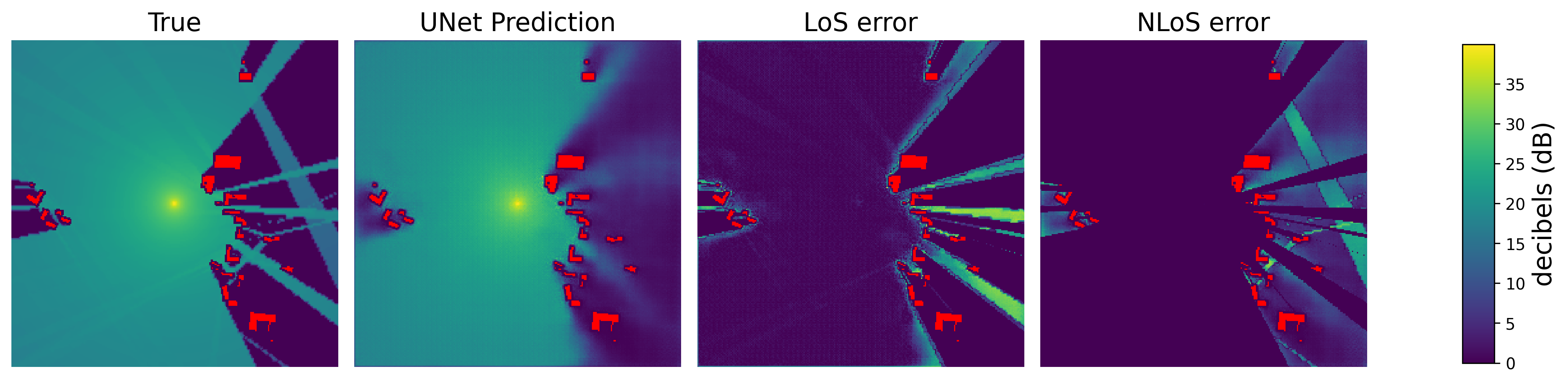}
    \caption{\small UNet}
    \label{fig:imageA}
\end{subfigure}
\vfill
\begin{subfigure}[b]{1\linewidth}
    \includegraphics[width=\linewidth]{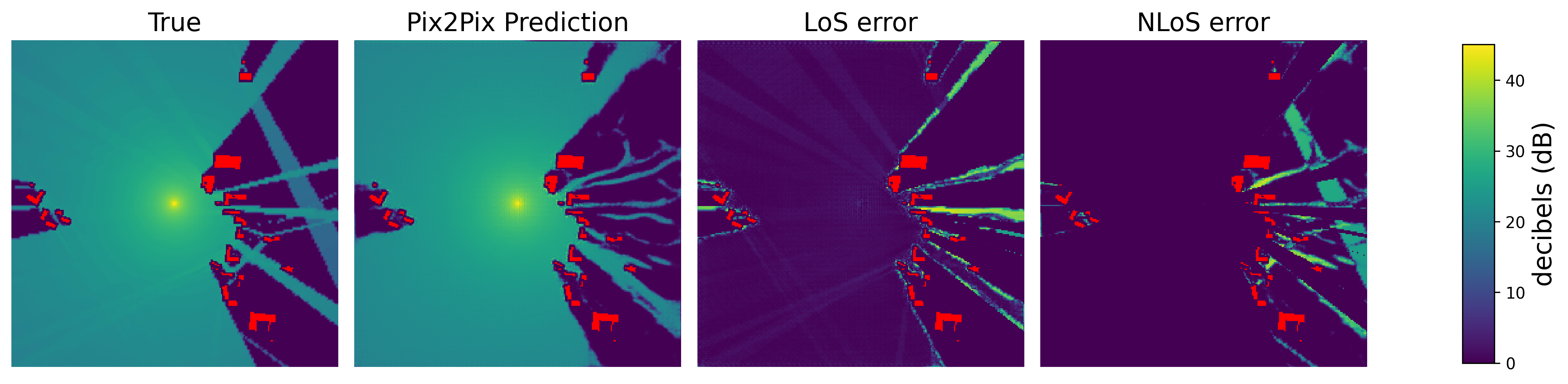}
    \caption{\small Pix2Pix}
    \label{fig:imageB}
\end{subfigure}
\vfill
\begin{subfigure}[b]{1\linewidth}
    \includegraphics[width=\linewidth]{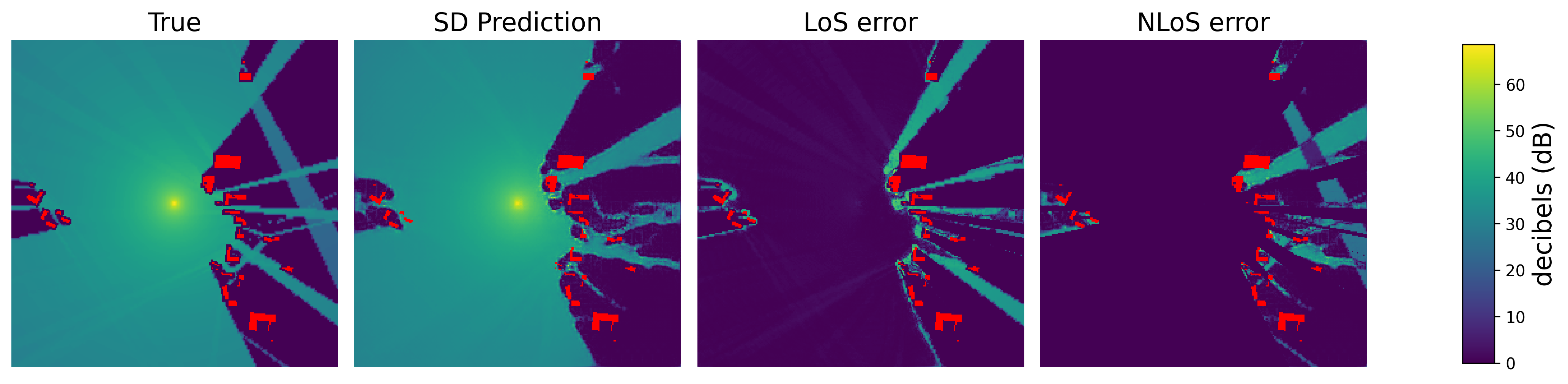}
    \caption{\small SD}
    \label{fig:imageC}
\end{subfigure}
\vfill
\begin{subfigure}[b]{1\linewidth}
    \includegraphics[width=\linewidth]{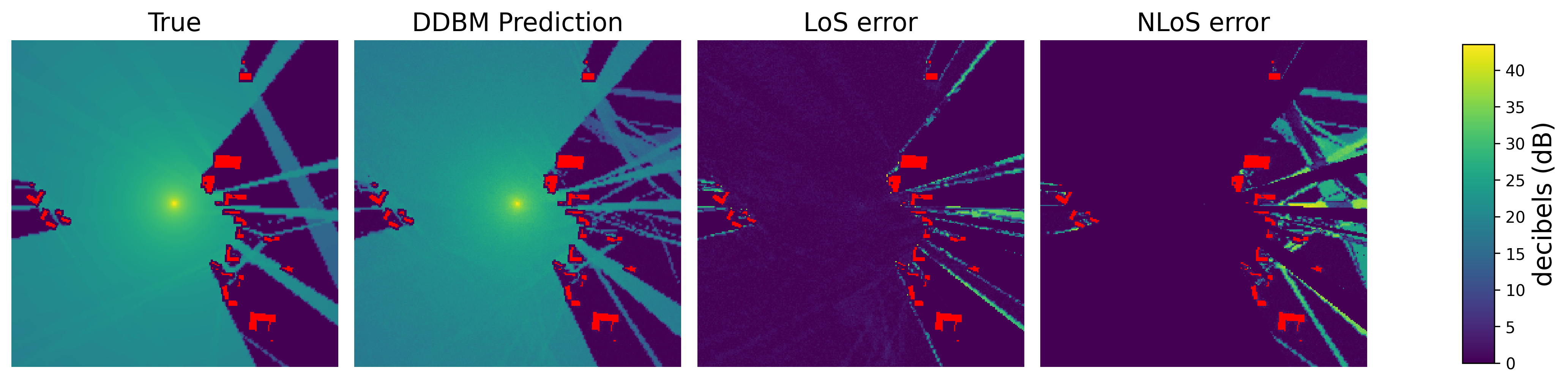}
    \caption{\small DDBM}
    \label{fig:imageD}
\end{subfigure}
\caption{\small Comparing the output of the physical simulation with the prediction of UNet (a), Pix2Pix (b), stable diffusion model (c) and DDBM (d) for a single sample within the \textbf{reflection} task dataset, distinguishing between the MAE in LoS and NLoS.}

\end{figure}

%% file: appendix/lens_appendix.tex
\clearpage
\section{Lens Appendix} \label{a:lens}

\subsection{Evaluation Metrics for Facial Landmark Detection} \label{app:face_landmark_metrics}

This appendix section details the evaluation metrics used to assess the accuracy of facial landmark detection, focusing on the Euclidean distance and mean absolute error calculations for both combined and separate x and y coordinates.

\noindent\textbf{Euclidean Distance:} The primary metric for evaluating the accuracy of facial landmark predictions is the Euclidean distance between predicted and true landmarks. This measure computes the root mean square error across all landmarks, providing a comprehensive gauge of overall prediction accuracy.

\noindent\textbf{Mean Absolute Error for X and Y Coordinates:} Additionally, the MAE for x and y coordinates is calculated separately. This breakdown allows for a detailed analysis of the model's performance along each axis, highlighting any directional biases or discrepancies in landmark prediction.

These metrics provide a dual approach to evaluating landmark detection performance—offering both a holistic measure via Euclidean distance and a directional sensitivity via separate MAE calculations.

\subsection{Lens Physics} \label{app:lens_physics}

The Brown-Conrady model is used in image processing to correct lens distortions by modeling both radial and tangential components~\cite{brownConrady}. While the paper focuses on the tangential aspects of distortion, here we provide a detailed mathematical formulation of the entire Brown-Conrady model, including both radial and tangential distortion corrections.

Radial distortion is typically dominant in optical systems and is characterized by its impact varying with the square of the distance from the optical center. This effect is modeled using three coefficients: $k_1$, $k_2$, and $k_3$, which adjust the coordinates as a function of radial distance $r^2 = x^2 + y^2$. The full expressions for the distorted coordinates incorporating both radial and tangential distortions are as follows:

\begin{align}
x_{dist} = x(1 + k_1 r^2 + k_2 r^4 + k_3 r^6) \\ + x + \left[2p_1xy + p_2\left(r^2 + 2x^2\right)\right], \\
y_{dist} = y(1 + k_1 r^2 + k_2 r^4 + k_3 r^6) \\ + y + \left[p_1\left(r^2 + 2y^2\right) + 2p_2xy\right].
\end{align}

These equations describe how each image point $(x, y)$ is displaced to $(x_{dist}, y_{dist})$ due to lens distortions.

\subsection{Lens Distortion Pipeline} \label{app:lens_distortion}

This appendix details the methodology for applying lens distortions to images within our dataset, using a computational pipeline built on Python and OpenCV~\cite{opencv_library}. The process leverages a pre-defined set of distortion parameters stored in a CSV file, including tangential coefficients, to systematically alter each image. The dataset consists of 100k images and was divided into training, evaluation, and testing sets with a split of 80\%, 15\%, and 5\%.

\textbf{Distortion Computation:} Each image is processed sequentially. The script calculates the distortion for each pixel using the specified parameters (radial coefficients \(k_1, k_2, k_3\) and tangential coefficients \(p_1, p_2\)) and the camera calibration data (focal lengths \(fx, fy\) and principal point coordinates \(cx, cy\)). These calculations transform the original pixel coordinates to their new distorted positions.

\textbf{Image Remapping:} Using OpenCV’s \texttt{remap} function, the distorted pixel coordinates are mapped back onto the original image, creating the visually distorted output. This process ensures that each pixel's new location reflects the simulated lens.

\subsection{Additional Qualitative Results} \label{app:lens_results}
This section provides a visual evaluation of the generative models' performance on the lens distortion task, showcasing some examples of model predictions and landmark detection.

\begin{figure}
\centering
\begin{subfigure}[b]{1\linewidth}
    \includegraphics[width=\linewidth]{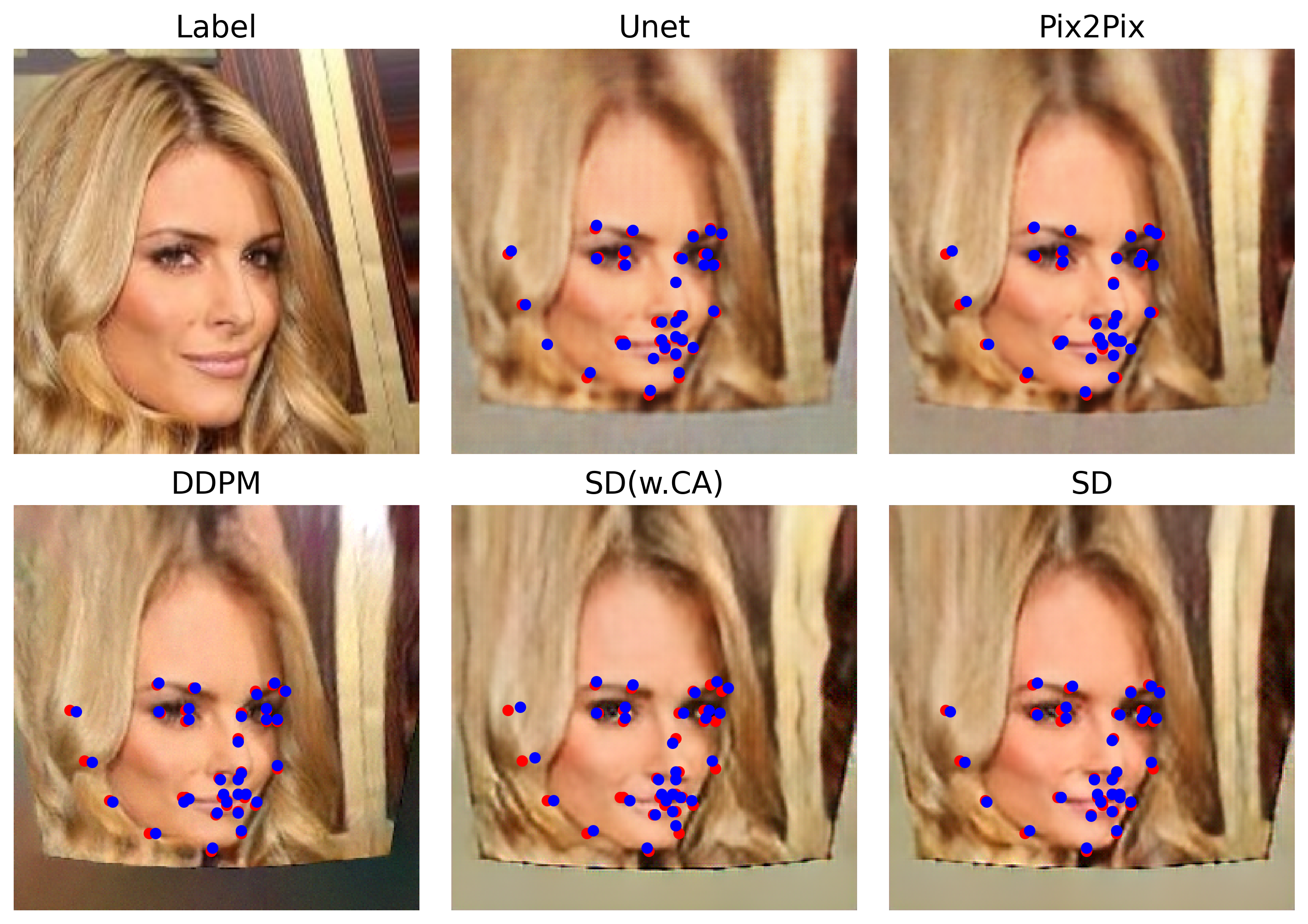}
    \label{fig:imageA}
\end{subfigure}
\vfill
\begin{subfigure}[b]{1\linewidth}
    \includegraphics[width=\linewidth]{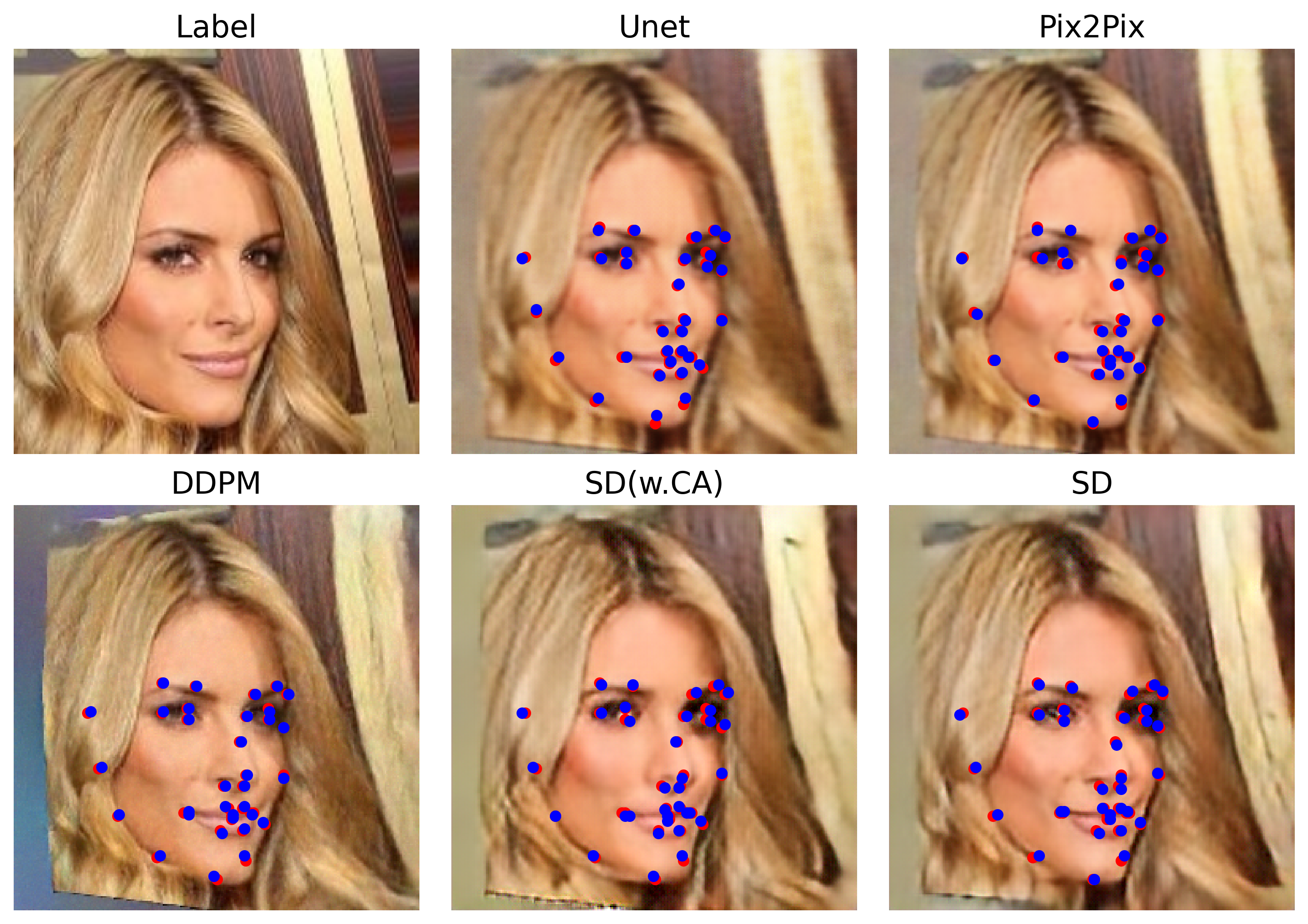}
    \label{fig:imageB}
\end{subfigure}
\caption{\small Comparison of simulation data with predicted landmarks for a single sample. The left image shows the input, while the right image overlays the predicted (blue dots) and true (red dots) landmark positions on the model output. For clarity, only every second landmark is visualized.}
\label{}
\end{figure}

\begin{figure}
\centering
\begin{subfigure}[b]{1\linewidth}
    \includegraphics[width=\linewidth]{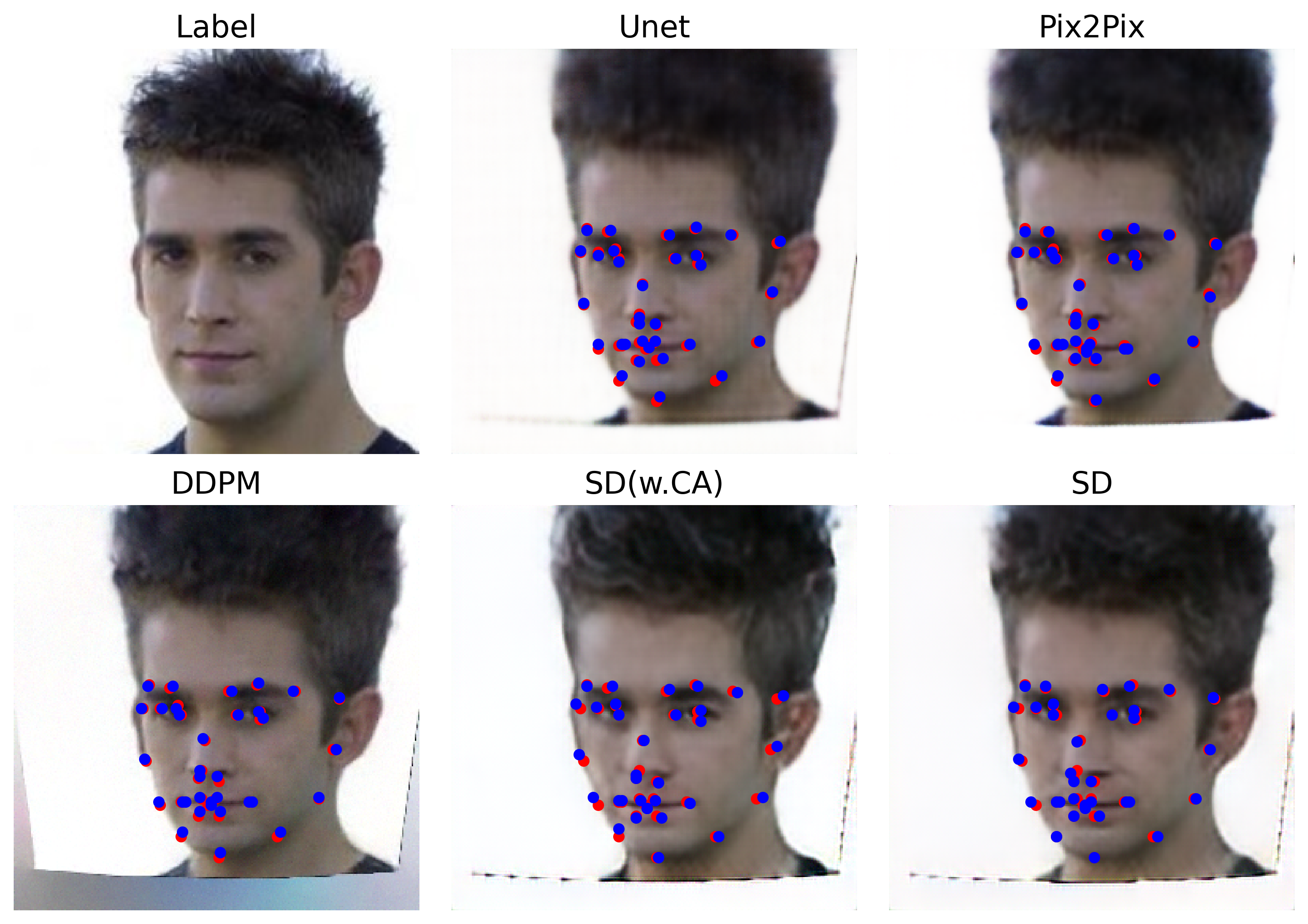}
    \label{fig:imageA}
\end{subfigure}
\vfill
\begin{subfigure}[b]{1\linewidth}
    \includegraphics[width=\linewidth]{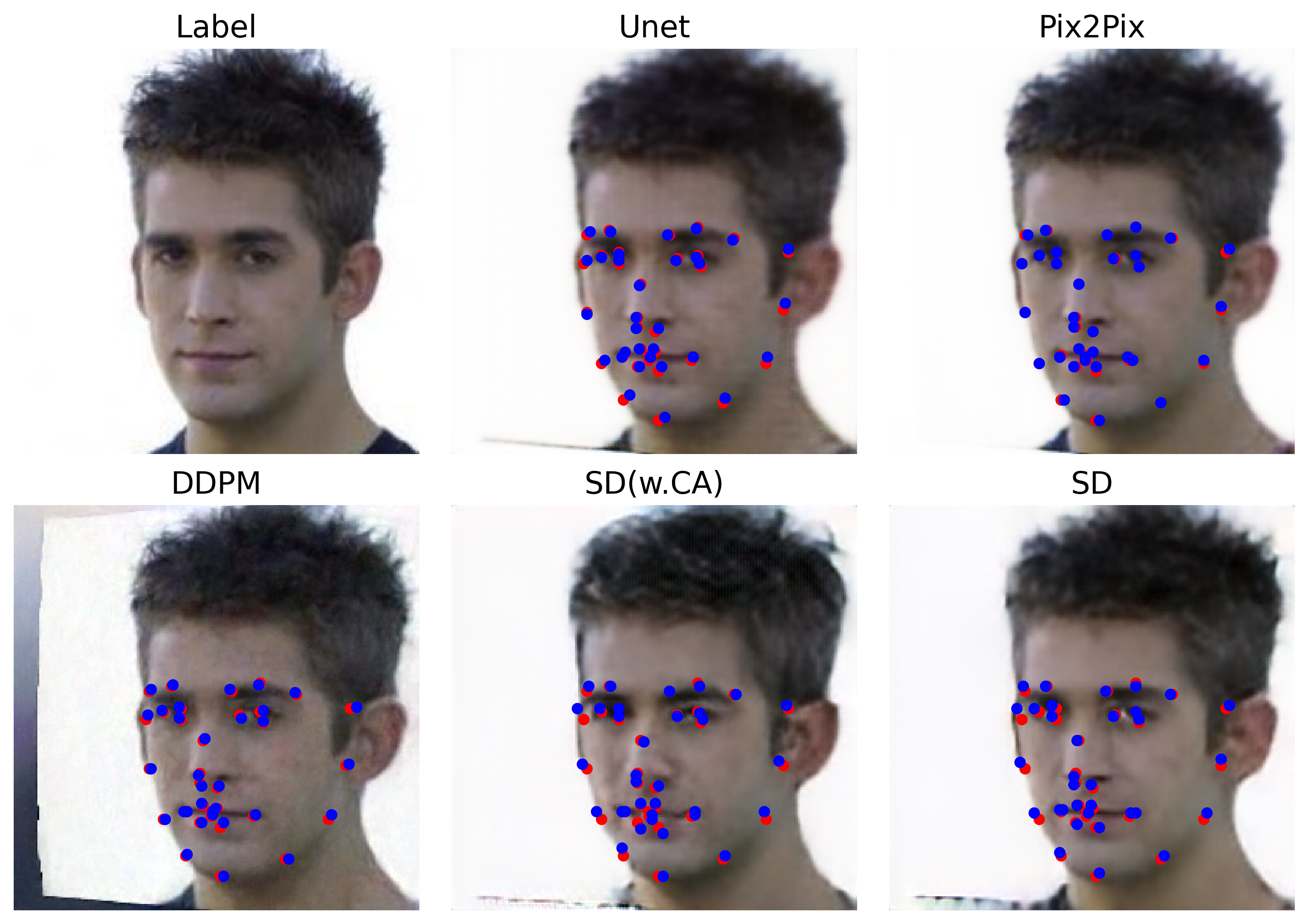}
    \label{fig:imageB}
\end{subfigure}
\caption{\small Comparison of simulation data with predicted landmarks for a single sample. The left image shows the input, while the right image overlays the predicted (blue dots) and true (red dots) landmark positions on the model output. For clarity, only every second landmark is visualized.}
\label{}
\end{figure}

\begin{figure}
\centering
\begin{subfigure}[b]{1\linewidth}
    \includegraphics[width=\linewidth]{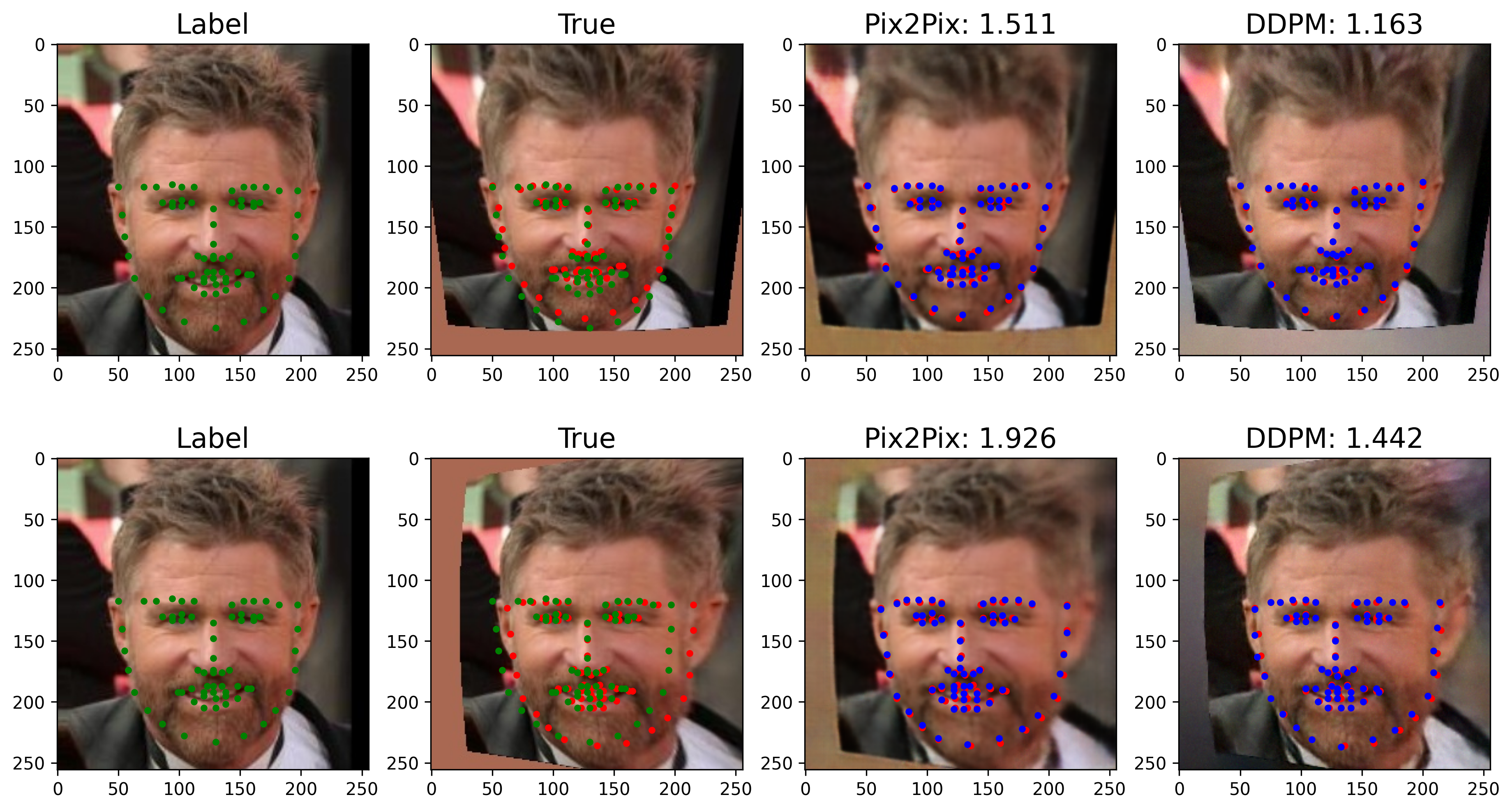}
    \label{fig:imageA}
\end{subfigure}
\vfill
\begin{subfigure}[b]{1\linewidth}
    \includegraphics[width=\linewidth]{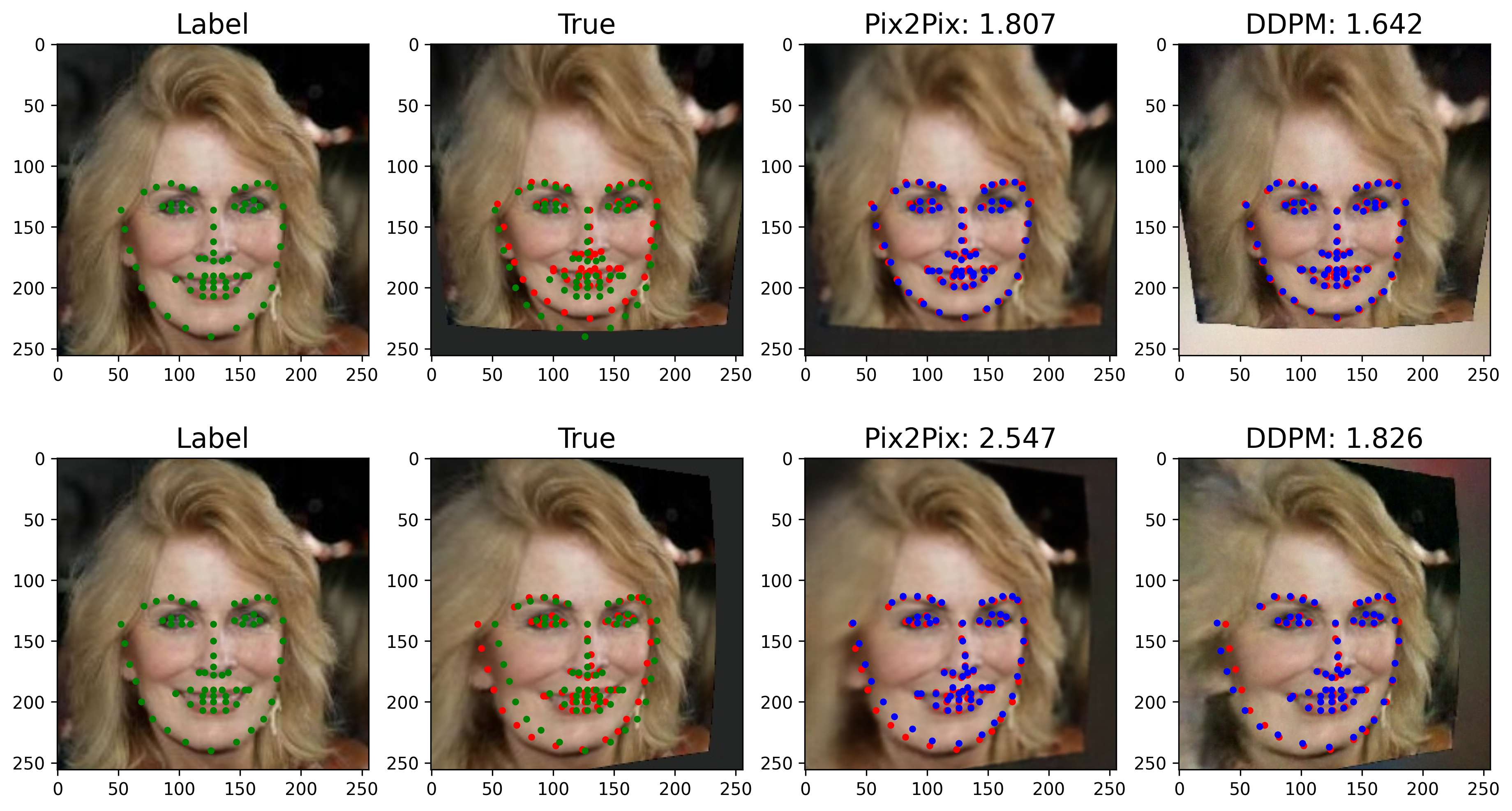}
    \label{fig:imageB}
\end{subfigure}
\caption{\small Comparing the input and output of the simulation with the predictions from Pix2Pix and DDPM for two random samples within the lens distortion task dataset. Green dots represent detected landmarks in the label image, red dots indicate the actual landmarks post-lens distortion, and blue dots denote the predicted landmarks by the models.
}
\label{}
\end{figure}

%% file: appendix/ball_appendix.tex
\clearpage
\section{Ball Appendix} \label{a:ball}

\subsection{Ball physics and kinematics} \label{a:ball_physics}

\begin{figure}[h!]
    \centering
    \includegraphics[width=0.5\textwidth]{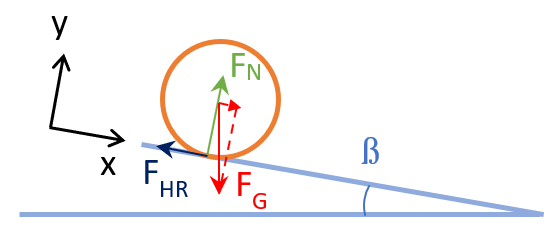}
    \caption{\small Forces overview for the rolling ball}
    \label{fig:RB_physics}
\end{figure}

\paragraph{Rolling ball.} \label{a:RB_physics}
The kinematic problem of the rolling ball can be relatively simply described using Newton's law and the angular momentum balance equation. \newline
The equations of motion for the X and Y directions are derived from the force balance acting on the ball, where no movement occurs in the Y direction (see figure~\ref{fig:RB_physics}):
\begin{center}
$\vec{F_{all}}=m\ast\vec{a}$ \:\: with \:\: $\vec{F_{all}}=\vec{F_G}+\vec{F_N}+\vec{F_{HR}}$
\end{center}

Equation of motion along x: 
\begin{equation}
\label{eq:RB_motion_x}
m\ast g\ast\sin{\left(\beta\right)}-F_{HR}=m\ast\ddot{x}
\end{equation}

Equation of motion along y (no movement): 
\begin{equation}
\label{eq:RB_motion_y}
F_N-m\ast g\ast\cos{\left(\beta\right)}=0
\end{equation}

To describe the rotation and the ball angle, the angular momentum balance from rotational mechanics can be used:
\begin{equation}
\label{eq:RB_rolling_angle}
F_{HR}\ast r=J\ast\alpha>0\Leftrightarrow F_{HR}\ast r=J\ast\ddot{\varphi}>0
\end{equation}

In the description of this first simple case, the ball position along the X-axis and the ball angle are derived twice.\newline
\vspace{-0.5cm}

\begin{figure}
    \centering
    \includegraphics[width=0.45\textwidth]{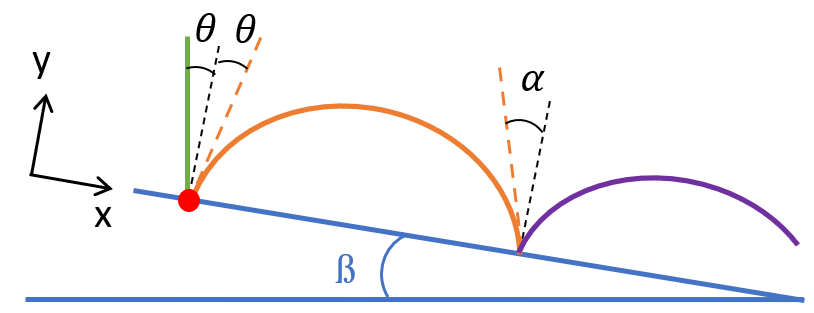}
    \caption{\small Splitting the movement of the bouncing ball}
    \label{fig:BB_movement}
\end{figure}

\paragraph{Bouncing ball.} \label{a:BB_physics}
The movement of the bouncing ball is much more complex to describe than the rolling case. For a better overview, the movement is divided into different sections (see different colors in figure~\ref{fig:BB_movement}). \newline
First, the ball is released from a defined height without any initial velocity (green area). Until it hits the inclined ground surface, it is in free fall. Derived from Newton's law, the following equations applies:

\begin{gather}
\label{eq:BB_free_fall}
F_g=m\ast a\ \Leftrightarrow m\ast\ddot{y}=-m\ast g\Leftrightarrow\ \ddot{y}=-g \\
\label{eq:BB_free_fall_time}
\Rightarrow y\left(t\right)=y_0-\frac{g}{2}\ast t^2
\end{gather}

\begin{figure}[h!]
    \centering
    \includegraphics[width=0.5\textwidth]{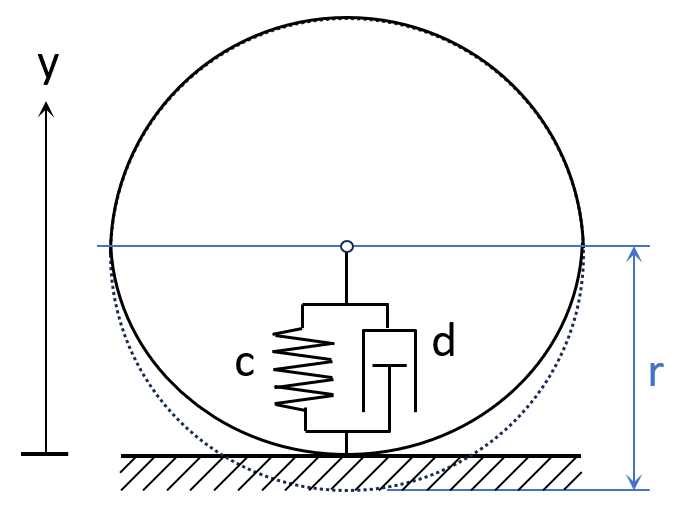}
    \caption{\small Spring-damper model to simulate the ball impact and bounce on the ground}
    \label{fig:BB_ground_impact}
\end{figure}

Subsequently, the contact with the ground (red point on figure~\ref{fig:BB_movement}) can be represented using a simple spring-damper model (see figure~\ref{fig:BB_ground_impact}). It is assumed that the \textit{Pymunk}~\cite{Pymunk} physics simulation used calculates similarly, as an elasticity factor can be defined in the settings and not all calculations of the physics engine could be retraced. To be able to describe the whole ball movement, this section mainly focuses on determining the impact velocity of the ball. This is where the weight force, the spring force and the damping force with their respective constants m (ball mass), c (spring stiffness) and d (damping constant) occur.
\begin{equation}
    \begin{split}
        F_g+F_c+F_d=m\ast\ddot{y} \\
        m\ast\ddot{y}+d\ast\dot{y}-c\ast\left(r-y\right)=-m\ast g
        \label{eq:BB_ground_contact}
    \end{split}
\end{equation}

After the impact, the next motion section (the orange part in figure~\ref{fig:BB_movement}) is assumed to be an oblique throw with the initial velocity $v_0$, which was calculated using equation~\ref{eq:BB_ground_contact} above. Additionally, the principle "angle of incidence = angle of reflection" is used to determine the angle of rebound $\theta$ from the ball; this corresponds to the ground slope $\beta$ after the first ground contact. \newline
The following equations of motion for X and Y are derived from Newton's law:
\begin{gather}
    \label{eq:BB_MovementX}m\ast\ddot{x}=m\ast g\ast\sin{(\beta)} \\
    \label{eq:BB_MovementY}m\ast\ddot{y}=-m\ast g\ast\cos{(\beta)}
\end{gather}

The equations can also be expressed as functions of time by integrating them. $v_0$ is the initial velocity of this third phase of motion, which was calculated above in the second part of the movement (impact on the ground).
\begin{gather}
    \label{eq:BB_MovementX_time}
    x=\frac{g\ast\sin{(\beta)}}{2}\ast t^2+v_0\ast\sin{\left(\theta\right)}\ast t+x_0 \\
    \label{eq:BB_MovementY_time}
    y=-\frac{g\ast\cos{(\beta)}}{2}\ast t^2+v_0\ast\cos{\left(\theta\right)}\ast t+y_0
\end{gather}

When the ball hits the ground for the second time after the "oblique throw", the orange motion phase ends and the purple phase begins. In between, there is another bounce, where equation~\ref{eq:BB_ground_contact} helps to determine the second rebound velocity. \newline 
The further movement is then described by the above equations~\ref{eq:BB_MovementX} and~\ref{eq:BB_MovementY}. This requires the impact angle $\alpha$, which can be calculated at the end of the orange phase as follows:
\begin{equation}
    \label{eq:BB_impact_angle}
    \tan{\left(\alpha\right)}=\frac{v_y}{v_x}=\frac{-g\ast t_{Auftreff}+v_{y,0}}{v_{x,0}}
\end{equation}
\begin{center}
$t_{Auftreff}$ can be calculated using the equation of motion in the y-direction by setting it to zero.
\end{center}

Finally the entire ball movement can be described step by step using the equations listed above. The rotation of the ball is determined as for the rolling case, by forming the angular momentum balance for each bounce, taking into account the friction force of the ground. \newline
In the bouncing case, all degrees of freedom (movements in the X and Y directions, as well as the ball rotation) contain second-order terms. In addition, it becomes evident through the more complex equation~\ref{eq:BB_ground_contact} of the ball impact that the motion along the Y-axis is more complex here than along the ground.

\subsection{Bouncing ball} \label{a:BB_training_all}
\subsubsection{Further evaluation and results} \label{a:BB_eval_results}

\begin{figure}
    \centering
    \includegraphics[width=1\linewidth]{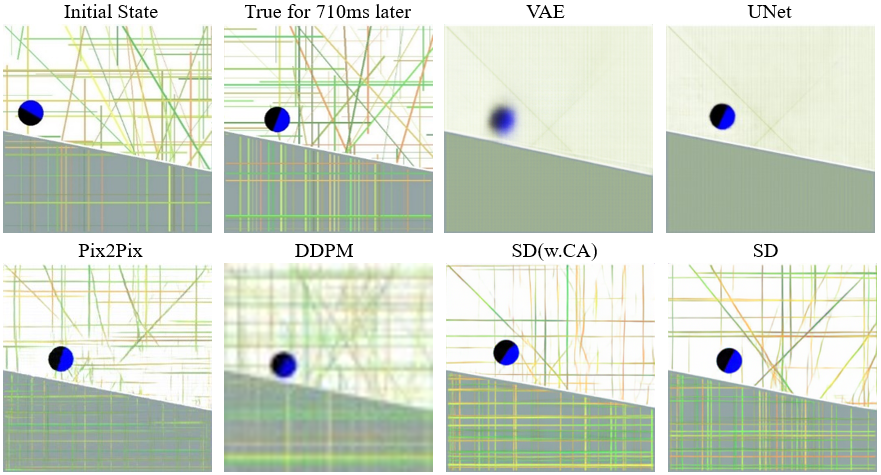}
    \caption{\small Predictions of the generative AI compared to the physics simulation for the rolling ball}
    \label{fig:BB_PredImgs2}
\end{figure}

Another result for the bouncing ball is shown in figure~\ref{fig:BB_PredImgs2}. Here it can be seen that the ball position varies slightly depending on the algorithm. The biggest error is made by the diffusion network along the vertical: the ball is drawn about 9px too low.

In the table~\ref{tab:BB_Results_SumTable}, the average results for all evaluation criteria and the three AI methods are shown.\\
\begin{table*}[ht]
\caption{\small Prediction results for the Pix2Pix, UNet, autoencoder and diffusion networks for the bouncing ball}
\label{tab:BB_Results_SumTable}
\centering
\begin{tabular}{c|c|c|>{\centering\arraybackslash}p{1.4cm}|c|>{\centering\arraybackslash}p{2.3cm}|>{\centering\arraybackslash}p{2cm}}
    \textbf{Model} & \textbf{Metric} & \textbf{Mean} & \textbf{Standard deviation} & \textbf{Error} & \textbf{Number of balls} & \textbf{Position to start ball}\\
    & & & & & 0 / \textgreater1 / Error & Ahead / Error \\
    \midrule
    \multirow{5}{*}{Pix2Pix} & Position X & 6.28 & 7.98 & 7\% & & \\
     & Position Y & 11.7 & 12.8 & 7\% & & \\
     & Rotation & 17.2 & 20.8 & 15\% & 7\% / 0\% / 0\% & 1\% / 7\% \\
     & Roundness & 0.56 & 0.14 & 10\% & & \\
     & Ground slope & 0 & 0 & 0\% & & \\
    \midrule
    \multirow{5}{*}{UNet} & Position X & 5.53 & 7.48 & 20\% & &  \\
     & Position Y & 10.8 & 12.2 & 20\% & & \\
     & Rotation & 15.2 & 22.9 & 30\% & 19\% / 0\% / 1\% & 0\% / 20\% \\
     & Roundness & 0.74 & 0.21 & 35\% & &  \\
     & Ground slope & 0 & 0 & 0\% & &  \\
    \midrule
    \multirow{5}{*}{VAE} & Position X & 4.69 & 6.13 & 89\% & &  \\
     & Position Y & 6.25 & 6.94 & 89\% & & \\
     & Rotation & 31.0 & 39.8 & 97\% & 20\% / 0\% / 68\% & 1\% / 89\% \\
     & Roundness & 0.90 & 0.14 & 99\% & &  \\
     & Ground slope & 1.00 & 0.04 & 25\% & &  \\
    \midrule
    \multirow{5}{*}{convAE} & Position X & 4.24 & 3.85 & 89\% & &  \\
     & Position Y & 6.08 & 5.93 & 97\% & & \\
     & Rotation & 12.2 & 8.61 & 99\% & 11\% / 0\% / 86\% & 0\% / 97\% \\
     & Roundness & 1.06 & 0 & 99.9\% & &  \\
     & Ground slope & 1.00 & 0.05 & 26\% & &  \\
    \midrule
    \multirow{5}{*}{DDPM} & Position X & 7.91 & 9.04 & 3\% & &  \\
     & Position Y & 15.5 & 13.7 & 3\% & & \\
     & Rotation & 32.9 & 33.8 & 8\% & 0\% / 1\% / 2\% & 6\% / 3\% \\
     & Roundness & 0.61 & 0.17 & 6\% & & \\
     & Ground slope & 0 & 0.08 & 3\% & & \\
     \midrule
     \multirow{5}{*}{SD(w.CA)} & Position X & 40.0 & 48.5 & 0\% & &  \\
     & Position Y & 24.8 & 22.9 & 0\% & & \\
     & Rotation & 61.1 & 52.5 & 21\% & 0\% / 0\% / 0\% & 12\% / 0\% \\
     & Roundness & 0.53 & 0.15 & 2\% & &  \\
     & Ground slope & 1.80 & 1.51 & 0\% & &  \\
     \midrule
     \multirow{5}{*}{SD} & Position X & 8.55 & 11.9 & 0\% & &  \\
     & Position Y & 16.2 & 14.1 & 0\% & & \\
     & Rotation & 34.2 & 37.8 & 6\% & 0\% / 0\% / 0\% & 5\% / 0\% \\
     & Roundness & 0.47 & 0.11 & 0\% & &  \\
     & Ground slope & 0.14 & 1.08 & 0\% & &  \\
\end{tabular}
\end{table*}

In order to supplement the main section with a few more details on the respective AI networks, their strengths and weaknesses are now summarised. A few typical error patterns are also listed for each network.

\paragraph{Pix2Pix (GAN architecture):}
\begin{itemize}
    \item Most stable mapping of the bouncing ball problem: best results achieved relative to the number of errors
    \item Most error cases for the ball rotation: 15\% of the generated images are not interpretable regarding the error metric. For the "valid" 85\%, there is an average error of 17,2° with a standard deviation of 20,8°, reflecting the high variance in the results.
    \item Only one ball appears on the generated images in 93\% of cases, while in the remaining 7\%, no proper ball is represented. In these cases, either no ball is present on the image or the depicted shape is too fragmented and has no roundness at all.
    \item The network most frequently places the "target ball" in the X-direction (i.e. along the ground slope) behind the start ball. In fact, the generated image must always show the ball with a larger X-coordinate than on the input image, otherwise the correctness of the physics is not given. Only in 1\% of the cases the target ball is in front of the start ball; otherwise all valid predictions correctly represent the rolling along the inclined surface. 
    \item Good representation of the background: the line structures are most accurately represented for the horizontal and vertical lines. The diagonal segments are usually only partially drawn.
    \item Correlations between error and simulation parameters:
    
    \begin{itemize}
        \item Position error in the X-direction increases the most with a larger time interval between input image and prediction, followed by an increasing ground inclination.
        \item Position error in the Y-direction: a slight correlation is visible here between increasing start height of the ball and increasing error, although the ground inclination and the time interval have almost the same influence.
        \item Ball angle deteriorates when the time interval between the start and target images increases.
        \item No real correlation is observed for the roundness error.
    \end{itemize}
    
\end{itemize}

\paragraph{UNet:}
\begin{itemize}
    \item General: slightly better results in position and rotation compared to the Pix2Pix, if only the successfully analysable images are considered. Accuracy in the X-direction is about 1px better and rotation about 2° better.
    \item Algorithm stability: much more unstable: 20\% to 35\% of the predictions are not evaluable. Many images are generated that contain no ball or a shape that does not approximate a ball. 
    \item \textbf{Target ball placement:} the network also places very well the target ball "behind" the start ball in terms of the X-coordinate. All valid result images comply with the physical correctness of rolling along the ground slope. Where this is not fulfilled, the images are not evaluable.
    \item Background depiction: blurry representation of the background. As seen in the result images, both the ground and the sky are depicted almost monotonously. The UNet seems unable to represent fine line structures; only very attenuated vertical lines can be guessed. Similarly, the different colours of the lines are not shown: the ground appears uniformly gray and the sky light green.
    \item Correlations between error and simulation parameters:

    \begin{itemize}
        \item Position error in the X-direction increases most with a larger time interval, followed by an increasing ground inclination (similar to the Pix2Pix)
        \item Position error in Y-direction increases mainly with greater starting height of the ball, although the correlation remains relatively weak (as with the Pix2Pix)
        \item Ball angle deteriorates when the time interval between the start and target image increases
        \item For roundness, unlike the Pix2Pix, there is a dependence on the time interval; if this becomes larger, the error also increases.
    \end{itemize}
\end{itemize}

\begin{figure}[h!]
    \centering
    \includegraphics[width=1\linewidth]{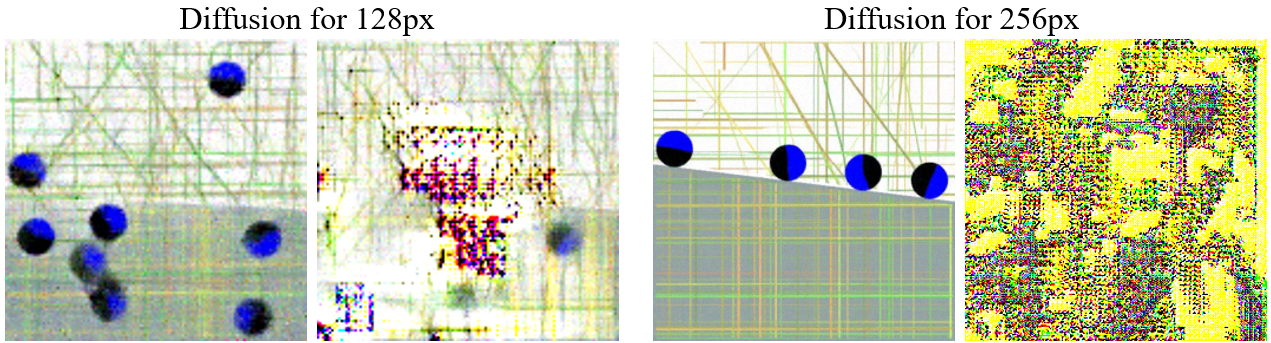}
    \caption{\small Typical predictions for the DDPM model trained with 128 and 256px images}
    \label{fig:Err_DiffTraining_highRes}
\end{figure}

\paragraph{Diffusion:} 
Three different models were examined under the category of "diffusion": the DDPM, stable diffusion with cross-attention (SDw.CA) and without (SD). In general, the results of the DDPM and SD are very similar to each other, with the SD showing a little more stability; the SDw.CA performs significantly worse except for the mapping of ball roundness, which is very good.\newline
For the DDPM network different approaches and settings were tested because the results of the first runs with 256px images were very poor. On one image, several balls appeared at randomly distributed positions or the prediction was noisy and contained artefacts (see figure~\ref{fig:Err_DiffTraining_highRes}).
Since no approach or parameter set could be found to achieve better predictions with this algorithm, the image size was progressively reduced. This had a very positive influence on the training, making the result images from the 64px network comparable to the Pix2Pix and UNet approaches. Therefore, only results for 64px images are shown in this paper for the DDPM, which explains the blurriness caused by up-scaling the predictions to 256px.
With the stable diffusion approach there were no such resolution problems, so that the 256px images could be used directly.

\begin{itemize}
    \item In general, the error criteria of the diffusion images are worse than those of the other analysed networks. However, depending on the error measure, between 4 and 7\% fewer non-evaluable images are produced compared to the Pix2Pix and significantly fewer than with the UNet, which has the highest error rate. 
    \item In the position, the accuracy is between 25-40\% worse. Rotation proves to be the biggest weakness of the approach, as the error here is twice as high as with Pix2Pix and UNet. The roundness is mapped similarly to the other methods and even a bit better for the stable diffusion. This is the strength of this method, which delivers very high-quality images with round balls. The SD approach depicts the roundness the best without generating error images regarding this criterion.
    \item It is noticeable that the ground slope cannot be evaluated in approx. 3\% of the images for the DDPM. Depending on the diffusion method, the mean error is also different from zero, which was not the case either with Pix2Pix or with UNet.
    \item Regarding the number of balls in the result images, the diffusion approach performs best. For the DDPM in 97\% of the images, only one ball is visible (which is not even achieved by GAN at 93\%) and the remaining 3\% is divided between images with multiple balls and error images. With stable diffusion, there are even no images with several or no balls.
    \item The approach is a bit less reliable in terms of mapping the roll of the target ball along the ground slope. SDw.CA in particular performs worst with 12\% of the cases where the target ball is located ahead the start ball along the X-axis, which is physically impossible. The DDPM and SD shows quite the same results with about 5 to 6\% ahead target balls. Nevertheless, the predictions that correctly represent this relationship are not much worse with diffusion, as the error rate is reduced.
    \item The prediction images correctly represent the different colors as well as the line structures in the sky and ground. Due to the blurred image, the visual evaluation is worse for the DDPM, but all "image elements" are present. Both stable diffusion algorithms generate very high-quality images in which the colors, the background pattern and the ball itself are mapped very accurately.
    \item Correlations between errors and simulation parameters:

    \begin{itemize}
        \item Generally, there are hardly any recognizable correlations with the simulation settings: only for the position error along the X axis and the ball rotation some relationships can be noticed.
        \item Position error in X direction has a slight dependency on the time interval between input and target image (a larger interval leads to greater inaccuracy)
        \item Ball angle deteriorates the further the ball of the input image has moved along the ground surface. For input images with a ball already in the middle of the image, the rotation is slightly worse. This correlation is slightly stronger for the SD approach compared to the DDPM.
        \item No correlations are present for the Y position and ball roundness errors.
    \end{itemize}
\end{itemize}

\begin{figure}[h!]
    \centering
    \includegraphics[width=1\linewidth]{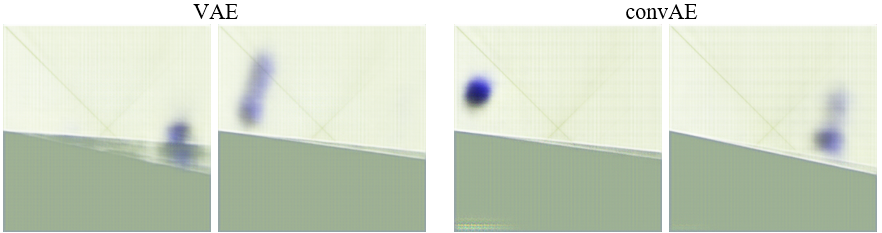}
    \caption{\small Typical error patterns with auto encoders (VAE or convAE) for the bouncing case - no clean representation of the ground surface and very blurred ball}
    \label{fig:Err_AE_typical_patterns}
\end{figure}

\paragraph{Auto encoder:}
Two approaches were examined for the autoencoders: the variational AE (VAE) and the convolutional AE (convAE). However, no evaluable results could be generated by the networks for the bouncing ball. Depending on the error criterion, between 89 and 99\% of the predictions are evaluated as invalid (see results in Table~\ref{tab:BB_Results_SumTable}). \newline
In general, it can be said that the obtained images are very blurred. In addition, the ball is rarely displayed as a single point but is often drawn out as if it were taking up several positions in the image. The ground surface is also not displayed cleanly as a single line; instead, it looks as if several slopes are depicted using different lines. There are also small white stripes in the ground at some images. These typical artefacts or errors of the autoencoder can be seen in Figure~\ref{fig:Err_AE_typical_patterns}. Structurally, meaning in terms of colors and background pattern, the predictions look similar to the UNet results, as the networks are closely related in design. \newline
Since so many generated predictions are considered invalid when evaluating the error criteria, the results for the bouncing ball in terms of position, angle and roundness are hardly relevant. Therefore, no correlations between simulation parameters and error measures are provided here.

In table~\ref{tab:BB_Results_DetailedTable} the detailed results of different training and evaluations are visible for the Pix2Pix and UNet approaches. Indeed for these two networks many runs were carried out to see how the models behave over multiple trainings. In the overview tables already shown in this paper, the mean values of all runs of one method are shown for each error criterion. 
\begin{table*}
\caption{\small Detailed results of each training run for the Pix2Pix (GAN) and UNet networks where multiple runs were carried out to get more representative mean values for the bouncing ball}
\label{tab:BB_Results_DetailedTable}
\centering
\small
\begin{tabular}{c|c|c|c|c|c}
    & \textbf{Rotation} & \textbf{Position X} & \textbf{Position Y} & \textbf{Position} & \textbf{Roundness}\\
    & Mean / Std / Err & Mean / Std & Mean / Std & Error & Mean / Std / Err \\
    \midrule
    \multirow{4}{*}{Pix2Pix} & 15.9 / 18.8 / 58 (4\%)\:\: & 6.46 / 8.27 & 12.2 / 12.9 & 6 (0\%) &  0.50 / 0.11 / 16 (1\%)\:\: \\
     & 16.7 / 19.9 / 112 (7\%)\: & 6.31 / 8.17 & 12.1 / 13.0 & 14 (1\%) & 0.51 / 0.12 / 23 (1\%)\:\: \\
     & 14.5 / 20.5 / 587 (37\%) & 5.71 / 7.25 & 10.3 / 11.9 & 433 (27\%) & 0.73 / 0.20 / 612 (38\%) \\
     & 21.6 / 24.1 / 229 (14\%) & 6.65 / 8.23 & 12.4 / 13.3 & 14 (1\%) & 0.51 / 0.11 / 19 (1\%)\:\: \\
    \midrule
    \multirow{4}{*}{UNet} & 15.1 / 23.4 / 495 (31\%) & 5.30 / 7.44 & 10.8 / 11.8 & 310 (19\%) & 0.73 / 0.22 / 542 (34\%) \\
     & 15.9 / 23.7 / 398 (25\%) & 5.55 / 7.62 & 10.9 / 12.1 & 261 (16\%) & 0.75 / 0.21 / 540 (34\%) \\
     & 15.2 / 24.1 / 458 (29\%) & 5.54 / 7.62 & 11.3 / 12.9 & 274 (17\%) & 0.74 / 0.22 / 520 (33\%) \\
     & 14.5 / 20.5 / 587 (37\%) & 5.71 / 7.25 & 10.3 / 11.9 & 433 (27\%) & 0.73 / 0.20 / 612 (38\%) \\
    \noalign{\vskip 0.2cm} 
\end{tabular}
\begin{tabular}{c|c|c|c}
    & \textbf{Ground slope} & \textbf{Number of balls} & \textbf{Position to start ball} \\
    & Mean / Std / Error & 0 \: / \: \textgreater 1 \: / \: Error & Ahead \: / \: Error \\
    \midrule
    \multirow{4}{*}{Pix2Pix} & 0 / 0 / 0 (0\%) & 3 (0\%) / 1 (0\%) / 0 (0\%) & 9 (1\%) / 6 (0\%) \\
     & 0 / 0 / 0 (0\%) & 12 (1\%) / 2 (0\%) / 0 (0\%) & 16 (1\%) / 14 (1\%) \\
     & 0 / 0 / 0 (0\%) & 416 (26\%) / 0 (0\%) / 17 (1\%) & 0 (0\%) / 433 (27\%) \\
     & 0 / 0 / 0 (0\%) & 14 (1\%) / 0 (0\%) / 0 (0\%) & 5 (0\%) / 14 (1\%) \\
    \midrule
    \multirow{4}{*}{UNet} & 0 / 0 / 0 (0\%) & 304 (19\%) / 0 (0\%) / 4 (0\%) & 3 (0\%) / 310 (19\%) \\
     & 0 / 0 / 0 (0\%) & 256 (16\%) / 1 (0\%) / 2 (0\%) & 3 (0\%) / 261 (16\%) \\
     & 0 / 0 / 0 (0\%) & 263 (16\%) / 0 (0\%) / 8 (1\%) & 3 (0\%) / 274 (17\%) \\
     & 0 / 0 / 0 (0\%) & 416 (26\%) / 0 (0\%) / 17 (1\%) & 96 (6\%) / 44 (3\%)\:\: \\
\end{tabular}
\end{table*}
\subsubsection{Typical error patterns} \label{a:BB_errors}
In this section, typical incorrect predictions of the analysed generative networks for the bouncing ball are presented to illustrate the above analysis and the error rate of the networks. Due to the poor results of the autoencoders, no error images are presented for this network structure (the typical appearance of the predictions has already been explained above). The visible errors or inaccuracies are described in the caption of the respective figures.
\begin{figure}[h!]
    \centering
    \includegraphics[width=1\linewidth]{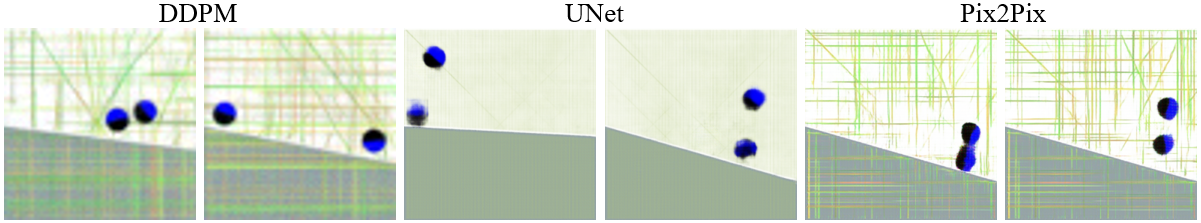}
    \caption{\small Several balls represented on the predictions of the DDPM, the UNet and the Pix2Pix algorithms.}
    \label{fig:Err_BB_2balls}
\end{figure}
\begin{figure}
    \centering
    \includegraphics[width=1\linewidth]{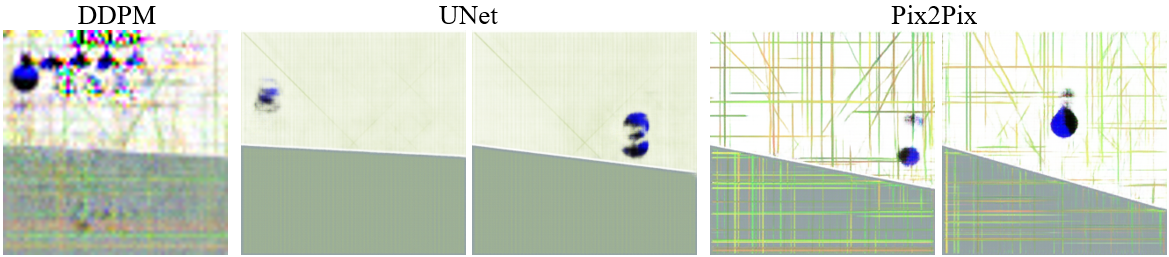}
    \caption{\small Typical artefacts that occur for different trained model}
    \label{fig:Err_BB_artefacts}
\end{figure}
\begin{figure}
    \centering
    \includegraphics[width=1\linewidth]{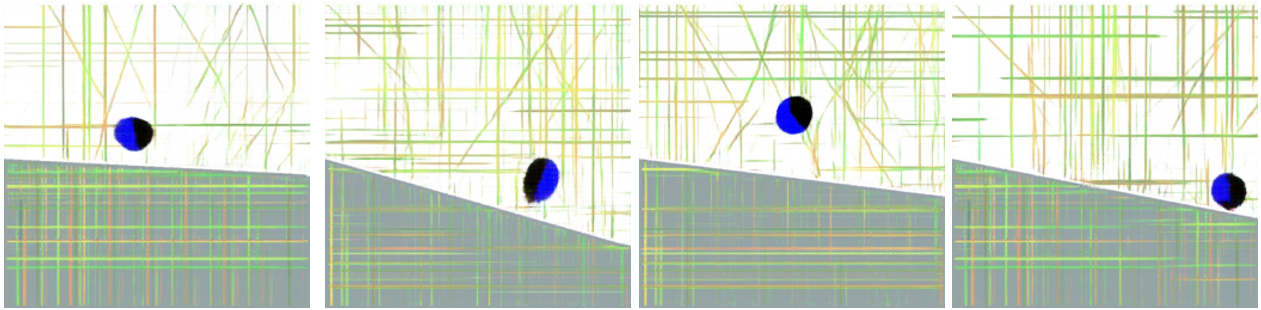}
    \caption{\small Not-round ball and imperfect color separation between both ball halves (only Pix2Pix predictions)}
    \label{fig:Err_BB_unround_colorsep}
\end{figure}
\begin{figure}
    \centering
    \includegraphics[width=1\linewidth]{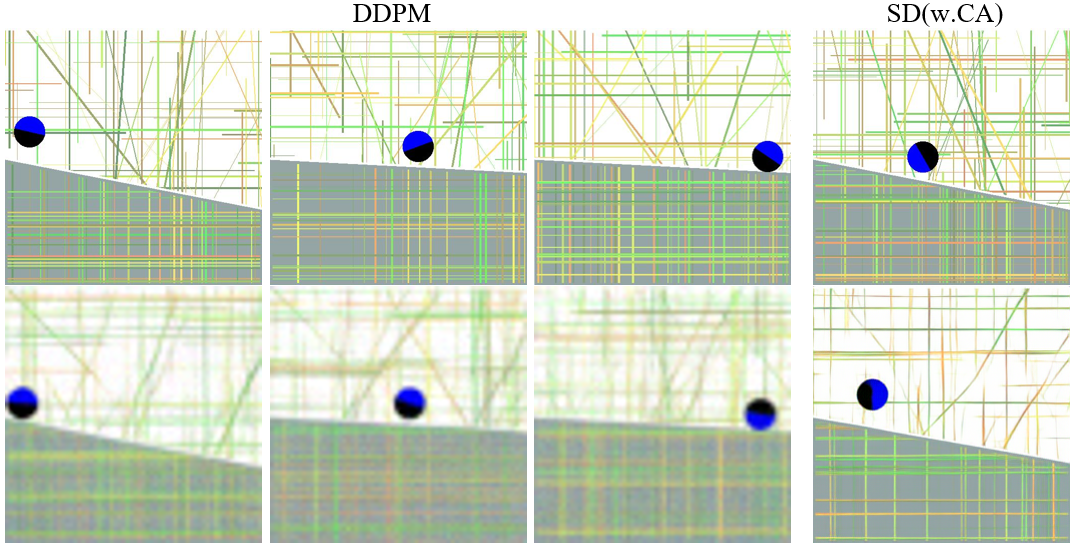}
    \caption{\small Position of the target ball ahead of the start ball along the X-axis (1st line: start ball; 2nd line: corresponding network prediction)}
    \label{fig:Err_BB_TargetAheadStart}
\end{figure}
\begin{figure}
    \centering
    \includegraphics[width=1\linewidth]{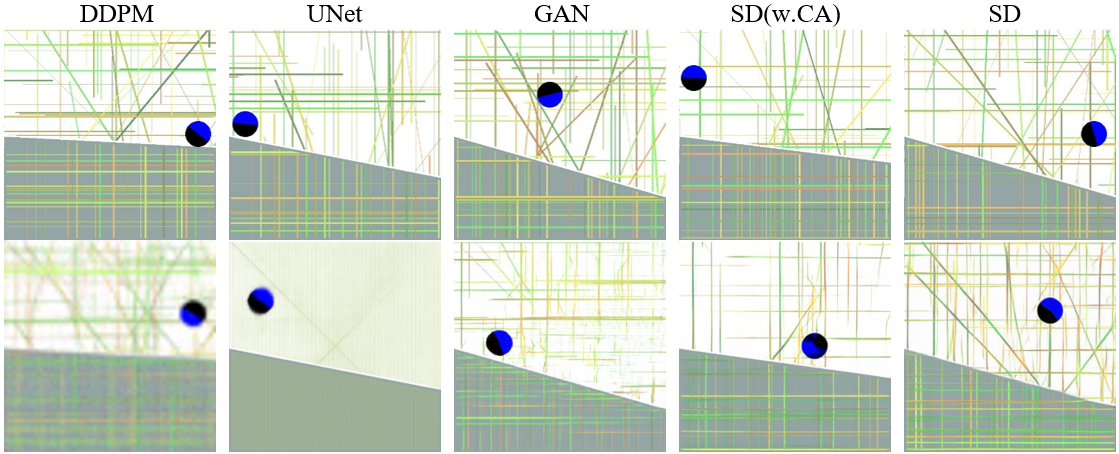}
    \caption{\small Errors concerning the ball position and rotation (1st line: true images; 2nd line: corresponding net prediction)}
    \label{fig:Err_BB_WrongPosRot}
\end{figure}
\subsection{Rolling ball} \label{a:RB_training_all}
\subsubsection{Further evaluation and results} \label{a:RB_eval_results}
\begin{figure}[h!]
\centering
\includegraphics[width=1\linewidth]{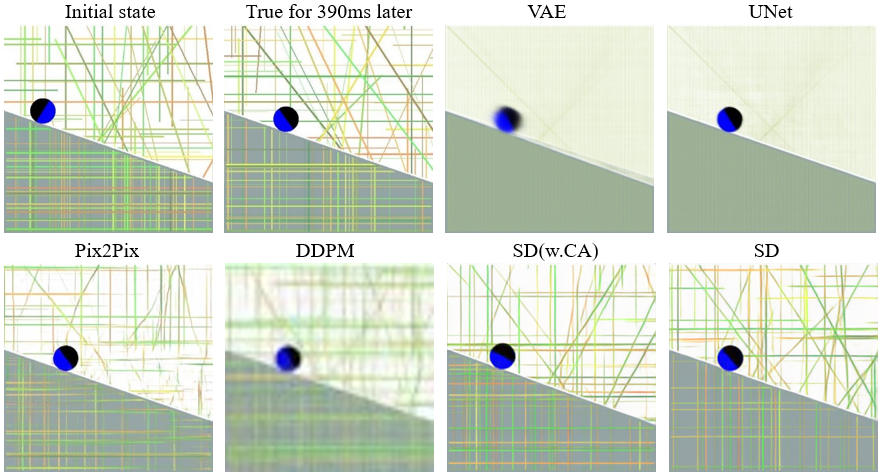}
\caption{\small Predictions of the generative AI compared to the physics simulation for the rolling ball}
\label{fig:RB_PredImgs_1}
\end{figure}

Result images of the predictions for the rolling ball are shown below on figure~\ref{fig:RB_PredImgs_1}. For this simpler physics problem, there are only two variable parameters in the simulation: the ground slope and the position of the "start ball" on the image. Of course, predictions are also expected from the AI networks for any given time interval. 
Visually, the same advantages and disadvantages as with the bouncing ball can be recognized for the analysed AI algorithms (see section~\ref{a:BB_eval_results}). Here too, the DDPM network creates a 64px image, which is then scaled up to 256px. That is because the network trains better with the 64px images and delivers more "stable" results. The up-scaling explains the blurriness of the diffusion image.  

In the table~\ref{tab:RB_Results_SumTable}, the average results of several runs for the investigated AI networks for the rolling ball are shown.
\begin{table*}[h]
\caption{\small Prediction results for the Pix2Pix, UNet, autoencoder and diffusion networks for the rolling ball}
\label{tab:RB_Results_SumTable}
\centering
\footnotesize
\begin{tabular}{c|c|c|>{\centering\arraybackslash}p{1.3cm}|c|>{\centering\arraybackslash}p{1.8cm}|>{\centering\arraybackslash}p{1.8cm}|>{\centering\arraybackslash}p{1.5cm}}
    \textbf{Model} & \textbf{Metric} & \textbf{Mean} & \textbf{Standard deviation} & \textbf{Error} & \textbf{Number of balls} & \textbf{Position to start ball} & \textbf{Distance to ground}\\
    & & & & & 0 / \textgreater1 / Error & Ahead / Error & Error \\
    \midrule
    \multirow{5}{*}{Pix2Pix} & Position X & 4.26 & 8.36 & 1\% & & & \\
     & Position Y & 1.61 & 3.31 & 1\% & & & \\
     & Rotation & 20.9 & 35.2 & 13\% & 1\% / 0\% / 0\% & 0\% / 1\% & 2\% \\
     & Roundness & 0.56 & 0.13 & 3\% & & & \\
     & Ground slope & 0 & 0 & 0\% & & & \\
    \midrule
    \multirow{5}{*}{UNet} & Position X & 3.69 & 8.49 & 1\% & & & \\
     & Position Y & 1.4 & 3.63 & 1\% & & & \\
     & Rotation & 16.1 & 32.7 & 11\% & 3\% / 0\% / 0\% & 0\% / 3\% & 3\% \\
     & Roundness & 0.53 & 0.15 & 4\% & & & \\
     & Ground slope & 0 & 0 & 0\% & & & \\
    \midrule
    \multirow{5}{*}{VAE} & Position X & 5.54 & 16.2 & 3\% & & & \\
     & Position Y & 2.44 & 6.26 & 3\% & & & \\
     & Rotation & 21.5 & 36.0 & 34\% & 3\% / 0\% / 0\% & 1\% / 3\% & 25\% \\
     & Roundness & 0.70 & 0.20 & 25\% & & & \\
     & Ground slope & 1.10 & 0.33 & 0\% & & & \\
    \midrule
    \multirow{5}{*}{convAE} & Position X & 4.57 & 9.47 & 2\% & & & \\
     & Position Y & 2.10 & 3.80 & 2\% & & & \\
     & Rotation & 21.4 & 38.2 & 20\% & 2\% / 0\% / 0\% & 0\% / 2\% & 12\% \\
     & Roundness & 0.65 & 0.21 & 12\% & & & \\
     & Ground slope & 1.11 & 0.32 & 0\% & & & \\
    \midrule
    \multirow{5}{*}{DDPM} & Position X & 7.38 & 8.56 & 1\% & & & \\
     & Position Y & 2.34 & 3.39 & 1\% & & & \\
     & Rotation & 34.2 & 38.6 & 10\% & 0\% / 0\% / 1\% & 3\% / 1\% & 3\%\\
     & Roundness & 0.74 & 0.16 & 6\% & & & \\
     & Ground slope & 0 & 0.05 & 1\% & & & \\
     \midrule
     \multirow{5}{*}{SD(w.CA)} & Position X & 24.1 & 33.2 & 1\% & & & \\
     & Position Y & 7.65 & 11.6 & 1\% & & & \\
     & Rotation & 59.1 & 52.1 & 12\% & 1\% / 0\% / 0\% & 3\% / 1\% & 2\%\\
     & Roundness & 0.55 & 0.13 & 2\% & & & \\
     & Ground slope & 1.06 & 0.39 & 0\% & & & \\
     \midrule
     \multirow{5}{*}{SD} & Position X & 7.36 & 14.7 & 0\% & & & \\
     & Position Y & 2.57 & 5.32 & 0\% & & & \\
     & Rotation & 29.9 & 40.4 & 4\% & 0\% / 0\% / 0\% & 1\% / 0\% & 1\%\\
     & Roundness & 0.53 & 0.11 & 1\% & & & \\
     & Ground slope & 0.02 & 0.13 & 0\% & & & \\
\end{tabular}
\end{table*}

The first noticeable point in the results is that the autoencoders provide evaluable predictions for the rolling case, even though they remain the most unstable method. In fact, the error images drop to around 3\% for the position and regarding the ball angle and roundness a maximum of 34\% invalid images are generated. Therefore, the results of this model structure can be discussed below.
Otherwise it can be seen that the errors for the position are smaller compared to the bouncing case. This is especially evident for the Y-coordinate, where a significant reduction of the error can be seen, as the ball here “only” rolls along the ground surface and no longer moves in the Y-direction. The third physical error measure, which describes whether the ball is lying on the ground surface, is approximately the same for all AI methods, ranging between 97-99\%. Only the autoencoders have difficulties to represent the ball rolling down on the ground. The ball roundness is mapped similarly to the bouncing ball, although it is much better with the UNet. Finally, the ball angle is represented about 2° worse by each algorithm, except for the stable diffusion models which are improving by approx. 2-4°. A few special features of the analysed AI approaches and differences compared to the bouncing ball are listed below.

\paragraph{Pix2Pix (GAN architecture):}

\begin{itemize}
    \item Significant reduction of "error images" created by the network that cannot be analysed regarding all error criteria except for the ball rotation. Here, the ball angle is still not recognisable for 13\% of the images.
    \item Slight increase in the rotation error but strong increase in the standard deviation compared to the bouncing case. This is the error measure that is most poorly mapped by the Pix2Pix model.
    \item With 99\% reliability, exactly one ball is represented on the predictions of the AI network (6\% improvement over the bouncing case). Similarly, in 99\% of cases, the target ball is placed behind the start ball in the X-direction (7\% improvement).
    \item Correlations between error and simulation parameters:

    \begin{itemize}
        \item Position errors in both directions (X and Y) increase the most with increasing ground slope, followed by a larger time interval. Here, the X and Y errors show the same dependencies on the physics simulation settings.
        \item Ball angle deteriorates with greater ground inclination and when the time interval between the start and target image increases. 
        \item For the roundness error, unlike the bouncing case, there is a slight dependence on the starting position of the ball: the further the ball of the input image has moved along the ground surface, the less round the ball is represented.
    \end{itemize}
    
\end{itemize}

\paragraph{UNet:}

\begin{itemize}
    \item The same insights and improvements apply to the UNet as to the Pix2Pix model concerning the position and rotation of the ball, the number of balls and the placement in X-direction of the target ball behind the input ball. \\
    With this approach, the error images have decreased the most compared to the bouncing case: the largest reduction of non-analysable images from 35\% to 4\% is observed for the ball roundness.
    \item The roundness, which is worse with the bouncing ball, achieves the best error measure for the rolling case.
    \item Overall, the UNet provides the best results for the rolling ball, slightly ahead of the Pix2Pix algorithm. However, the background of the predicted images (ground and sky) remains blurred, so that the GAN performs better in terms of overall appearance.
    \item Correlations between error and simulation parameters:

    \begin{itemize}
        \item Position errors in both directions increase as the time interval increases; there is a significant dependence here. There is also a smaller correlation with the increasing ground slope, but it is weaker.
        \item For ball rotation, the time interval is more dominant compared to the GAN, followed by the ground slope. As the parameters increase, the inaccuracy in the prediction of the ball angle also increases.
        \item For the roundness error, the same dependence as for the bouncing case is observed: a larger time interval between input and target images results in a poorer ball representation.
    \end{itemize}
    
\end{itemize}

\paragraph{Diffusion:} Even in the rolling case, the best results for the DDPM model were obtained with 64px images, explaining the blurriness of the predictions due to up-scaling. 

\begin{itemize}
    \item Like the bouncing ball, the error criteria of the diffusion method are worse than for GAN and UNet. Here, stable diffusion with cross-attention delivers the worst results, with the network performing two or three times worse than the other diffusion methods, depending on the criterion. The DDPM and SD approaches differ only slightly in the results, with the SD model demonstrating greater stability with a very low error rate and a better representation of the ball roundness. Compared to the bouncing case, the position error in the Y-direction is mainly reduced, as with the other algorithms, while the other error metrics remain nearly identical.
    \item Both the ball position (in both directions) and the angle are represented with about twice the inaccuracy compared to Pix2Pix and UNet for the DDPM and the SD model. The roundness is about 35\% worse.
    \item Regarding the mapping of the ground slope, the same is observed as for the bouncing case: very few error images and a small standard deviation around the mean value of 0 are present for the DDPM, while the UNet and the GAN architecture draw the ground almost "perfectly". The stable diffusion approaches even show a mean error greater than zero.
    \item With a 99\% reliability going to almost 100\% for the SD, exactly one ball is shown on the predictions of the diffusion networks and the few remaining predictions are not evaluable. Here, diffusion performs best, as about 1\% of the Pix2Pix and 3\% of the UNet  predictions do not contain a ball.
    \item Regarding the correct rolling of the ball on the ground surface, 3\% of predictions represent the target ball ahead of the input ball along the X-direction, which is physically impossible. Here, both other methods discussed above perform slightly better with almost 0\% target balls ahead. However, the diffusion network, like the Pix2Pix and the UNet, correctly depicts the ball on the ground surface in 97 to 99\% of cases.\\
    Overall the SD model shows the best results concerning the three criteria expected to verify the physical correctness.
    \item Correlation between error and simulation parameters:

    \begin{itemize}
        \item Position error in the X direction most increases with a larger time interval. There is also a smaller correlation with the increasing ground slope.
        \item Position error in the Y direction increases the most with increasing ground slope, followed by a larger time interval. \\ 
        Concerning the position errors along both axes the DDPM algorithm shows the strongest dependencies.
        \item Ball angle deteriorates the most when the time interval between the start and finish images increases. There is also a slightly weaker correlation with the ground slope.  
        \item No correlations are present for the roundness error.
    \end{itemize}
\end{itemize}

\paragraph{Autoencoder:} For the rolling ball, the two examined autoencoder approaches generate evaluable results. With a maximum of 34\% invalid images for ball angle predictions and an error rate of 3\% for the position, the quality of the predictions is significantly improved. Additionally, the average error metrics are close to those of the Pix2Pix model and even slightly better than the diffusion networks. Nonetheless, autoencoders remain the most unstable structure for modeling the ball problem, as they generate the highest number of non-evaluable images. The following observations can be made regarding the VAE and convAE approaches:

\begin{itemize}
    \item The convAE network generally provides slightly better results than the VAE. The error rates of convAE are a bit lower, as well as reduced average errors and standard deviations. However, the mapping of the ground and the representation of the ball angle are nearly identical between the two methods.
    \item As with the other models, the ball's position is represented less accurately along the X-axis compared to the Y-direction for the autoencoders. 
    \item The ball rotation is learned as effectively as with the Pix2Pix approach, but the error rate increases by 50\% to more than double in the case of the VAE algorithm.
    \item A notable observation is that the ball's roundness is represented worse compared to all other approaches examined, with significantly more invalid images. Similarly, the ground slope is learned less accurately by both models. With the SDw.CA method, the autoencoders are the only models showing a mean error greater than zero.
    \item Regarding physical correctness, these models generate the highest number of images without any ball present (3\%). Additionally, in 12\% (convAE) and 25\% (VAE) of the cases, the ball is not depicted on the ground surface, although this should always be the case for a rolling ball on an inclined surface. This error criterion stands out most significantly compared to the other models, where only 1 to 3\% of the images show the ball not in contact with the ground.
    \item Correlation between error and simulation parameters:
    
    \begin{itemize}
        \item Position errors in both directions increase the most as the time interval increases. A slight dependency can be observed to the ground slope, which generates more inaccuracy when getting greater (more pronounced for the Y-direction).
        \item Concerning the ball rotation the correctness of the model is worser when the time interval between start and target images increases. There is also a slightly weaker correlation with the ground slope.  
        \item For the roundness error, a correlation to the position of the ball on the image can be observed. The further the ball of the input image has moved along the ground surface, the less round it is depicted. This is the only model that shows a dependency between the roundness error and this simulation parameter.
    \end{itemize}
\end{itemize}

\vspace{\baselineskip}

In table~\ref{tab:RB_Results_DetailedTable} the detailed results of different training and evaluations are visible for each of the three analysed AI approaches. In the overview tables already shown in this paper, the mean values of all runs of one method are shown for each error criterion. 

\begin{table*}[h]
\caption{\small Detailed results of each training run for the Pix2Pix and UNet networks where multiple runs were carried out to get more representative mean values for the rolling ball}
\label{tab:RB_Results_DetailedTable}
\centering
\small
\begin{tabular}{c|c|c|c|c|c}
    & \textbf{Rotation} & \textbf{Position X} & \textbf{Position Y} & \textbf{Position} & \textbf{Roundness}\\
    & Mean / Std / Err & Mean / Std & Mean / Std & Error & Mean / Std / Err \\
    \midrule
    \multirow{5}{*}{Pix2Pix} & 21.7 / 36.0 / 199 (11\%) & 4.17 / 8.28 & 1.58 / 3.24 & 16 (1\%) & 0.60 / 0.14 / 34 (2\%) \\
     & 24.4 / 36.7 / 385 (21\%) & 4.35 / 8.14 & 1.67 / 3.32 & 26 (1\%) & 0.62 / 0.15 / 85 (5\%) \\
     & 21.8 / 35.4 / 377 (21\%) & 4.92 / 9.36 & 1.91 / 3.82 & 21 (1\%) & 0.57 / 0.14 / 63 (4\%) \\
     & 16.2 / 32.9 / 121 (7\%)\: & 3.80 / 8.43 & 1.38 / 3.26 & 3 (0\%) & 0.50 / 0.11 / 16 (1\%) \\
     & 20.2 / 34.8 / 98 (5\%)\:\: & 4.07 / 7.60 & 1.49 / 2.92 & 14 (1\%) & 0.51 / 0.11 / 16 (1\%) \\
    \midrule
    \multirow{4}{*}{UNet} & 16.8 / 33.5 / 176 (10\%) & 3.69 / 8.33 & 1.30 / 3.18 & 39 (2\%) & 0.52 / 0.15 / 62 (3\%) \\
     & 15.7 / 32.1 / 194 (11\%) & 3.59 / 8.35 & 1.33 / 3.26 & 40 (2\%) & 0.51 / 0.14 / 66 (4\%) \\
     & 16.2 / 33.6 / 187 (10\%) & 3.77 / 8.69 & 1.40 / 3.46 & 23 (1\%) & 0.53 / 0.15 / 62 (3\%) \\
     & 15.6 / 31.5 / 203 (11\%) & 3.71 / 8.59 & 1.57 / 4.61 & 9 (1\%) & 0.55 / 0.15 / 68 (4\%) \\
    \noalign{\vskip 0.3cm} 
\end{tabular}
\begin{tabular}{c|c|c|c|c}
    & \textbf{Ground slope} & \textbf{Number of balls} & \textbf{Position to start ball} & \textbf{Distance to ground} \\
    & Mean / Std / Err & 0 \: / \: \textgreater 1 \: / \: Error & Ahead \: / \: Error & Error \\
    \midrule
    \multirow{5}{*}{Pix2Pix} & 0 / 0 / 0 (0\%) & 16 (1\%) / 0 (0\%) / 0 (0\%) & 5 (0\%) / 16 (1\%) & 19 (1\%) \\
     & 0 / 0 / 0 (0\%) & 13 (1\%) / 0 (0\%) / 0 (0\%) & 8 (0\%) / 26 (1\%) & 34 (2\%) \\
     & 0 / 0 / 0 (0\%) & 21 (1\%) / 0 (0\%) / 0 (0\%) & 0 (0\%) / 21 (1\%) & 35 (2\%) \\
     & 0 / 0 / 0 (0\%) & 36 (2\%) / 0 (0\%) / 2 (0\%) & 0 (0\%) / 38 (2\%) & 38 (2\%) \\
     & 0 / 0 / 0 (0\%) & 14 (1\%) / 0 (0\%) / 0 (0\%) & 2 (0\%) / 14 (1\%) & 14 (1\%) \\
    \midrule
    \multirow{4}{*}{UNet} & 0 / 0 / 11 (1\%) & 37 (2\%) / 1 (0\%) / 1 (0\%) & 0 (0\%) / 39 (2\%) & 40 (2\%) \\
     & 0 / 0 / 0 (0\%) & 36 (2\%) / 0 (0\%) / 4 (0\%) & 0 (0\%) / 40 (2\%) & 40 (2\%) \\
     & 0 / 0 / 1 (0\%) & 44 (2\%) / 0 (0\%) / 8 (0\%) & 0 (0\%) / 52 (3\%) & 52 (3\%) \\
     & 0 / 0 / 0 (0\%) & 57 (3\%) / 0 (0\%) / 7 (0\%) & 0 (0\%) / 64 (4\%) & 64 (4\%) \\
\end{tabular}
\end{table*}

\subsubsection{Typical error patterns} \label{a:RB_errors}
In this section, typical incorrect predictions of the analysed networks for the rolling ball are presented to illustrate the above analysis and the error rate of the networks. The visible errors or inaccuracies are described in the caption of the respective figures.
\begin{figure}[h!]
    \centering
    \includegraphics[width=1\linewidth]{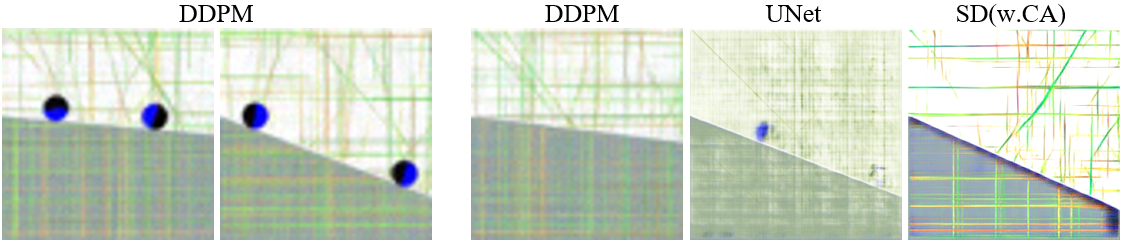}
    \caption{\small Several or no balls are present in the predictions (left: several balls only occur with the diffusion network; right: no ball present)}
    \label{fig:Err_RB_No_or_2_balls}
\end{figure}
\begin{figure}[h!]
    \centering
    \includegraphics[width=1\linewidth]{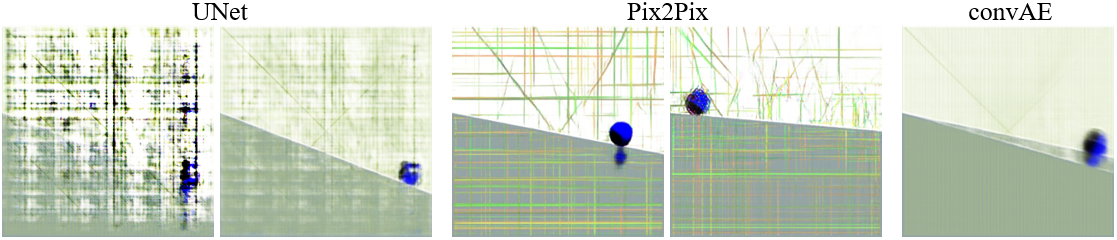}
    \caption{\small Typical artefacts that occur for different trained model}
    \label{fig:Err_RB_artefacts}
\end{figure}
\begin{figure}[h!]
    \centering
    \includegraphics[width=1\linewidth]{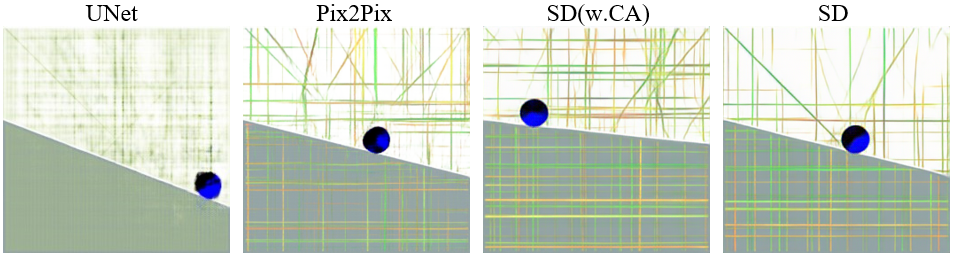}
    \caption{\small Imperfect color separation between both ball halves (occurs mostly for the Pix2Pix model)}
    \label{fig:Err_RB_colorsep}
\end{figure}
\begin{figure}[h!]
    \centering
    \includegraphics[width=1\linewidth]{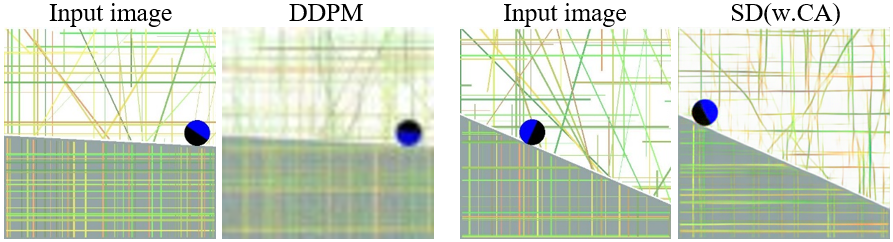}
    \caption{\small Position of the target ball ahead of the start ball along the X-axis (mostly for diffusion). The predicted ball is slightly to the left of the starting ball position on the generated samples.}
    \label{fig:Err_RB_TargetAheadStart}
\end{figure}
\begin{figure}[h!]
    \centering
    \includegraphics[width=1\linewidth]{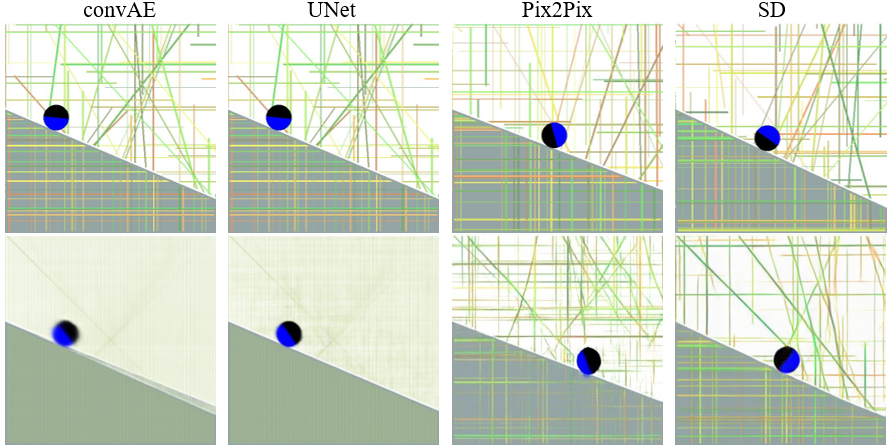}
    \caption{\small Errors concerning the ball position and rotation (1st line: true images; 2nd line: corresponding network prediction)}
    \label{fig:Err_RB_WrongPosRot}
\end{figure}

\vspace{0.5em}

Finally, the evaluation of the rolling case can be concluded with a brief look at the physics equations.
As already indicated in the main part of this paper, the largest errors can be traced back to the parts of the equation that contain a double derivative. The prediction of the x-coordinate and the ball angle are the most error-prone and also occur in the equations with $x$ and $\psi$ double derivatives. In the Y direction, the significant improvement in the position prediction can be explained by the fact that there is no movement along this axis and therefore the complex calculation parts of the bouncing case are omitted.

\clearpage

%% file: appendix/datacard.tex
\FloatBarrier


\section*{Supplementary material}

# E    Datacard

| PhysicsGen | This assembled dataset comprises three distinct physical-based image-to-image translation tasks, totaling 300,000 samples across diverse domains. The datasets include: |
|---|---|
| | **Urban Sound Propagation:** Expanding on research into urban sound propagation, this dataset contains 25,000 data points extracted from 10 different cities. Each city is represented by 5,000 locations within a 500m x 500m area, utilizing OpenStreetMap imagery where buildings are marked with black pixels and open spaces with white pixels. |
| | **Lens Distortion Correction:** Derived from the CelebA dataset, this dataset focuses on correcting lens distortion in facial images. |
| | **Rolling and Bouncing Ball Dynamics:** This dataset captures the physical dynamics of rolling and bouncing balls, intended for studies in motion prediction. |
| | Each dataset facilitates the development of generative models by providing high-resolution (256x256) imagery and diverse scenarios. |
| **DATASET LINK** | **DATA CARD AUTHOR(S)** |
| https://doi.org/10.5281/zenodo.11448582 | **Martin Spitznagel, IMLA**: Owner |
| | **Jan Vaillant, IMLA / Herrenknecht AG**:  Owner |
| | **Janis Keuper, IMLA**:  Owner |

# Authorship

## Publishers

| PUBLISHING ORGANIZATION(S) | INDUSTRY TYPE(S) | CONTACT DETAIL(S) |
|---|---|---|
| Institute for Machine Learning and Analytics (IMLA) | Academic – Tech | **Affiliation:** Offenburg University<br><br>**Website:** https://imla.hs-offenburg.de/ |

## Funding Sources

| INSTITUTION(S) | FUNDING OR GRANT SUMMARY(IES) |
|---|---|
| German Federal Ministry of Education and Research (BMBF) | This project is part of the "Forschung an Fachhochschulen in Kooperation mit Unternehmen (FH-Kooperativ)" program, within the joint project KI-**Bohrer** [https://www.ki-bohrer.de/] and is funded under the grant number 13FH525KX1. |

# Sound Propagation - Dataset Overview

| DATA SUBJECT(S) | DATASET SNAPSHOT | CONTENT DESCRIPTION |
|---|---|---|
| Data about natural phenomena<br><br>Data about places and objects | | A data point in this dataset consists of two main components: an urban layout image from OpenStreetMap and a corresponding sound distribution map. The urban layout image is a 500m x 500m area depiction where buildings are marked in black and open spaces in white. The sound distribution map generated using NoiseModelling v4.0, illustrates the sound dynamics within that urban environment at resolutions of 512x512 or 256x256.<br><br>The dataset comprises four subsets: **Baseline** for basic sound behavior, **Reflection** examining sound wave interactions with surfaces, **Diffraction** focusing on sound navigating around objects, and **Combined** which merges reflection, diffraction and changing environmental factors like temperature and humidity. |

| Size of Dataset | ~5 GB |
|---|---|
| Number of Instances | ~100,000 |
| Training | 19908 x 4 |
| Evaluation | 3732 x 4 |
| Test | 1244 x 4 |

**Additional Notes:** The dataset is segmented into four distinct subsets, each tailored to explore specific aspects of sound propagation in urban environments: Baseline, Reflection, Diffraction, and Combined.

## Dataset Version and Maintenance

| MAINTENANCE STATUS | VERSION DETAILS | MAINTENANCE PLAN |
|---|---|---|
| Regularly Updated<br><br>(New versions of the dataset have been or will continue to be made available.) | **Current Version:** 2.0<br><br>**Last Updated:** 06/2024<br><br>**Release Date:** 06/2024 | **Feedback:**<br>martin.spitznagel@hs-offenburg.de |

# Sound Propagation - Example of Data Points

| PRIMARY DATA MODALITY | SAMPLING OF DATA POINTS | DATA FIELDS |
|---|---|---|

Multimodal
- Image Data
- Geospatial Data
- Tabular Data

| Field Name | Field Value | Description |
|---|---|---|
| **lat** | float | Latitude of the sound measurement location. |
| **long** | float | Longitude of the sound measurement location. |
| **db** | Object | Key-value pairs of sound levels in decibels for a given frequency (lwd{fqz}). |
| **soundmap** | string | Path to 256x256 resolution sound distribution image. |
| **soundmap_512** | string | Path to 512x512 resolution sound distribution image. |
| **osm** | string | Path to Open Street Map image showing urban layout. |
| **temperature** | float | Temperature (°C) at the location. |
| **humidity** | float | Humidity (%) at the location. |
| **yaw** | float | Orientation of the noise source. Can be empty. |
| **sample_id** | int | Unique identifier for the data point. |

```
{
  "lat": 48.030229082138526,
  "long": 11.367773397906852,
  "db": {"lwd500": 69},
  "soundmap":
  "./soundmaps/256/0_LEQ_256.png",
  "soundmap_512":
  "./soundmaps/512/0_LEQ_512.png",
  "osm":
"./buildings/osm_23747.png",
  "temperature": 12,
  "humidity": 35,
  "yaw": None,
  "sample_id": "23747"
}
```

Below is an example of an OSM and Simulated Sound Propagation pair:

OSM:

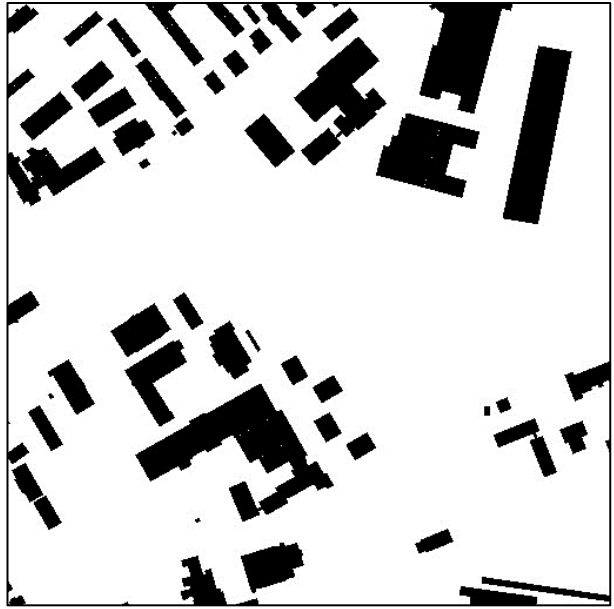

Simulated Sound Propagation:

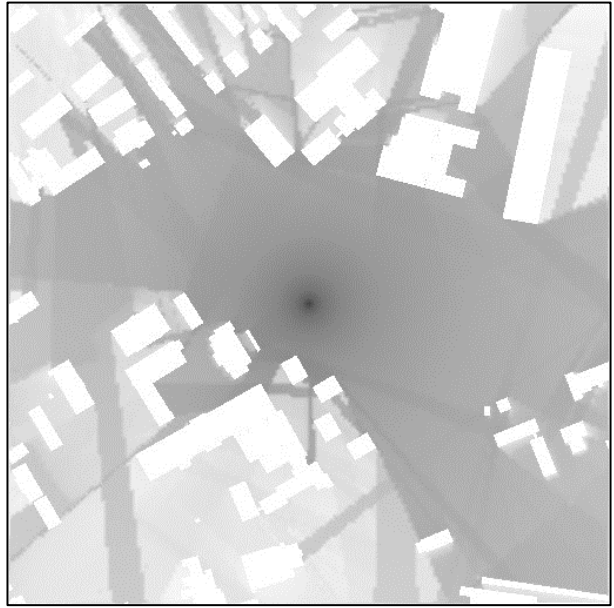

# Sound Propagation - Provenance

## Collection

| METHOD(S) USED | METHODOLOGY DETAIL(S) | SOURCE DESCRIPTION(S) |
|---|---|---|
| - API<br>- Physical Simulation Framework | **Overpass API**<br><br>**Source:** The Overpass API is a read-only API that serves up custom selected parts of the OSM map data. It acts as a database over the web: the client sends a query to the API and gets back the data set that corresponds to the query.<br><br>**Platform:** https://overpass-api.de/<br><br>**Is this source considered sensitive or high-risk?** [Yes / **No**]<br><br>**Dates of Collection:** [10 2023 - 12 2024]<br><br>**Primary modality of collected data:**<br>Geospatial Data<br><br>**NoiseModelling v4.0**<br><br>**Source:** An advanced simulation tool engineered for accurate modeling of sound dynamics within urban environments.<br><br>**Platform:** https://github.com/Universite-Gustave-Eiffel/NoiseModelling<br><br>**Is this source considered sensitive or high-risk?** [Yes / **No**]<br><br>**Dates of Collection:** [10 2023 - 12 2024]<br><br>**Primary modality of collected data:**<br>Geospatial Data | **OSM Buildings:** This source provides images from Open Street Map (OSM) that depict urban layouts, specifically focusing on buildings within cities. In these images, black pixels represent buildings, and white pixels indicate open spaces.<br><br>**Sound Propagation:** This component of the dataset involves simulated sound distribution images around urban centers, where the noise source is placed at the center. |

## Collection Criteria

### DATA SELECTION

**Location Sampling:** The locations are randomly sampled across 5 cities/areas:
["Hamburg", "Hannover", "Augsburg", "Bonn", "Muenchen"]

### DATA INCLUSION

**Enough Obstacles:** At least 10 Buildings within a circle r=200m around the sound source.
**No Obstacle to close:** No Building within a r=50m circle around the sound source.

**Additional Notes:**

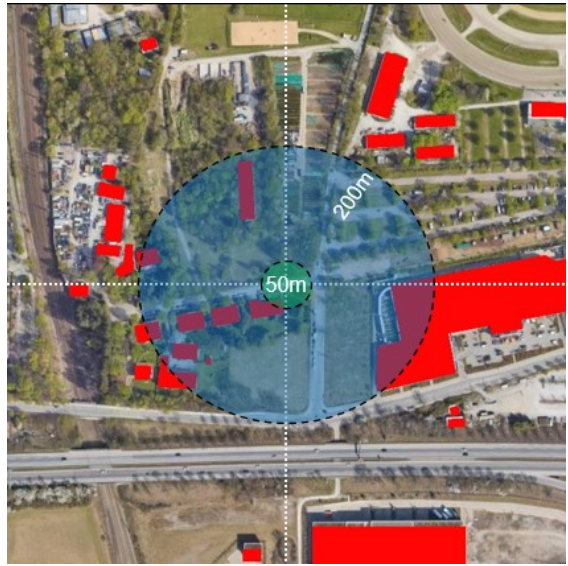

### DATA EXCLUSION

**Additional Notes:** If the Data Inclusion criteria is not met, the data is excluded.
No additional exclusion criteria are introduced.

# Lens Distortion - Dataset Overview

| DATA SUBJECT(S) | DATASET SNAPSHOT | CONTENT DESCRIPTION |
|---|---|---|
| Data about natural phenomena | | This dataset provides the lens parameters corresponding to images from the CelebA dataset in a structured CSV format. Intended for use in computer vision research focusing on the effects of different lens configurations on image characteristics, the dataset organizes parameters across training, evaluation, and testing subsets. It includes scripts available on our [GitHub repository](#) for researchers to reproduce the image samples within the bounds of copyright law. |

| Size of Dataset | ~2 MB |
|---|---|
| Number of Instances | ~100,000 |
| Training | 40000 x 2 |
| Evaluation | 7500 x 2 |
| Test | 2500 x 2 |

**Additional Notes:** The dataset is segmented into two subsets for p1 and p2.

# Dataset Version and Maintenance

| MAINTENANCE STATUS | VERSION DETAILS | MAINTENANCE PLAN |
|---|---|---|
| Regularly Updated<br>(New versions of the dataset have been or will continue to be made available.) | **Current Version:** 1.0<br><br>**Last Updated:** 06/2024<br><br>**Release Date:** 06/2024 | **Feedback:**<br>martin.spitznagel@hs-offenburg.de |

# Lens Distortion - Example of Data Points

| PRIMARY DATA MODALITY | SAMPLING OF DATA POINTS | DATA FIELDS |
|---|---|---|
| Tabular Data | | |

| Field Name | Field Value | Description |
|---|---|---|
| **label_path** | string | Path to the label file associated with the image sample. |
| **fx** | float | The focal length of the camera lens used to capture the image. |
| **k1, k2, k3** | float | Coefficients representing the radial distortion introduced by the lens. |
| **p1, p2** | float | Coefficients representing the tangential distortion of the lens. |
| **cx** | float | The x-coordinate of the principal point of the image, which is the point on the image sensor where the lens is focused. |
| **distortion_path** | string | Path to the file containing distortion image. |

TYPICAL DATA POINT

```
{
  "label_path": "labels_50k/label_0.jpg",
  "fx": 200,
  "k1": 0.0,
  "k2": 0.0,
  "k3": 0.0,
  "p1": 0.13393540720902974,
  "p2": 0.0,
  "cx": 128,
  "distortion_path": "true/y_0.jpg"
}
```

# Lens Distortion - Provenance

## Collection

| METHOD(S) USED | METHODOLOGY DETAIL(S) | SOURCE DESCRIPTION(S) |
|---|---|---|
| - Python Package | **OpenCV**<br><br>**Source:** OpenCV (Open Source Computer Vision Library) is an open-source computer vision and machine learning software library. The library provides a common infrastructure for computer vision applications and accelerates the use of machine perception in commercial products. It contains more than 2500 optimized algorithms, including a comprehensive set of both classic and state-of-the-art computer vision and machine learning techniques.<br><br>**Platform:** https://opencv.org/<br><br>**Is this source considered sensitive or high-risk?** [Yes / **No**]<br><br>**Dates of Collection:** Not applicable (continuously updated)<br><br>**Primary modality of collected data:**<br>Image Processing Data | |

# BALL - Dataset Overview

| DATA SUBJECT(S) | DATASET SNAPSHOT | CONTENT DESCRIPTION |
|---|---|---|
| Data about natural phenomena | | The third physics problem investigated in this publication is the movement of a rolling or bouncing ball. The aim here is to evaluate the ability of generative AI to map kinematic movements of physics: to investigate how well the networks can predict the position and rotation of the ball along an inclined surface for a defined time interval after the input image. |

| Size of Dataset | ~7 GB |
|---|---|
| Number of Instances | ~100,000 |
| Rolling - Training | 27840 |
| Rolling - Evaluation | 50 |
| Rolling - Test | 1800 |
| Bouncing - Training | 44835 |
| Bouncing - Evaluation | 58 |
| Bouncing - Test | 1600 |

**Additional Notes:** The dataset is segmented into two distinct subsets, for a rolling ball and a bouncing ball.

# Dataset Version and Maintenance

| MAINTENANCE STATUS | VERSION DETAILS | MAINTENANCE PLAN |
|---|---|---|
| Regularly Updated (New versions of the dataset have been or will continue to be made available.) | **Current Version:** 1.0 <br> **Last Updated:** 06/2024 <br> **Release Date:** 06/2024 | **Feedback:** martin.spitznagel@hs-offenburg.de |

# BALL - Example of Data Points

| PRIMARY DATA MODALITY | SAMPLING OF DATA POINTS | DATA FIELDS |
|---|---|---|

**Multimodal**

- Image Data
- Tabular Data

| Field Name | Field Value | Description |
|---|---|---|
| **ImgName** | float | The filename of the image pair capturing the rolling ball at a specific instance. |
| **StartHeight** | int | The initial height from which the ball starts. |
| **GroundIncli** | int | The angle of incline of the ground on which the ball is rolling, expressed in degrees. |
| **InputTime** | int | The simulation time when the ball begins its movement. |
| **TargetTime** | int | The timestamp at which the prediction should be made, indicating when the model estimates the ball will reach a predetermined target point or condition. |

## TYPICAL DATA POINT

```
 {
  "ImgName": "17",
  "StartHeight": 0,
  "GroundIncli": -3,
  "InputTime": 99,
  "TargetTime": 5910
 }
```

## EXAMPLE OF DATA POINT

Below is an example of an OSM and Simulated Sound Propagation pair:

t_0:

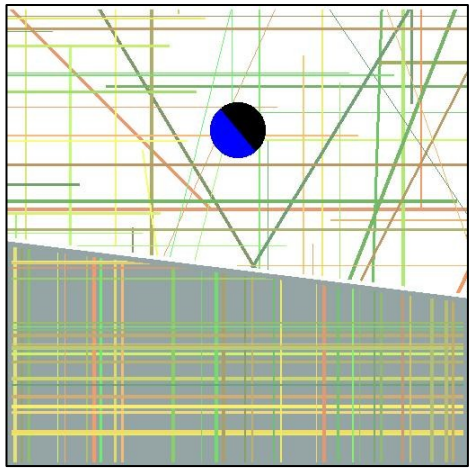

t_n:

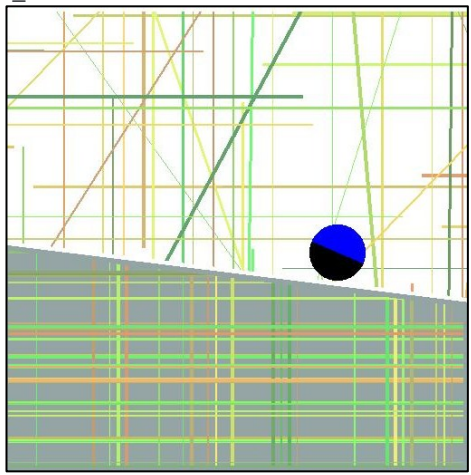

# BALL - Provenance

## Collection

| METHOD(S) USED | METHODOLOGY DETAIL(S) | SOURCE DESCRIPTION(S) |
|---|---|---|
| - Physical Simulation Framework | **Pymunk**<br><br>**Source:** Pymunk is a physics simulation library based on Chipmunk, which provides a fast, lightweight 2D rigid body physics library for Python. It enables the creation and manipulation of physics simulations in a simple manner, offering precise control over elements like mass, gravity, and friction.<br><br>**Platform:** http://www.pymunk.org/<br><br>**Is this source considered sensitive or high-risk?** [Yes /**No**]<br><br>**Primary modality of collected data:**<br>Physical Simulation Data | |

# Motivations & Intentions

## Motivations

| PURPOSE(S) | DOMAIN(S) OF APPLICATION | MOTIVATING FACTOR(S) |
|---|---|---|
| Research | `Generative Models`, `1-step Physic Simulation`, `Sound Propagation`, `Machine Learning` | The release of this dataset is driven by the potential of generative learning models to understand and simulate complex physical phenomena. <br><br> Our dataset of 300,000 image-pairs supports research into three specific physical simulation tasks: urban sound propagation, lens distortion correction, and the dynamics of rolling and bouncing balls. By providing these data, along with baseline evaluations, we aim to address several fundamental research questions: <br><br> - Learning Capability: Can generative models accurately learn and replicate complex physical relationships from input-output image pairs? <br> - Efficiency Gains: What computational speedups can be achieved by substituting traditional, differential equation-based simulations with generative model predictions? |

## Intended Use

| DATASET USE(S) | RESEARCH AND PROBLEM SPACE(S) |
|---|---|
| Safe for research use | The dataset targets three key physical phenomena: urban noise propagation, lens distortion, and the dynamics of rolling and bouncing balls. It is crafted to develop models capable of: <br> - **Urban Noise Propagation**: Predicting sound distribution around urban buildings. <br> - **Lens Distortion**: Correcting optical distortions in photographs. <br> - **Rolling and Bouncing Balls**: Simulating motion trajectories under various conditions. |

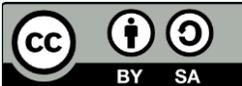

